\documentclass{article} % For LaTeX2e
\usepackage{iclr2025_conference,times}

\usepackage{amsmath,amsfonts,bm}

\def\eqref#1{equation~\ref{#1}}
\def\1{\bm{1}}

\DeclareMathAlphabet{\mathsfit}{\encodingdefault}{\sfdefault}{m}{sl}
\SetMathAlphabet{\mathsfit}{bold}{\encodingdefault}{\sfdefault}{bx}{n}

\usepackage{hyperref}

\usepackage{url}
\usepackage{booktabs}       % professional-quality tables
\usepackage{amsfonts}       % blackboard math symbols
\usepackage{nicefrac}       % compact symbols for 1/2, etc.
\usepackage{microtype}      % microtypography
\usepackage{amsmath,bm}
\usepackage{mathtools}
\usepackage{booktabs} % for professional tables
\usepackage{multirow} % for multirow in tables
\usepackage{tabularx}
\usepackage{pifont}
\usepackage{algorithm}
\usepackage{algpseudocode}
\usepackage{subfig}
\usepackage{amssymb}
\usepackage{wrapfig}
\usepackage{enumitem}
\usepackage{makecell}

\usepackage[pdftex]{graphicx}

\usepackage{parskip}

\newcommand{\xhdr}[1]{\noindent{{\bf #1.}}}

\newcommand{\our}{TGDMat}

\newcommand{\blfootnote}[1]{{\renewcommand{\thefootnote}{\roman{footnote}}\footnotetext[0]{#1}}}

\title{Periodic Materials Generation using Text-Guided Joint Diffusion Model}

\author{Kishalay Das$^{1,\dag }$, Subhojyoti Khastagir$^1$, Pawan Goyal$^1$, Seung-Cheol Lee$^2$,\\
\textbf{Satadeep Bhattacharjee$^2$}, \textbf{Niloy Ganguly$^1$}  \\ 
$^1$ Indian Institute of Technology, Kharagpur, India\\
$^2$ Indo Korea Science and Technology Center, Bangalore, India
}
\iclrfinalcopy % Uncomment for camera-ready version, but NOT for submission.
\begin{document}

\maketitle

\blfootnote{
\dag Correspondence to Kishalay: kishalaydas@kgpian.iitkgp.ac.in
}

\begin{abstract}
Equivariant diffusion models have emerged as the prevailing approach for generating novel crystal materials due to their ability to leverage the physical symmetries of periodic material structures. However, current models do not effectively learn the joint distribution of atom types, fractional coordinates, and lattice structure of the crystal material in a cohesive end-to-end diffusion framework.  Also, none of these models work under realistic setups, where users specify the desired characteristics that the generated structures must match. In this work, we introduce \our{}, a novel text-guided diffusion model designed for 3D periodic material generation. Our approach integrates global structural knowledge through textual descriptions at each denoising step while jointly generating atom coordinates, types, and lattice structure using a periodic-E(3)-equivariant graph neural network (GNN).  Extensive experiments using popular datasets on benchmark tasks reveal that \our{} outperforms existing baseline methods by a good margin. Notably, for the structure prediction task, with just one generated sample, \our{} outperforms all baseline models, highlighting the importance of text-guided diffusion. Further, in the generation task, \our{} surpasses all baselines and their text-fusion variants, showcasing the effectiveness of the joint diffusion paradigm. Additionally, incorporating textual knowledge reduces overall training and sampling computational overhead while enhancing generative performance when utilizing real-world textual prompts from experts. Code is available at \url{https://github.com/kdmsit/TGDMat}
\end{abstract}

\section{Introduction}
\label{intro}
Screening 3D periodic structures and their atomic compositions to identify novel crystal materials with specific chemical properties remains a long-standing challenge in the materials design community. These materials have been fundamental to key innovations such as the development of batteries, solar cells, semiconductors etc.~\citep{ butler2018machine,desiraju2002cryptic}. Historically, there have been attempts to generate novel materials by conducting resource-intensive and time-consuming simulations based on Density Functional Theory (DFT)~\citep{kohn1965self}.  Recently, the equivariant diffusion models~\citep{jiao2023crystal,luo2023towards,xie2021crystal} have demonstrated great potential to generate stable 3D periodic structures of new crystal materials. 

However, these models possess several inherent limitations. 1) None of these existing SOTA models learns the joint distribution of atom coordinates, types, and lattice structure of the material through an end-to-end diffusion network. Existing models like CDVAE~\citep{xie2021crystal} and SyMat~\citep{luo2023towards} learn lattice parameters and atom types separately using a VAE model and further use a score network to learn the conditional distribution of atom coordinates given atom types and lattice. DiffCSP~\citep{jiao2023crystal}, on the other hand, focuses primarily on structure prediction task where it assumes atom types are given and predict the stable crystal structure (lattice and coordinates). 
2) Furthermore, these models use SE(3)-equivariant GNNs as backbone denoising network, which largely relies on messages passing around the local neighborhood of the atoms. Hence they fail to incorporate global structural knowledge into the diffusion process, which can enhance the diffusion performance. 3) Finally, these models are unconditional by design. From initial noisy structures without any external constraints, they generate stable crystal structures, which are distributionally similar to structures of the training dataset. 
This setup may have limited utility
in real-world scenarios, as it lacks a mechanism for users to specify a criteria for the material to be generated.  In a realistic setup, users would want to specify certain key details about the target material, like the chemical formula, space group, crystal symmetry, bond lengths, chemical properties, etc as input to the diffusion model, which the generated structure must then match.

In this paper, we propose, \our{}, a novel \textit{\underline{T}ext-\underline{G}uided \underline{D}iffusion Model for \underline{Mat}erial Generation} that mitigates the limitations mentioned above and enhances the generation capability. Though Text Guided Diffusion Models (TGDMs) produce impressively high-quality data in the form of images~\citep{nichol2021glide,ramesh2022hierarchical,rombach2022high,saharia2022photorealistic}, audio~\citep{kreuk2022audiogen,yang2207discrete}, video~\citep{du2024learning}, molecules~\citep{gong2024text,luo2023text} etc, it remains largely unexplored in periodic material generation.
% \noteng{my suggestion is to taken the below part above -- the readers know what is the capability and then the design}
Text-guided diffusion model for new material generation has some key benefits. First, we can leverage popular tools like Robocrystallographer~\citep{ganose2019robocrystallographer} to generate a textual description of the material which provides a rich and diverse set of global structural knowledge like chemical formula, lattice constraint, space group number, crystal symmetry, chemical properties, etc. We believe this additional information is helpful for diffusion models in learning underlying crystal geometry. Second, it provides end users the flexibility to use custom prompts to guide the material generation process, ensuring that the resulting material aligns with the user's provided description. Towards that goal, we first develop a diffusion model that jointly generates the atom coordinates, atom types, and lattice structure of crystal materials using a periodic E(3)-equivariant denoising model, satisfying periodic E(3) invariance properties of learned data distribution. Subsequently, we fuse textual information into the reverse diffusion process, which guides the denoising process in predicting material structure as specified by the textual description.\\
To sum up, our novel contributions in this work are as follows:
\begin{itemize}
    % \setlength\itemsep{0.3em}
    % \noteng{we perhaps need to write we prove the importance of text-guided generation}
    \item To the best of our knowledge, we are the first to explore text-guided diffusion for material generation. Our proposed \our{} bridges the gap between natural language understanding and material structure generation.
    \item Unlike prior models, \our{} conducts joint diffusion on lattices, atom types, and coordinates, enhancing its ability to accurately capture the crystal geometry. Additionally, incorporating global structural knowledge through textual descriptions at each denoising step improves TGDMat’s ability to generate plausible materials with valid and stable structures.
    % \item Through extensive experiments using popular datasets on benchmark tasks we show that text guidance can improve the generation capability of existing SOTA diffusion models for crystal materials. Moreover, in the generation task, \our{} outperforms text-fusion variants of SOTA models with good margin, showcasing the effectiveness of the text guided joint diffusion paradigm.
    \item Extensive experiments using popular datasets on benchmark tasks show \our{} outperforms baseline models with a good margin. Notably, in CSP task, with just one generated sample, \our{} outperforms all baseline models, highlighting the importance of text-guided diffusion. Moreover, in the generation task, \our{} outperforms all baselines and their text-fusion variants, showcasing the effectiveness of the joint diffusion paradigm.
    \item Fusing textual knowledge reduces the overall computational cost for both training and inference of the diffusion model. Moreover, when applied to real-world custom text prompts by experts, \our{} demonstrates rich generative capability under general textual conditions.
\end{itemize}
\begin{figure}[ht]
	\centering
	\includegraphics[width=\columnwidth]{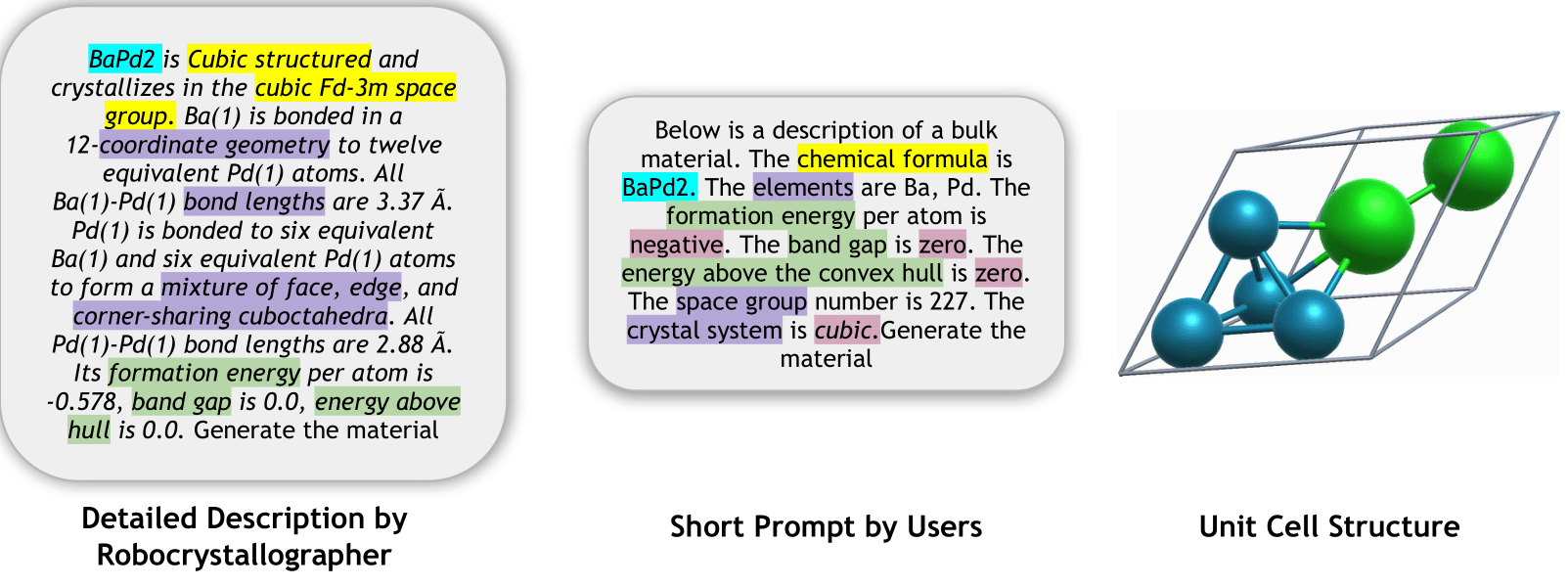}
	\caption{Detailed textual description generated by Robocrystallographer, less-detailed prompts by domain experts, and crystal unit cell structure of $\mathbf{BaPd_2}$.}
	\label{fig:text}
\end{figure}
\section{Preliminaries : Crystal Structure Representation}
\label{prelim}
Crystal material can be modeled by a minimal \textit{unit cell}, which gets repeated infinite times in 3D space on a regular lattice to form the periodic crystal structure. Given a material with $N$ number of atoms in its unit cell, we can describe the unit cell by two matrices: \textit{Atom Type Matrix (\textbf{\textit{A}})} and \textit{Coordinate Matrix ($\textbf{\textit{X}}$)}. Atom Type Matrix $\textbf{\textit{A}} =[\textbf{\textit{a}}_1,\textbf{\textit{a}}_2,...,\textbf{\textit{a}}_N]^T \in \mathbb{R}^{N \times k}$ denotes set of atomic type in one hot representation (k: maximum possible atom types). On the other hand, Coordinate Matrix $\textbf{\textit{X}}=[\textbf{\textit{x}}_1,\textbf{\textit{x}}_2,...,\textbf{\textit{x}}_N]^T \in \mathbb{R}^{N \times 3}$ denotes atomic coordinate positions, where $\textbf{\textit{x}}_i \in \mathbb{R}^{3}$ corresponds to coordinates of $i^{th}$ atom in the unit cell. Further, there is an additional \textit{Lattice Matrix} $\textbf{\textit{L}}=[\textbf{\textit{l}}_1,\textbf{\textit{l}}_2,\textbf{\textit{l}}_3]^T \in \mathbb{R}^{3 \times 3}$, which describes how a unit cell repeats itself in the 3D space towards $\textbf{\textit{l}}_1,\textbf{\textit{l}}_2$ and $\textbf{\textit{l}}_3$ direction to form the periodic 3D structure of the material. Formally, a given material can be defined as $ \mathbf{\textit{M}}=(\textbf{\textit{A}},\textbf{\textit{X}},\textbf{\textit{L}})$ and we can represent its infinite periodic structure as $\mathbf{\hat{X}}  = \{ \hat{x}_i |  \hat{x}_i = \textit{x}_i + \sum_{j=1}^{3} k_j{l}_j \}$; $\mathbf{\hat{A}}  = \{ \hat{a}_i |  \hat{a}_i = {a}_i \}$ where $k_1,k_2,k_3, i \in Z, 1 \leq i \leq N$.\\\\
\xhdr{Invariances in Crystal Structure}
\label{prelim_inv}
The basic idea of using generative models for crystal generation is to learn the underlying data distribution of material structure $p(\textbf{\textit{M}})$. Since crystal materials satisfy physical symmetry properties~\citep{dresselhaus2007group,zee2016group}, one of the major challenges here is the learned distribution must satisfy periodic E(3) invariance i.e. invariance to permutation, translation, rotation, and periodic transformations. A formal definition of these invariance properties is provided in Appendix \ref{appendix_prelim_inv}.
\section{Related Work: Periodic Material Generation}
\label{related}
Recently, the majority of the research on material generation focuses on using popular generative models like VAEs~\citep{kingma2013auto}, GANs~\citep{goodfellow2014generative} or Diffusion Models~\citep{song2019generative,song2020improved,ho2020denoising} to generate 3D periodic structures of materials~\citep{hoffmann2019data, noh2019inverse, ren2020inverse, kim2020generative, court20203, long2021constrained, zhao2021high, xie2021crystal, jiao2023crystal, luo2023towards,zeni2023mattergen, yang2023scalable, jiao2024space, miller2024flowmm}.  In specific, state-of-the-art models like CDVAE~\citep{xie2021crystal} and SyMat~\citep{luo2023towards} combine VAEs and score-based diffusion models to work directly with atomic coordinates, ensuring euclidean and periodic invariance using equivariant graph neural networks(GNNs). 
Moreover, DiffCSP~\citep{jiao2023crystal} focuses on structure prediction, jointly optimizing atom coordinates and lattice using a diffusion framework given atomic composition. We provided a comprehensive literature review of other related works in Appendix \ref{appendix_related}. \\\\
\xhdr{Relations with Prior Methods} Among existing models, DiffCSP comes close to our methodology, however, our work differs in multiple ways.  DiffCSP primarily focuses only on the Structure Prediction (\textit{CSP}) task and they didn't explore the Random Generation (\textit{Gen}) task, whereas \our{} focuses on both tasks.  Moreover, unlike DiffCSP, \our{} can leverage the informative textual descriptions during the denoising process and can jointly learn lattices, atom types, and coordinates, which makes \our{} a more flexible and robust generative model for new material generation.
\section{Methodology}
\label{method}
\subsection{Problem Formulation}
\label{problem_formulation}
In this work, given the textual description, we focus on generating a stable crystal structure that aligns with the provided textual description. Formally, given a dataset $\mathcal{M}=\{ \textbf{\textit{M}}_i,\textbf{\textit{T}}_i \}$, containing crystal structure {$\textbf{\textit{M}}_i = (\textbf{\textit{A}}_i,\textbf{\textit{X}}_i,\textbf{\textit{L}}_i)$} and its text description ($\textbf{\textit{T}}_i$), the goal of text guided crystal generation problem is to capture the underlying conditional data distribution $p(\textbf{\textit{M}} | \textbf{\textit{T}})$ via learning a generative model $f_{\theta} (\textbf{\textit{M}} | \textbf{\textit{T}})$, where $\theta$ is a set of learnable parameters. While training, we need $f_{\theta}$ to ensure that the learned distribution is invariant to different symmetry transformations mentioned in Section \ref{prelim_inv}. Once trained, given a text description of a plausible material, the learned generative model can sample a valid and stable structure of the material, that is invariant to different symmetry transformations.
\subsection{Textual Datasets}
\label{text_data}
Leveraging textual information to guide the reverse diffusion process remains unexplored in the material design community. 
To the best of our knowledge, there is currently no text data available for materials in benchmark databases (mentioned in Section \ref{results_data_task}). Hence, we first curate the textual data of these material databases.
Specifically, we propose two approaches for generating textual descriptions of materials, which are easy to follow. First, we utilize a freely available utility tool, \textit{Robocrystallographer}~\citep{ganose2019robocrystallographer} to generate detailed textual descriptions about the periodic structure of crystal materials. These descriptions encompass local compositional details like atomic coordination, geometry, etc. as well as global structural aspects like crystal formula, mineral type, space group information, etc. 
Secondly, we utilized shorter and less detailed prompts that are more easily interpretable by users. We extend the prompt template proposed by~\citep{gruver2024fine}, which encodes minimal information about the material like its chemical formula, constituent elements, crystal system it belongs to, and its space group number. Further, we specify a few chemical properties, and instead of mentioning their actual values, we provide generic information like negative/positive formation energy, zero/nonzero band gaps, etc. 
Detailed information regarding the two textual datasets, including their curation process is provided in Appendix \ref{appendix_text_data}. We have publicly shared the textual datasets for both benchmark material databases with the community for future use.
\subsection{Proposed Methodology : \our{}}
\label{text2mat_intro}
\begin{figure*}
	\centering
	\includegraphics[width=\columnwidth]{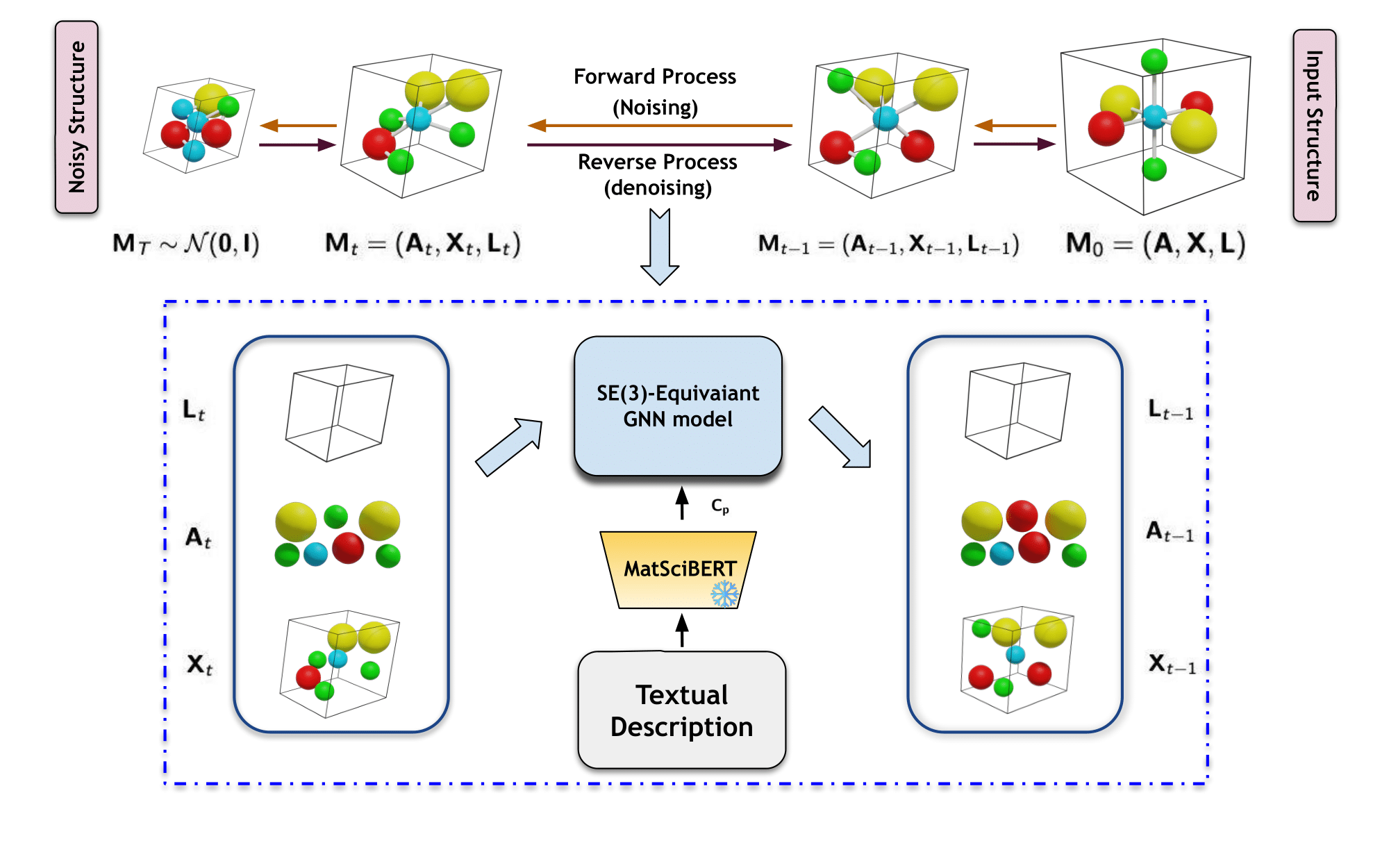}
    % \vspace{-9pt}
	\caption{Model Architecture of our proposed text guided diffusion model \our{}. At $t^{th}$ step of reverse diffusion, given $ \textbf{\textit{M}}_t=(\textbf{\textit{A}}_t,\textbf{\textit{X}}_t,\textbf{\textit{L}}_t)$, we use periodic-E(3)-equivariant GNN model guided by contextual representation of the textual prompts ($\textbf{\textit{C}}_\textbf{\textit{p}}$) to generate $ \textbf{\textit{M}}_{t-1}=(\textbf{\textit{A}}_{t-1},\textbf{\textit{X}}_{t-1},\textbf{\textit{L}}_{t-1})$}
	\label{fig:architecture}
\end{figure*}

 Our proposed model, \our{} (Fig. \ref{fig:architecture}), uses an equivariant diffusion model guided by contextual representation of the textual description ($\textbf{\textit{C}}_\textbf{\textit{p}}$) to generate a new crystal structure $ \textbf{\textit{M}}=(\textbf{\textit{A}},\textbf{\textit{X}},\textbf{\textit{L}})$. Unlike prior methods~\citep{jiao2023crystal,luo2023towards,xie2021crystal}, our method jointly diffuses \textbf{\textit{A}}, \textbf{\textit{X}}, \textbf{\textit{L}} to learn the underlying data distribution of crystal structure $p (\mathbf{\textit{M}} | \textbf{\textit{C}}_\textbf{\textit{p}})$. Diffusion models~\citep{ho2020denoising,song2019generative,song2020improved} are popular generative models that are formulated using a T steps Markov Chain. Given an input crystal material $ \textbf{\textit{M}}_0=(\textbf{\textit{A}}_0,\textbf{\textit{X}}_0,\textbf{\textit{L}}_0)$, the forward process gradually add noise to $\textbf{\textit{A}}_0,\textbf{\textit{X}}_0,\textbf{\textit{L}}_0$ independently over T steps and the reverse denoising process samples a noisy structure $ \textbf{\textit{M}}_T=(\textbf{\textit{A}}_T,\textbf{\textit{X}}_T,\textbf{\textit{L}}_T)$ from a prior distribution and reconstruct back $ \textbf{\textit{M}}_0$ using some GNN model. At each $t^{th}$ step of denoising $( T \geq t \geq 0)$, the contextual representation of the crystal textual description ($\textbf{\textit{C}}_\textbf{\textit{p}}$) will guide the diffusion process so that the intermediate structure $ \textbf{\textit{M}}_t$ aligns the target 3D structure constrained on textual conditions. Moreover, the learned distribution of material structure must satisfy periodic E(3) invariance. It is well studied in the literature~\citep{xu2022geodiff} that if the prior distribution $p(x)$ is invariant to a group and the transition probabilities of a Markov chain $y \sim p(y|x)$ exhibit equivariance, the marginal distribution of $y$ at any given time step also remains invariant to group transformations. Hence the learned distribution $p(\textbf{\textit{M}}_0)$ of the denoising model will satisfy periodic E(3) invariance if the prior distribution $p(\textbf{\textit{M}}_T)$ is invariant and the neural network used to parameterize the transition probability $q(\textbf{\textit{M}}_{t-1} | \textbf{\textit{M}}_t)$ is equivariant to permutational, translation, rotational, and periodic transformations. To satisfy that, we use periodic-E(3)-equivariant GNN model as a backbone denoising network to guide the denoising process. Next in this section, we first explain diffusion on \textbf{\textit{M}} in \ref{diffusion}, then demonstrate the text-guided denoising network in \ref{text_guided_diffusion} and finally training details in \ref{train_sample}.
\subsubsection{Joint Equivariant Diffusion on \textbf{\textit{M}}}\label{diffusion}
\xhdr{Diffusion on Lattice (\textbf{\textit{L}})} 
Since the Lattice Matrix $\textbf{\textit{L}}=[{l}_1,{l}_2,{l}_3]^T \in \mathbb{R}^{3 \times 3}$ is in continuous space, we leverage the idea of the Denoising Diffusion Probabilistic Model (DDPM)~\citep{ho2020denoising} for diffusion on \textbf{\textit{L}}. Specifically, given input lattice matrix $\textbf{\textit{L}}_0 \sim p(\textbf{\textit{L}})$, at each $t^{th}$ step, the forward diffusion process iteratively diffuses it through a transition probability $q(\textbf{\textit{L}}_{t} | \textbf{\textit{L}}_{0})$  which can be derived as $q(\textbf{\textit{L}}_{t}|\textbf{\textit{L}}_{0}) = \mathcal{N} (\textbf{\textit{L}}_{t}|\sqrt{\bar{\alpha}_{t}}\textbf{\textit{L}}_{0}, \ (1 \ - \ \bar{\alpha}_{t})\mathbf{I} )$
where, $\bar{\alpha}_{t} = \prod_{k=1}^{t} \alpha_{k}$, $\alpha_{t} = 1-\beta_{t}$ and $\{ \beta_{t} \in (0,1) \}^{T}_{t=1}$ controls the variance of diffusion step following certain noise scheduler. 
By reparameterization, we can rewrite
$\textbf{\textit{L}}_{t} = \sqrt{\bar{\alpha}_{t}}\textbf{\textit{L}}_{0} +  \sqrt{1-\bar{\alpha}_{t}}\bm{\epsilon}^{\textbf{\textit{L}}}$
where, $\bm{\epsilon}^{\textbf{\textit{L}}}$ is noise sampled from $\mathcal{N}(\mathbf{0},\mathbf{I})$, added with $\textbf{\textit{L}}_{0}$  at $t^{th}$ step to generate $\textbf{\textit{L}}_{t}$. 
After T such diffusion steps, noisy lattice matrix $\textbf{\textit{L}}_T \sim \mathcal{N}(\mathbf{0},\mathbf{I})$ is generated. During reverse denoising process, given noisy $\textbf{\textit{L}}_T \sim \mathcal{N}(\mathbf{0},\mathbf{I})$ we reconstruct true lattice structure $\textbf{\textit{L}}_0$ thorough iterative denoising step via learning reverse conditional distribution, which we formulate as $p(\textbf{\textit{L}}_{t-1} | \textbf{\textit{M}}_{t},\textbf{\textit{C}}_\textbf{\textit{p}}) = \mathcal{N} \{ \textbf{\textit{L}}_{t-1} | \mu^{\textbf{\textit{L}}}(\textbf{\textit{M}}_{t},\textbf{\textit{C}}_\textbf{\textit{p}}),\beta_{t}\frac{(1 - \bar{\alpha}_{t-1})}{(1 -\bar{\alpha}_{t})}\mathbf{I} \}$
where $\mu^{\textbf{\textit{L}}}(\textbf{\textit{M}}_{t},\textbf{\textit{C}}_\textbf{\textit{p}}) = \frac{1}{\sqrt{\alpha_{t}}}(\textbf{\textit{L}}_{t}-\frac{1 - \alpha_{t}}{\sqrt{1 - \bar{\alpha}_{t}}} \ \hat{\bm{\epsilon}}^{\textbf{\textit{L}}} (\textbf{\textit{M}}_{t},\textbf{\textit{C}}_\textbf{\textit{p}},t))$.
% Intuitively, the mean of the target reverse distribution depends on the scaled $\textbf{\textit{L}}_{t}$, after subtracting the noise added to it during the forward process which is a time-conditioned function depends on $\textbf{\textit{M}}_{t}$.
Intuitively, $\hat{\bm{\epsilon}}^{\textbf{\textit{L}}}$ needs to be subtracted from $\textbf{\textit{L}}_{t}$ to generate $\textbf{\textit{L}}_{t-1}$ and textual representation $\textbf{\textit{C}}_\textbf{\textit{p}}$ will steer this reverse diffusion process. We use a text-guided denoising network $\Phi_\theta(\textbf{\textit{A}}_{t},\textbf{\textit{X}}_{t},\textbf{\textit{L}}_{t},t,\textbf{\textit{C}}_\textbf{\textit{p}})$ to model the noise term $\hat{\bm{\epsilon}}^{\textbf{\textit{L}}} (\textbf{\textit{M}}_{t},\textbf{\textit{C}}_\textbf{\textit{p}},t)$. Following the simplified training objective proposed by~\citep{ho2020denoising}, we train denoising model using $l_2$ loss between $\hat{\bm{\epsilon}}^{\textbf{\textit{L}}}$ and $\bm{\epsilon}^{\textbf{\textit{L}}}$
\setlength{\abovedisplayskip}{0pt} 
\setlength{\abovedisplayshortskip}{0pt}
\begin{equation}
    \label{eq:lattice_loss}
        \mathcal{L}_{lattice} = \mathbb{E}_{\bm{\epsilon}^{\textbf{\textit{L}}},t \sim \mathcal{U}(1,T)}
        \lVert  \bm{\epsilon}^{\textbf{\textit{L}}} \ - \ \hat{\bm{\epsilon}}^{\textbf{\textit{L}}} {\rVert}^2_2 
\end{equation}
\xhdr{Diffusion on Atom Types (\textbf{\textit{A}})} Prior studies~\citep{jiao2023crystal,xie2021crystal} consider Atom Type Matrix $\textbf{\textit{A}}$ as the probability distribution for k classes $\in$ $\mathbb{R}^{N \times k}$ (continuous variable) and apply DDPM to learn the distribution. However for discrete data these models are inappropriate and produce suboptimal results~\citep{austin2021structured,campbell2022continuous}. Hence we consider $\textbf{\textit{A}}$ as N discrete variables belonging to k classes and leverage discrete diffusion model (D3PM)~\citep{austin2021structured} for diffusion on \textbf{\textit{A}}. In specific, with \textbf{\textit{a}} as the one-hot representation of atom \textit{a}, the transition probability for the forward process is $q(\textbf{\textit{a}}_t | \textbf{\textit{a}}_{t-1}) = Cat(\textbf{\textit{a}}_t;\textbf{\textit{p}} = \textbf{\textit{a}}_{t-1}\textbf{\textit{Q}}_{t})$, where $Cat(\textbf{\textit{a}};\textbf{\textit{p}})$ is a categorical distribution over \textbf{\textit{a}} with probabilities \textbf{\textit{p}} and $\textbf{\textit{Q}}_{t}$ is the Markov transition matrix at time step t, defined as $[\textbf{\textit{Q}}_{t}]_{i,j} = q(a_t = i | a_{t-1} = j)$. Different choices of $\textbf{\textit{Q}}_{t}$ and corresponding stationary distributions are proposed by~\citep{austin2021structured} which provides flexibility to control the data corruption and denoising process. We adopted the absorbing state diffusion process, introducing a new absorbing state [MASK] in $\textbf{\textit{Q}}_{t}$. At each time step t, an atom either stays in its type state with probability $1-\beta_{t}$ or moves to [MASK] state with probability $\beta_{t}$ and once it moves to [MASK] state, it stays there. Hence, the stationary distribution
of this diffusion process has all the mass on the [MASK] state. During denoising process, given textual representation $\textbf{\textit{C}}_\textbf{\textit{p}}$, we first sample noisy $\textbf{\textit{a}}_T$ and obtain $\textbf{\textit{a}}_0$ thorough iterative denoising step via learning reverse conditional transition $p_{\theta}(\textbf{\textit{a}}_{t-1} | \textbf{\textit{a}}_t, \textbf{\textit{C}}_\textbf{\textit{p}}) \propto \sum_{\textbf{\textit{a}}_0} q(\textbf{\textit{a}}_{t-1}, \textbf{\textit{a}}_t | \textbf{\textit{a}}_0)p_{\theta}(\textbf{\textit{a}}_0 | \textbf{\textit{a}}_t, \textbf{\textit{C}}_\textbf{\textit{p}})$. We use the text-guided denoising network $\Phi_\theta(\textbf{\textit{A}}_{t},\textbf{\textit{X}}_{t},\textbf{\textit{L}}_{t},t,\textbf{\textit{C}}_\textbf{\textit{p}})$ to model this denoising process, which is trained using following loss function :
\begin{equation}
    \label{eq:type_loss}
        \mathcal{L}_{type} = \mathcal{L}_{VB}+\lambda\mathcal{L}_{CE}
\end{equation}
where $\mathcal{L}_{VB}$ and $\mathcal{L}_{CE}$ is the variational lower bound and cross-entropy loss respectively and $\lambda$ is a hyperparameter. Details about the diffusion process and the losses $\mathcal{L}_{VB}$, $\mathcal{L}_{CE}$ are in Appendix \ref{appendix_diffusion}\\\\
\xhdr{Diffusion on Atom Coordinates (\textbf{\textit{X}})}
We can diffuse the Coordinate Matrix $\textbf{\textit{X}}=[\textbf{\textit{x}}_1,\textbf{\textit{x}}_2,...,\textbf{\textit{x}}_N]^T \in \mathbb{R}^{N \times 3}$ in two ways: either by diffusing cartesian coordinates or fractional coordinates. Prior works like CDVAE~\citep{xie2021crystal} and SyMat~\citep{luo2023towards} diffuse cartesian coordinates whereas DiffCSP~\citep{jiao2023crystal} diffuses fractional coordinates. In our setup, as we are jointly learning atom coordinates and lattice matrix, hence we follow DiffCSP and diffuse fractional coordinates. Fractional coordinates in crystal material resides in quotient space $\mathbb{R}^{N \times 3} / \mathbb{Z}^{N \times 3}$ induced by the crystal periodicity. Since the Gaussian distribution used in DDPM is unable to model the cyclical and bounded domain of \textbf{\textit{X}}, it is not suitable to apply DDPM to model \textbf{\textit{X}}. Hence at each step of forward diffusion, we add noise sampled from Wrapped Normal (WN) distribution~\citep{de2022riemannian} to \textbf{\textit{X}} and during denoising leverage Score Matching Networks~\citep{song2019generative,song2020improved} to model underlying transition probability $q(\textbf{\textit{X}}_{t}| \textbf{\textit{X}}_{0}) = \mathcal{N}_W (\textbf{\textit{X}}_{t} |\textbf{\textit{X}}_{0}, \sigma_t^2 \mathbf{I})$. In specific, at each $t^{th}$ step of diffusion, we derive $\textbf{\textit{X}}_{t}$ as : 
$\textbf{\textit{X}}_{t} = f_w(\textbf{\textit{X}}_{0} + \bm{\sigma_t}\bm{\epsilon}^{\textbf{\textit{X}}})$
where, $\bm{\epsilon}^{\textbf{\textit{X}}}$ $\sim$ $\mathcal{N}(\mathbf{0},\mathbf{I})$, $\bm{\sigma_t}$ is the noise scheduler and $f_w(.)$ is a truncation function. Given a fractional coordinate matrix \textbf{\textit{X}}, truncation function $f_w(\textbf{\textit{X}}) = (\textbf{\textit{X}} - \lfloor \textbf{\textit{X}} \rfloor)$ returns the fractional part of each element of \textbf{\textit{X}}.\\ 
As argued in~\citep{jiao2023crystal}, $q(\textbf{\textit{X}}_{t}| \textbf{\textit{X}}_{0})$ is periodic translation equivariant, and approaches uniform distribution $\mathcal{U}(0,1)$ for sufficiently large values of $\bm{\sigma_T}$. Hence during the denoising process, we first sample $\textbf{\textit{X}}_{T} \sim \mathcal{U}(0,1)$ and iteratively denoise via score network for T steps to recover back the true fractional coordinates $\textbf{\textit{X}}_{0}$. We use the text-guided denoising network $\Phi_\theta(\textbf{\textit{A}}_{t},\textbf{\textit{X}}_{t},\textbf{\textit{L}}_{t},t,\textbf{\textit{C}}_\textbf{\textit{p}})$ to model the denoising process, which is trained using the following score-matching objective function :
\begin{equation}
    \label{eq:coord_loss}
        \mathcal{L}_{coord} = \mathbb{E}_{\substack{\textbf{\textit{X}}_{t} \sim q(\textbf{\textit{X}}_{t} | \textbf{\textit{X}}_{0}) \\
        t \sim \mathcal{U}(1,T)}}
        \lVert  \nabla_{\textbf{\textit{X}}_{t}} \text{log} q(\textbf{\textit{X}}_{t} | \textbf{\textit{X}}_{0}) - \hat{\bm{\epsilon}}^{\textbf{\textit{X}}}(\textbf{\textit{M}}_{t},\textbf{\textit{C}}_\textbf{\textit{p}},t) {\rVert}^2_2
\end{equation}
where $\nabla_{\textbf{\textit{X}}_{t}} \text{log} q(\textbf{\textit{X}}_{t} | \textbf{\textit{X}}_{0}) \propto \sum_{\textbf{\textit{K}} \in \mathbb{Z}^{N \times 3}} \text{exp}(- \ \frac{\lVert \textbf{\textit{X}}_{t} - \textbf{\textit{X}}_{0} + \textbf{\textit{K}} {\rVert}^2_F }{2\bm{\sigma_t}^2})$ is the score function of transitional distribution and $\hat{\bm{\epsilon}}^{\textbf{\textit{X}}}(\textbf{\textit{M}}_{t},\textbf{\textit{C}}_\textbf{\textit{p}},t)$ denoising term. More Details are provided in Appendix \ref{appendix_diffusion}
\subsubsection{Text Guided Denoising Network} 
\label{text_guided_diffusion}
In this subsection, we will illustrate the detailed architecture of our proposed Text Guided Denoising Network $\Phi_\theta(\textbf{\textit{A}}_{t},\textbf{\textit{X}}_{t},\textbf{\textit{L}}_{t},t,\textbf{\textit{C}}_\textbf{\textit{p}})$, which we used during denoising process to generate \textbf{\textit{A}}, \textbf{\textit{X}} and \textbf{\textit{L}}. As mentioned in \ref{prelim_inv}, the learned distribution of material structure $p(\textbf{\textit{M}})$  must satisfy periodic E(3) invariance. Hence we leverage a periodic-E(3)-equivariant Graph Neural Network (GNN) integrated with a pre-trained textual encoder to model the denoising process. In particular, as a text encoder, we adopt a pre-trained MatSciBERT~\citep{gupta_matscibert_2022} model, which is a domain-specific language model for materials science, followed by a projection layer. MatSciBERT is effectively a pre-trained SciBERT model on a scientific text corpus of 3.17B words, which is further trained on a huge text corpus of materials science containing around 285M words.  We feed textual description of material $\textbf{\textit{T}}$ into MatSciBERT and extract embedding of [CLS] token $\textbf{\textit{h}}_{CLS}$ as a representation of the whole text. Further. we feed $\textbf{\textit{h}}_{CLS}$ through a projection layer to generate the contextual textual embedding for the material $\textbf{\textit{C}}_\textbf{\textit{p}} \in \mathbb{R}^{d}$, which we pass to the equivariant GNN model to guide the denoising process. Practically, as the backbone network for the denoising process, we extend CSPNet architecture~\citep{jiao2023crystal}, originally developed for crystal structure prediction (CSP) task. CSPNet is built upon EGNN~\citep{satorras2021n}, satisfying periodic E(3) invariance condition on periodic crystal structure. At the $k^{th}$ layer message passing, the Equivariant Graph Convolutional Layer (EGCL) takes as input the set of atom embeddings $\textbf{\textit{h}}^{k}=[\textbf{\textit{h}}^{k}_1,\textbf{\textit{h}}^{k}_2,...,\textbf{\textit{h}}^{k}_N]$, atom coordinates $\textbf{\textit{x}}^{k}=[\textbf{\textit{x}}^{k}_1,\textbf{\textit{x}}^{k}_2,...,\textbf{\textit{x}}^{k}_N]$ and Lattice Matrix \textbf{\textit{L}} and outputs a transformation on $\textbf{\textit{h}}^{k+1}$. Formally, we can define the $k^{th}$ layer message passing operation as:
\setlength{\belowdisplayskip}{1pt} 
\setlength{\belowdisplayshortskip}{1pt}
\setlength{\abovedisplayskip}{0pt} 
\setlength{\abovedisplayshortskip}{0pt}
\begin{equation}
\label{eq:msg_pass}
    \textbf{\textit{m}}_{i,j} = \rho_m \{\textbf{\textit{h}}^{k}_i, \ \textbf{\textit{h}}^{k}_j, \ \textbf{\textit{L}}^T\textbf{\textit{L}}, \ 
    \psi_{FT}(\textbf{\textit{x}}^{k}_i - \textbf{\textit{x}}^{k}_j ) \} ; \;
    \textbf{\textit{m}}_{i} = \sum_{j=1}^{N} \textbf{\textit{m}}_{i,j};  \;
    \textbf{\textit{h}}^{k+1}_{i} = \textbf{\textit{h}}^{k}_{i} + \rho_h \{ \textbf{\textit{h}}^{k}_{i}, \textbf{\textit{m}}_{i} \}
\end{equation}
% \begin{equation}
% \label{eq:msg_pass_2}
%     \textbf{\textit{h}}^{k+1}_{i} = \textbf{\textit{h}}^{k}_{i} + \rho_h \{ \textbf{\textit{h}}^{k}_{i}, \textbf{\textit{m}}_{i} \}
% \end{equation}
where $\rho_m, \rho_h$ are MLPs and $\psi_{FT}$ is a Fourier Transformation function applied on relative difference between fractional coordinates $\textbf{\textit{x}}^{k}_i$, $ \textbf{\textit{x}}^{k}_j$. Fourier Transformation is used since it is invariant to periodic translation and extracts various frequencies of all relative fractional distances that are helpful for crystal structure modeling~\citep{jiao2023crystal}. \\
We fuse textual representation $\textbf{\textit{C}}_\textbf{\textit{p}}$ into input atom feature $\textbf{\textit{h}}^{0}_{i}$ as 
% $\textbf{\textit{h}}^{0}_{i} = \rho \ \{ \ f_{atom}(\textbf{\textit{a}}_i) \ \lvert \rvert \  f_{pos}(t) \ \lvert \rvert  \ \textbf{\textit{C}}_\textbf{\textit{p}}  \}$,
\begin{equation}
\label{eq:fuse}
    \textbf{\textit{h}}^{0}_{i} = \rho \ \{ \ f_{atom}(\textbf{\textit{a}}_i) \ \lvert \rvert \  f_{pos}(t) \ \lvert \rvert  \ \textbf{\textit{C}}_\textbf{\textit{p}}  \}
\end{equation}
where t is the timestamp of the diffusion model, $f_{pos}(.)$ is sinusoidal
positional encoding~\citep{ho2020denoising,vaswani2017attention}, $f_{atom}(.)$ learned atomic embedding function and $\lvert \rvert$ is concatenation operation. Input atom features $\textbf{\textit{h}}^{0}$ and coordinates $\textbf{\textit{x}}^{0}$ are fed through $\mathcal{K}$ layers of EGCL to produce $\hat{\bm{\epsilon}}^{\textbf{\textit{L}}}$, $p(\textbf{\textit{A}}_{t-1}|\textbf{\textit{M}}_{t})$ and $\hat{\bm{\epsilon}}^{\textbf{\textit{X}}}$ as follows :
\begin{equation}
    \label{eq:msg_pass_3}
    % \begin{split}
        \hat{\bm{\epsilon}}^{\textbf{\textit{L}}} = \textbf{\textit{L}} \rho_L (\frac{1}{N} \sum^{N}_{i=1} \textbf{\textit{h}}^{\mathcal{K}}) ; \; 
        p(\textbf{\textit{A}}_{t-1} \ | \ \textbf{\textit{M}}_{t}) = \rho_A (\textbf{\textit{h}}^{\mathcal{K}}) ; \; 
        \hat{\bm{\epsilon}}^{\textbf{\textit{X}}} = \rho_X (\textbf{\textit{h}}^{\mathcal{K}})
    % \end{split}
\end{equation}
where $\rho_L, \rho_A, \rho_X$ are MLPs on the final layer embeddings. Intuitively, we feed global structural knowledge about the crystal structure into the network by injecting contextual representation $\textbf{\textit{C}}_\textbf{\textit{p}}$ into input atom features. This added signal participates through message-passing operations in Eq. \ref{eq:msg_pass} and guides in denoising atom types, coordinates, and lattice parameters such that it can capture the global crystal geometry and aligned with the input stable structure specified by textual description.
% \noteng{this section is a bit dense, making it difficult to read and looks like a potpourri of methods copied from various sources, try to read carefully and bring in the originality quotient}
\subsection{Training and Sampling}
\label{train_sample}
\our{} is trained using the following combined loss:
% $\mathcal{L} = \lambda_{L} \mathcal{L}_{lattice} + \lambda_{A} \mathcal{L}_{type} + \lambda_{X} \mathcal{L}_{coord}$
\begin{equation}
    \label{eq:loss}
        \mathcal{L} = \lambda_{L} \mathcal{L}_{lattice} + \lambda_{A} \mathcal{L}_{type} + \lambda_{X} \mathcal{L}_{coord}
\end{equation}
where $\mathcal{L}_{lattice}$, $\mathcal{L}_{type}$ and $\mathcal{L}_{coord}$ are lattice $l_2$ loss (Eq. \ref{eq:lattice_loss}), type loss (Eq. \ref{eq:type_loss}) and coordinate score matching loss (Eq. \ref{eq:coord_loss}) respectively and $\lambda_{L}$, $\lambda_{A}$, $\lambda_{X}$ are hyperparameters control the relative weightage between these different loss components. During training, we freeze the MatSciBERT parameters and do not tune them further. During sampling, we use the Predictor-Corrector~\citep{song2020score} sampling mechanism to sample $\textbf{\textit{A}}_0$, $\textbf{\textit{X}}_0$ and $\textbf{\textit{L}}_0$. Training/Sampling algorithms are provided in Appendix \ref{appendix_training}
% TABLE 1 ------------>
\begin{table}[t] 
\centering
\setlength{\tabcolsep}{8 pt}
\resizebox{1.0\textwidth}{!}{
\begin{tabular}{c | c | c c | c c| c c }
\toprule
\multirow{2}*{Method} &  \multirow{2}*{\# Samples} & \multicolumn{2}{c}{Perov-5} & \multicolumn{2}{c}{Carbon-24} & \multicolumn{2}{c}{MP-20}  \\
& & Match Rate $\uparrow$ & RMSE $\downarrow$ & Match Rate $\uparrow $ & RMSE $\downarrow$ & Match Rate $\uparrow $ & RMSE $\downarrow$ \\
\midrule
% % RS
% \multirow{2}*{RS} 
% & 20    & 29.22 & 0.2924 & 14.63 & 0.4041 & 8.73 & 0.2501 \\
% & 5,000 & 36.56 & 0.0886 & 14.63 & 0.4041 & 11.49 & 0.2822 \\
% \midrule
% % BO
% \multirow{2}*{BO} 
% & 20    & 21.03 & 0.2830 & 0.44  & 0.3653 & 8.11  & 0.2402  \\
% & 5,000 & 55.09 & 0.2037 & 12.17 & 0.4089 & 12.68 & 0.2816 \\
% \midrule
% % PSO
% \multirow{2}*{PSO} 
% & 20    & 20.90 & 0.0836 & 6.40 & 0.4204 & 4.05  & 0.1567\\
% & 5,000 & 21.88 & 0.0844 & 6.50 & 0.4211 & 4.35  & 0.1670\\
% \midrule
% \midrule
% P-cG-SchNet
\multirow{2}*{\shortstack{P-cG-SchNet}} 
& 1  & 48.22 & 0.4179 & 17.29 & 0.3846 & 15.39  & 0.3762\\
& 20 & 97.94 & 0.3463 & 55.91 & 0.3551 & 32.64  & 0.3018\\
\midrule
% CDVAE
\multirow{2}*{CDVAE} 
& 1  & 45.31 & 0.1138 & 17.09 & 0.2969 & 33.90 & 0.1045\\
& 20 & 88.51 & 0.0464 & 88.37 & 0.2286 & 66.95 & 0.1026\\
\midrule
% DiffCSP
\multirow{2}*{DiffCSP} 
& 1  & 52.02 & 0.0760 & 17.54 & 0.2759 & 51.49 & 0.0631 \\
& 20 & \textbf{98.60}   & \underline{0.0128} & 88.47 & 0.2192 & 77.93 & 0.0492 \\
\midrule
% Ours - Short
\multirow{2}*{\shortstack{\our{}\\(Short)}} 
& 1 & \underline{56.54} & \underline{0.0583} & \underline{24.13} & \underline{0.2424} & \underline{52.22} & \underline{0.0597} \\
& 20 & 98.25 & 0.0137 & 88.28 & 0.2252 & \underline{80.97} & \underline{0.0443}  \\
\midrule
% Ours - Long
\multirow{2}*{\shortstack{\our{}\\(Long)}} 
& 1 & \textbf{90.46} & \textbf{0.0203} & \textbf{44.63} & \textbf{0.2266} & \textbf{55.15} & \textbf{0.0572} \\
& 20 & \underline{98.59} & \textbf{0.0072} & \textbf{95.27} & \textbf{0.1534} & \textbf{82.02} & \textbf{0.0483} \\
\bottomrule
\end{tabular} 
}
% \vspace{-10pt}
\caption{Summary of results on \textit{CSP} task. We highlight the best and second-best performances in bold and underlined, respectively.
}
  \label{tbl-recons}
\end{table}
\section{Experiments}\label{results}
We provide a comprehensive evaluation of our method against several baselines on two benchmark tasks. 
First, in \ref{results_data_task}, we provide a brief overview of the experimental setup, including benchmark tasks, and datasets. Next, in \ref{results_csp}, we demonstrate how textual data enhances the prediction of stable crystal structures. Following that, in \ref{results_gen}, we highlight the effectiveness of our proposed joint diffusion paradigm in enhancing its ability to generate novel crystal materials. Additionally, we present the correctness of our generated materials \ref{result_corectness}, the computational cost of training and inference \ref{results_time}, and a few ablation studies \ref{results_ablation}. Appendix \ref{appendix_results} describes additional experiments and more ablations studies.
\subsection{Benchmark Tasks and Datasets}
\label{results_data_task} 
Following the prior works~\citep{xie2021crystal, jiao2023crystal}, we evaluate our proposed model \our{} on two benchmark tasks for material generation, \textit{Crystal Structure Prediction (CSP)} and \textit{Random Material Generation (Gen)}, using three popular material datasets: \textbf{Perov-5}~\citep{castelli2012new,castelli2012computational}, \textbf{Carbon-24}~\citep{carbon} and \textbf{MP-20}~\citep{jain2013materials}. We curated textual data for these datasets with a textual description of each material. Specifically, we generate both long detailed textual descriptions and shorter prompts using approaches mentioned in \ref{text_data}. While training \our{}, we split the datasets into the train, test, and validation sets following the convention of 60:20:20~\citep{xie2021crystal}.  More details about the dataset and the experimental setup are in Appendix \ref{appendix_results_expsetup}
\subsection{Crystal Structure Prediction (CSP)}
\label{results_csp}
\xhdr{Setup} In \textit{CSP} task, the goal is to predict the crystal structure (atom coordinates and lattice) given atom types. In text guidance setup, \our{} utilizes textual descriptions during the denoising process and jointly predicts atom coordinates and lattice parameters from randomly sampled noise. To assess \our{}'s effectiveness, we choose three SOTA generative models: \textbf{P-cG-SchNet}~\citep{gebauer2022inverse}, \textbf{CDVAE}~\citep{xie2021crystal}, and \textbf{DiffCSP}~\citep{jiao2023crystal}. Following prior works~\citep{jiao2023crystal,xie2021crystal}, we evaluate the performance of all competing models using standard metrics \textbf{Match Rate (MR)} and \textbf{RMSE}, by matching the generated structure and the ground truth structure in the test set. In particular, we generate k samples for each material structure in the test set, and determine the matching metrics (MR and RMSE) if at least one sample aligns with the ground truth structure. Details about evaluation metrics in Appendix \ref{appendix_results_metrics}\\\\
\xhdr{Results and Discussions} We report the Match Rate and RMSE of all the baselines and \our{} for three benchmark datasets in Table \ref{tbl-recons}.
We trained \our{} using both detailed textual descriptions and short prompts, as outlined in \ref{text_data} and report them as \textbf{\our{}(Long)} and \textbf{\our{}(Short)} respectively in Table \ref{tbl-recons}. We observe that both variants of \our{} surpass all the baseline models with a good margin across three datasets, which shows the rich capability of text-guided joint diffusion to predict stable crystal structure. Notably, while prior diffusion models demonstrate improved Match Rates and lower RMSE when generating 20 samples per test material, they largely fail in both metrics when generating only one sample per test material. However, generating 20 samples per test material to match the structures is unrealistic and computationally burdensome. This highlights the importance of text-guided diffusion: incorporating textual knowledge during reverse diffusion aids in aligning the noisy structure with the 3D geometry of stable realistic materials. Specifically, with just one generated sample ($k = 1$) per test material, both the variants of \our{} outperform all baseline models, thereby reducing computational overhead. Moreover, even with 20 generated samples ($k = 20$) performance of \our{} is significantly better for Carbon-24 and MP-20, whereas comparable with DiffCSP for Perov-5.  In general, due to the extensive metadata provided by the detailed description about the crystal structure, the performance enhancement in the long variant surpasses that of the short one. However, these findings collectively demonstrate the effectiveness of text-guided diffusion in the stable crystal structure prediction task.
% TABLE 2 ------------>
\begin{table}[t]
\centering
\footnotesize
\setlength{\tabcolsep}{7 pt}
\renewcommand{\arraystretch}{0.9}

% \scalebox{0.85}{
\resizebox{1.0\textwidth}{!}{
\begin{tabular}{c | c | c c | c c| c c c}
\toprule
\multirow{2}*{Dataset} &  \multirow{2}*{Method} & \multicolumn{2}{c|}{Validity $\uparrow$} & \multicolumn{2}{c|}{Coverage $\uparrow$} & \multicolumn{3}{c}{Property Statistics (EMD) $\downarrow$}  \\
& & Compositional & Structural & COV-R & COV-P  & \# Element  & $\rho$ & $\mathcal{E}$ \\
\midrule
\multirow{8}*{Perov-5} 
& CDVAE   & 98.59 & \textbf{100} & 99.45 & 98.46 & 0.0628 & 0.1258 & 0.0264 \\
& CDVAE+  & 98.45 & 99.8 & 99.53 & 99.09 & 0.0609 & 0.1276 & 0.0223 \\
& SyMat   & 97.40 & \textbf{100} & 99.68 & 98.64 & 0.0177 & 0.1893 & 0.2364 \\
& SyMat+  & 97.88 & \underline{99.9} & 99.70 & 98.79 & 0.0172 & 0.1755 & 0.2566 \\
& DiffCSP & \textbf{98.85} & \textbf{100} & \underline{99.74} & 98.27 & 0.0128 & 0.1110 & 0.0263 \\
& DiffCSP+ & 98.44 & \textbf{100} & 99.85 & 98.53 & 0.0119 & 0.1070 & \underline{0.0241} \\
& \our{}(Short) & 98.28 & \textbf{100} & 99.7 & \underline{99.24} & \underline{0.0108} & \underline{0.0947} & 0.0257 \\
& \our{}(Long) & \underline{98.63} & \textbf{100} & \textbf{99.83} & \textbf{99.52} & \textbf{0.0090} & \textbf{0.0497} & \textbf{0.0187} \\

\midrule
\multirow{8}*{Carbon-24} 
& CDVAE    & - & \textbf{100} & \underline{99.8} & 83.08 & - & 0.1407 & 0.285 \\
& CDVAE+   & - & \textbf{100} & \underline{99.8} & 84.76 & - & 0.1377 & 0.266 \\
& SyMat    & - & \textbf{100} & \textbf{99.9} & \underline{97.59} & - & 0.1195 & 3.9576 \\
& SyMat+   & - & \textbf{100} & \textbf{99.9} & \textbf{97.63} & - & 0.1171 & 3.862 \\
& DiffCSP  & - & \textbf{100} & \textbf{99.9} & 97.27 & - & 0.0805 & \underline{0.082} \\
& DiffCSP+ & - & \textbf{100} & \textbf{99.9} & 97.33 & - & 0.0763 & 0.085 \\
& \our{}(Short) & - & \textbf{100} & \underline{99.8} & 91.77 & - & \underline{0.0681} & 0.087 \\
& \our{}(Long) & - & \textbf{100} & \textbf{99.9} & 92.43 & - & \textbf{0.043} &  \textbf{0.063} \\

\midrule
\multirow{8}*{MP-20} 
& CDVAE  &  86.70 & \textbf{100} & 99.15 &  99.49 &  1.432  & 0.6875 & 0.2778 \\
& CDVAE+ & 87.42 & \textbf{100} & 99.57 & 99.81 & 0.972 & 0.6388 & 0.2977 \\
& SyMat  & 88.26 & \textbf{100} & 98.97 & \textbf{99.97} &0.5067  & 0.3805 & 0.3506 \\
& SyMat+  & \underline{88.47} & \underline{99.9} & 99.01 & \underline{99.95} & 0.4865 & 0.3879 & 0.3489 \\
& DiffCSP & 83.25 & \textbf{100} & 99.71 & 99.76 &  0.3398 & 0.3502 & 0.1247 \\
& DiffCSP+ & 85.07 & \textbf{100} & \underline{99.8} & 99.89 & \underline{0.3122} & 0.3799 & 0.1355 \\
& \our{}(Short) & 86.60 & \textbf{100} & 99.79 & 99.88 & 0.3337 & \textbf{0.3296} & \textbf{0.1154} \\
& \our{}(Long) & \textbf{92.97} & \textbf{100} & \textbf{99.89} & \underline{99.95} & \textbf{0.2890} & \underline{0.3382} & \underline{0.1189}  \\
\bottomrule
\end{tabular} 
}
% \vspace{-9pt}
\caption{Summary of results on \textit{Gen} task, with the best and second-best performances in bold and underlined, respectively. The table contains "-" values for metrics that don't apply to certain datasets.}
\label{tbl-gen}
\end{table}
\subsection{Random Material Generation (Gen)}\label{results_gen} 
\xhdr{Setup} In \textit{Gen} task, the goal is to generate novel stable materials (both structure and atom types), that are distributionally similar to the materials in the test dataset. In \our{}, by design choice, we use the textual description of crystal materials in the test dataset during the reverse diffusion process to enhance the generation capability. To evaluate performance of \our{} in this task, we choose three popular state-of-the-art generative models: \textbf{CDVAE}~\citep{xie2021crystal}, \textbf{SyMat}~\citep{luo2023towards}, and \textbf{DiffCSP}~\citep{jiao2023crystal}. For a fair comparison, we also consider their text-guided variants such as \textbf{CDVAE+}, \textbf{SyMat+} and \textbf{DiffCSP+} respectively, where we fuse the contextual representation of (Long) text data into those models using the same process described in \ref{text_guided_diffusion}. Following ~\citep{xie2021crystal}, for evaluating all the competing models, we use seven metrics under three broad categories: \textbf{Validity}, \textbf{Coverage}, and \textbf{Property Statistics} (Details in Appendix \ref{appendix_results_metrics}). To ensure a fair comparison regarding sample size, we generate a number of samples equal to the test data size for all baseline models, both unconditional and text-guided variants, and evaluate their performance accordingly.\\\\
\xhdr{Results and Discussions} We report the result of \textbf{\our{} (Long and Short)} and all the baseline models in Table \ref{tbl-gen}. We observe that both variants of \our{} consistently enhances performance across almost all metrics across the benchmark datasets. Particularly on the Perov-5 dataset, \our{} outperforms all baseline models across all metrics except for compositional validity, where its performance is on par with state-of-the-art results. In the Carbon-24 and MP-20 datasets, \our{} exhibits performance improvements across all metrics except for COV-P. Additionally, our experiments indicate that utilizing shorter prompts results in a slight decrease in overall performance compared to the longer variant. Nonetheless, the performance remains comparable to baseline models. Overall, \our{} exhibits promising performance in the Random Material Generation task, indicating its effectiveness in integrating global textual knowledge into the reverse diffusion process to generate more stable periodic structures of 3D crystal materials. Moreover \our{ } outperforms text guided variant of baseline models CDVAE+, SyMat+ and DiffCSP+ as well, which highlights the effectiveness of using joint diffusion to learn  \textbf{\textit{A}},\textbf{\textit{X}} and \textbf{\textit{L}}.\\ 
Additionally, we present visualizations of a few generated materials based on the textual descriptions in Table \ref{tbl-samples} and compare them with the ground truth material structure. The generated samples exhibit clear matches to the ground truth structure, highlighting the generation capability of the \our{} model given information in text form. More visualization results are in Appendix \ref{appendix-visualization}.
% TABLE 3 ------------>
\begin{table}[t]
  \centering
    \setlength{\tabcolsep}{1.5 pt}
    \renewcommand{\arraystretch}{0.9}
    % \resizebox{1.0\textwidth}{!}{
    \scalebox{0.7}{
    \begin{tabular}{m{2.5in}|m{2.2in}|c|cccc } % angstrom symbol Å
  \textbf{Long (Detailed) Description} & \textbf{Short Prompt} &  \textbf{Ground truth} & \multicolumn{4}{c}{\textbf{Generated Samples}} \\
 \midrule
 % perov
  % HgCsO3
  LaNi2Ge2 crystallizes in the tetragonal I4/mmm space group. La is bonded in a 16-coordinate geometry to eight equivalent Ni and eight equivalent Ge atoms. All La-Ni bond lengths are 3.25 Å. \dots The Ge-Ge bond length is 2.66 Å. The formation energy is -0.691. The band gap is 0.0. The energy above the convex hull is 0.0. & 
  Below is a description of a bulk material. The chemical formula is La(NiGe)2. The elements are La, Ni, Ge. The formation energy is negative. The band gap is zero. The energy above the convex hull is zero. The spacegroup number is 138. The crystal system is tetragonal. Generate the material.& \begin{minipage}{.1\linewidth}
      \includegraphics[width=\linewidth]{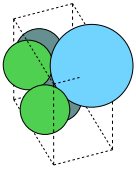}
  \end{minipage} & \begin{minipage}{.1\linewidth}
      \includegraphics[width=\linewidth]{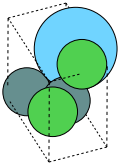}
  \end{minipage} & \begin{minipage}{.1\linewidth}
      \includegraphics[width=\linewidth]{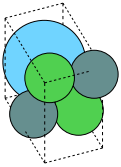}
  \end{minipage} & \begin{minipage}{.1\linewidth}
      \includegraphics[width=\linewidth]{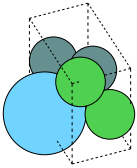}
  \end{minipage} & \begin{minipage}{.1\linewidth}
      \includegraphics[width=\linewidth]{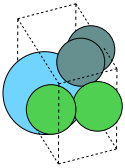}
  \end{minipage} \\
  \midrule
  % AuTaO2S
  HgScNOF is alpha Rhenium trioxide-derived structured and crystallizes in the orthorhombic Pmmm space group. The structure consists of one Hg cluster inside a ScNOF \dots linear geometry to two equivalent Sc atoms. The formation energy is 1.1428.&
  Below is a description of a bulk material. The chemical formula is HgScNOF. The elements are Sc,Hg,N,O,F. The formation energy is positive. The spacegroup number is 46. The crystal system is orthorhombic. Generate the material.& 
  \begin{minipage}{.1\linewidth}
      \includegraphics[width=\linewidth]{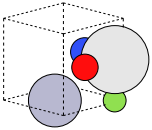}
  \end{minipage} & \begin{minipage}{.1\linewidth}
      \includegraphics[width=\linewidth]{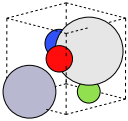}
  \end{minipage} & \begin{minipage}{.1\linewidth}
      \includegraphics[width=\linewidth]{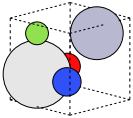}
  \end{minipage} & \begin{minipage}{.1\linewidth}
      \includegraphics[width=\linewidth]{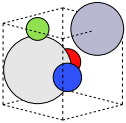}
  \end{minipage} & \begin{minipage}{.1\linewidth}
      \includegraphics[width=\linewidth]{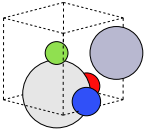}
  \end{minipage} 
  \\
  \bottomrule
 \end{tabular}
}
% \vspace{-10pt}
   \caption{Visualization of the generated structures given textual description. Note that our model produces rotated or translated versions of the ground truth material, owing to the periodic-E(3)invariance.}
  \label{tbl-samples}
  % \vspace*{-6pt}
\end{table}
\subsection{Correctness of Generated Materials}
\label{result_corectness}
\xhdr{Setup} In this section, we investigate whether the generated material matches different features specified by the textual prompts. \our{} has the capability to process textual prompts given by the user, enabling it to manage global attributes about crystal materials such as Formula, Space group, Crystal System, and different property values like formation energy, band-gap, etc. To ensure the fidelity of our model's outputs concerning these specified global attributes from the text prompt, We randomly generated 1000 materials (sampled from all three Datasets) based on their respective textual descriptions (both Long and Short) and assessed the percentage of generated materials that matched the global features outlined in the text prompt. In specific, we matched the Formula, Space group, and Crystal System of generated materials with the textual descriptions. Moreover, we examined whether properties such as formation energy and bandgap matched the specified criteria as per the text prompt (positive/negative, zero/nonzero). \\\\
\xhdr{Results and Discussions} We report the results for \our{} using long prompt in Table \ref{tbl-match} and short prompts in Table \ref{tbl-match-appendix} (\ref{result_corectness_appendix}). 
In general, using longer text, considering Perov-5 and Carbon-24 datasets, the generated material meets the specified criteria effectively. However, when dealing with the MP-20 dataset, which is more intricate due to its complex structure and composition, performance tends to decline. Additionally, when using shorter prompts, overall performance suffers across all datasets compared to longer text inputs. This is because the longer text, provided by the robocrystallographer, offers a comprehensive range of information, both global and local

\begin{table}
  \centering
  \small
    \setlength{\tabcolsep}{10 pt}
    \scalebox{1}{
    % \resizebox{1.0\textwidth}{!}{
\begin{tabular}{c | c | c c c }
\toprule
 & Global Features  & \multicolumn{3}{c}{\% of Matched Materials}  \\
& in Text Prompt   & Perov-5 & Carbon-24 & MP-20 \\
\midrule
% RS
\multirow{4}*{\our{}(Long)} 
& Formula           & 97.50 & 98.20 & 70.54\\
& Space Group       & 87.00 & 80.79 & 67.88\\
& Crystal System    & 92.60 & 91.55 & 73.54\\
& Formation Energy  & 95.49 & - & 92.88\\
& Band Gap          & - & 98.61 & 96.73\\
\bottomrule
\end{tabular} 
}
% \vspace{-8pt}
\caption{Correctness of generated materials matching conditions specified by the textual prompts.}
  \label{tbl-match}
\end{table}
\subsection{Computational Cost for Training and Sampling}\label{results_time}
Integrating textual knowledge during reverse diffusion for crystal material generation offers a key advantage: it accelerates convergence towards realistic structures and reduces computational overhead. We observe, compared to other baseline models, \our{} incurs substantially lower computation costs during training and sampling processes. Compared to baseline models, our approach notably cuts down on training time, requiring only 500 epochs compared to 3K or 4K epochs for CDVAE and DiffCSP on Perov-5 and Carbon-24 datasets respectively. Additionally, our method reduces sampling steps, making it faster to generate new structures. While CDVAE and DiffCSP need 5K and 1K steps respectively, our model only requires 500 steps. We compare the performance of  CDVAE and DiffCSP with different \our{}(Long) variants with 50, 100, 200, 500, and 1K steps and report the match rate of the predicted crystal structure vs running time (GPU hours in P100 GPU server) for Perov-5 and Carbon-24 datasets in Fig. \ref{fig:ablation}(a). We notice that the inference time for CDVAE is lengthier as it necessitates 5K steps for each generation. However, for Carbon-24, \our{} with 200 or 500 steps outperforms DiffCSP with 1K steps. Additionally, for Perov-5, \our{} with 500 steps achieves results comparable to DiffCSP with 1K steps.
\subsection{Smoothness of Model's Generation}
\label{results_ablation}
\begin{figure}
    \centering
	\subfloat[Match Rate vs Running time]{\includegraphics[width=0.45\columnwidth,height=3cm]{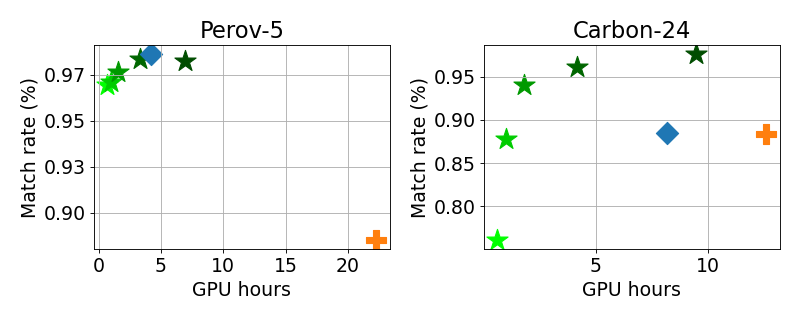}}
    \hspace{10mm}
    \subfloat[Smoothness of \our{}]{\boxed{\includegraphics[width=0.30\columnwidth]{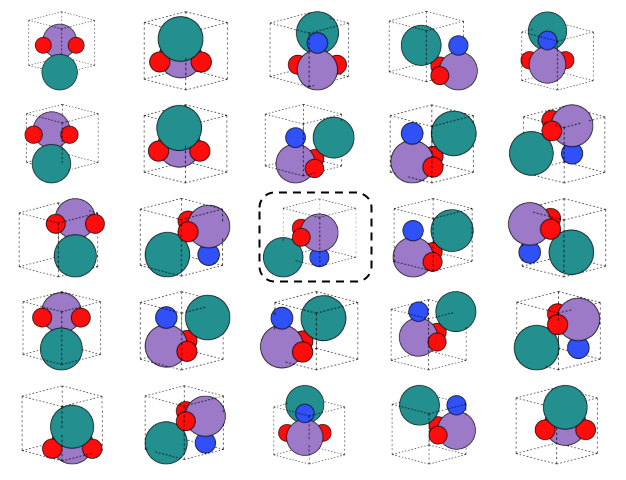}}}
    % \vspace{-8pt}
	\caption{(a) Match Rate vs Running time (GPU Hours) for different variants of \our{}(Long) \{50 Steps $\textcolor[rgb]{0,1,0,1}{\star}$, 100 Steps $\textcolor[rgb]{0,0.8,0,1}{\star}$, 200 Steps $\textcolor[rgb]{0,0.6,0,1}{\star}$, 500 Steps $\textcolor[rgb]{0,0.4,0,1}{\star}$, 1K Steps $\textcolor[rgb]{0,0.3,0,1}{\star}$ \}, DiffCSP $\textcolor[rgb]{0.12,0.46,0.70}{\blacklozenge}$ and CDVAE \textcolor[rgb]{1,0.498,0.054}{\textcolor[rgb]{1, 0.498, 0.054}{
        $\mathord{\text{\ding{58}}}$}}. (b) Materials sampled given the textual description of the center ground truth material $\boxed{\textbf{M}}$. The sampled materials are structurally similar (rotated or translated) to each other as well as the ground truth.}
	\label{fig:ablation}
\end{figure}

We qualitatively demonstrate the smoothness of the crystal generation process in our model. We provide a textual description of a ground truth material and generate several samples of materials to assess the diversity of the generated structures. Figure \ref{fig:ablation}(b) summarizes the results for one crystal material from the Perov-5 dataset, which shows that the generated materials are structurally similar to each other and the given input material.
\section{Conclusion}
\label{conclusion}
In this work, we explore a practical approach of generating stable crystal materials given a textual description of the material. We propose \our{}, which jointly diffuse atom types, fractional coordinates, and lattice structure for crystal materials using a periodic-E(3)-equivariant denoising model. We further integrate textual information into the reverse diffusion process through a pre-trained transformer model, which guides the denoising process in learning the crystal 3D geometry matching the specification by textual description. Extensive experiments conducted on two benchmark generative tasks reveal that \our{} surpasses all popular baseline models by a good margin. Furthermore, integrating textual knowledge reduces the overall computational cost for both training and inference of the diffusion model. Moreover, when applied to real-world custom text prompts by experts, \our{} demonstrates rich generative capability under general textual conditions.
\section{Acknowledgment}
\label{acknowledgment}
This work was funded by Indo Korea Science and Technology Center, Bangalore, India, under the project name “Generating Stable Periodic Materials using Diffusion Models”. We thank the Ministry of Education, Govt of India, for supporting Kishalay with Prime Minister Research Fellowship (PMRF) during his Ph.D. tenure.
\bibliography{iclr2025_conference}

\begin{thebibliography}{73}
\providecommand{\natexlab}[1]{#1}
\providecommand{\url}[1]{\texttt{#1}}
\expandafter\ifx\csname urlstyle\endcsname\relax
  \providecommand{\doi}[1]{doi: #1}\else
  \providecommand{\doi}{doi: \begingroup \urlstyle{rm}\Url}\fi

\bibitem[Austin et~al.(2021)Austin, Johnson, Ho, Tarlow, and Van Den~Berg]{austin2021structured}
Jacob Austin, Daniel~D Johnson, Jonathan Ho, Daniel Tarlow, and Rianne Van Den~Berg.
\newblock Structured denoising diffusion models in discrete state-spaces.
\newblock \emph{Advances in Neural Information Processing Systems}, 34:\penalty0 17981--17993, 2021.

\bibitem[Butler et~al.(2018)Butler, Davies, Cartwright, Isayev, and Walsh]{butler2018machine}
Keith~T Butler, Daniel~W Davies, Hugh Cartwright, Olexandr Isayev, and Aron Walsh.
\newblock Machine learning for molecular and materials science.
\newblock \emph{Nature}, 559\penalty0 (7715):\penalty0 547--555, 2018.

\bibitem[Campbell et~al.(2022)Campbell, Benton, De~Bortoli, Rainforth, Deligiannidis, and Doucet]{campbell2022continuous}
Andrew Campbell, Joe Benton, Valentin De~Bortoli, Thomas Rainforth, George Deligiannidis, and Arnaud Doucet.
\newblock A continuous time framework for discrete denoising models.
\newblock \emph{Advances in Neural Information Processing Systems}, 35:\penalty0 28266--28279, 2022.

\bibitem[Castelli et~al.(2012{\natexlab{a}})Castelli, Landis, Thygesen, Dahl, Chorkendorff, Jaramillo, and Jacobsen]{castelli2012new}
Ivano~E Castelli, David~D Landis, Kristian~S Thygesen, S{\o}ren Dahl, Ib~Chorkendorff, Thomas~F Jaramillo, and Karsten~W Jacobsen.
\newblock New cubic perovskites for one-and two-photon water splitting using the computational materials repository.
\newblock \emph{Energy \& Environmental Science}, 5\penalty0 (10):\penalty0 9034--9043, 2012{\natexlab{a}}.

\bibitem[Castelli et~al.(2012{\natexlab{b}})Castelli, Olsen, Datta, Landis, Dahl, Thygesen, and Jacobsen]{castelli2012computational}
Ivano~E Castelli, Thomas Olsen, Soumendu Datta, David~D Landis, S{\o}ren Dahl, Kristian~S Thygesen, and Karsten~W Jacobsen.
\newblock Computational screening of perovskite metal oxides for optimal solar light capture.
\newblock \emph{Energy \& Environmental Science}, 5\penalty0 (2):\penalty0 5814--5819, 2012{\natexlab{b}}.

\bibitem[Chen et~al.(2019)Chen, Ye, Zuo, Zheng, and Ong]{chen2019graph}
Chi Chen, Weike Ye, Yunxing Zuo, Chen Zheng, and Shyue~Ping Ong.
\newblock Graph networks as a universal machine learning framework for molecules and crystals.
\newblock \emph{Chem. Mater.}, 31\penalty0 (9):\penalty0 3564--3572, 2019.

\bibitem[Choudhary \& DeCost(2021)Choudhary and DeCost]{choudhary2021atomistic}
Kamal Choudhary and Brian DeCost.
\newblock Atomistic line graph neural network for improved materials property predictions.
\newblock \emph{npj Computational Materials}, 7\penalty0 (1):\penalty0 1--8, 2021.

\bibitem[Court et~al.(2020)Court, Yildirim, Jain, and Cole]{court20203}
Callum~J Court, Batuhan Yildirim, Apoorv Jain, and Jacqueline~M Cole.
\newblock 3-d inorganic crystal structure generation and property prediction via representation learning.
\newblock \emph{Journal of Chemical Information and Modeling}, 60\penalty0 (10):\penalty0 4518--4535, 2020.

\bibitem[Das et~al.(2022)Das, Samanta, Goyal, Lee, Bhattacharjee, and Ganguly]{das2022crysxpp}
Kishalay Das, Bidisha Samanta, Pawan Goyal, Seung-Cheol Lee, Satadeep Bhattacharjee, and Niloy Ganguly.
\newblock Crysxpp: An explainable property predictor for crystalline materials.
\newblock \emph{npj Computational Materials}, 8\penalty0 (1):\penalty0 43, 2022.

\bibitem[Das et~al.(2023{\natexlab{a}})Das, Goyal, Lee, Bhattacharjee, and Ganguly]{das2023crysmmnet}
Kishalay Das, Pawan Goyal, Seung-Cheol Lee, Satadeep Bhattacharjee, and Niloy Ganguly.
\newblock Crysmmnet: multimodal representation for crystal property prediction.
\newblock In \emph{Uncertainty in Artificial Intelligence}, pp.\  507--517. PMLR, 2023{\natexlab{a}}.

\bibitem[Das et~al.(2023{\natexlab{b}})Das, Samanta, Goyal, Lee, Bhattacharjee, and Ganguly]{das2023crysgnn}
Kishalay Das, Bidisha Samanta, Pawan Goyal, Seung-Cheol Lee, Satadeep Bhattacharjee, and Niloy Ganguly.
\newblock Crysgnn: Distilling pre-trained knowledge to enhance property prediction for crystalline materials.
\newblock \emph{arXiv preprint arXiv:2301.05852}, 2023{\natexlab{b}}.

\bibitem[Davies et~al.(2019)Davies, Butler, Jackson, Skelton, Morita, and Walsh]{davies2019smact}
Daniel~W Davies, Keith~T Butler, Adam~J Jackson, Jonathan~M Skelton, Kazuki Morita, and Aron Walsh.
\newblock Smact: Semiconducting materials by analogy and chemical theory.
\newblock \emph{Journal of Open Source Software}, 4\penalty0 (38):\penalty0 1361, 2019.

\bibitem[De~Bortoli et~al.(2022)De~Bortoli, Mathieu, Hutchinson, Thornton, Teh, and Doucet]{de2022riemannian}
Valentin De~Bortoli, Emile Mathieu, Michael Hutchinson, James Thornton, Yee~Whye Teh, and Arnaud Doucet.
\newblock Riemannian score-based generative modelling.
\newblock \emph{Advances in Neural Information Processing Systems}, 35:\penalty0 2406--2422, 2022.

\bibitem[Desiraju(2002)]{desiraju2002cryptic}
Gautam~R Desiraju.
\newblock Cryptic crystallography.
\newblock \emph{Nature materials}, 1\penalty0 (2):\penalty0 77--79, 2002.

\bibitem[Devlin et~al.(2018)Devlin, Chang, Lee, and Toutanova]{devlin2018bert}
Jacob Devlin, Ming-Wei Chang, Kenton Lee, and Kristina Toutanova.
\newblock Bert: Pre-training of deep bidirectional transformers for language understanding.
\newblock \emph{arXiv preprint arXiv:1810.04805}, 2018.

\bibitem[Dhariwal \& Nichol(2021)Dhariwal and Nichol]{dhariwal2021diffusion}
Prafulla Dhariwal and Alexander Nichol.
\newblock Diffusion models beat gans on image synthesis.
\newblock \emph{Advances in neural information processing systems}, 34:\penalty0 8780--8794, 2021.

\bibitem[Dresselhaus et~al.(2007)Dresselhaus, Dresselhaus, and Jorio]{dresselhaus2007group}
Mildred~S Dresselhaus, Gene Dresselhaus, and Ado Jorio.
\newblock \emph{Group theory: application to the physics of condensed matter}.
\newblock Springer Science \& Business Media, 2007.

\bibitem[Du et~al.(2024)Du, Yang, Dai, Dai, Nachum, Tenenbaum, Schuurmans, and Abbeel]{du2024learning}
Yilun Du, Sherry Yang, Bo~Dai, Hanjun Dai, Ofir Nachum, Josh Tenenbaum, Dale Schuurmans, and Pieter Abbeel.
\newblock Learning universal policies via text-guided video generation.
\newblock \emph{Advances in Neural Information Processing Systems}, 36, 2024.

\bibitem[Ganea et~al.(2021)Ganea, Pattanaik, Coley, Barzilay, Jensen, Green, and Jaakkola]{ganea2021geomol}
Octavian Ganea, Lagnajit Pattanaik, Connor Coley, Regina Barzilay, Klavs Jensen, William Green, and Tommi Jaakkola.
\newblock Geomol: Torsional geometric generation of molecular 3d conformer ensembles.
\newblock \emph{Advances in Neural Information Processing Systems}, 34:\penalty0 13757--13769, 2021.

\bibitem[Ganose \& Jain(2019)Ganose and Jain]{ganose2019robocrystallographer}
Alex~M Ganose and Anubhav Jain.
\newblock Robocrystallographer: automated crystal structure text descriptions and analysis.
\newblock \emph{MRS Communications}, 9\penalty0 (3):\penalty0 874--881, 2019.

\bibitem[Gebauer et~al.(2022)Gebauer, Gastegger, Hessmann, M{\"u}ller, and Sch{\"u}tt]{gebauer2022inverse}
Niklas~WA Gebauer, Michael Gastegger, Stefaan~SP Hessmann, Klaus-Robert M{\"u}ller, and Kristof~T Sch{\"u}tt.
\newblock Inverse design of 3d molecular structures with conditional generative neural networks.
\newblock \emph{Nature communications}, 13\penalty0 (1):\penalty0 973, 2022.

\bibitem[Gong et~al.(2024)Gong, Liu, Wu, and Wang]{gong2024text}
Haisong Gong, Qiang Liu, Shu Wu, and Liang Wang.
\newblock Text-guided molecule generation with diffusion language model.
\newblock \emph{arXiv preprint arXiv:2402.13040}, 2024.

\bibitem[Goodfellow et~al.(2014)Goodfellow, Pouget-Abadie, Mirza, Xu, Warde-Farley, Ozair, Courville, and Bengio]{goodfellow2014generative}
Ian Goodfellow, Jean Pouget-Abadie, Mehdi Mirza, Bing Xu, David Warde-Farley, Sherjil Ozair, Aaron Courville, and Yoshua Bengio.
\newblock Generative adversarial nets.
\newblock \emph{Advances in neural information processing systems}, 27, 2014.

\bibitem[Gruver et~al.(2024)Gruver, Sriram, Madotto, Wilson, Zitnick, and Ulissi]{gruver2024fine}
Nate Gruver, Anuroop Sriram, Andrea Madotto, Andrew~Gordon Wilson, C~Lawrence Zitnick, and Zachary Ulissi.
\newblock Fine-tuned language models generate stable inorganic materials as text.
\newblock \emph{arXiv preprint arXiv:2402.04379}, 2024.

\bibitem[Gupta et~al.(2022)Gupta, Zaki, Krishnan, and Mausam]{gupta_matscibert_2022}
Tanishq Gupta, Mohd Zaki, N.~M.~Anoop Krishnan, and Mausam.
\newblock {MatSciBERT}: A materials domain language model for text mining and information extraction.
\newblock \emph{npj Computational Materials}, 8\penalty0 (1):\penalty0 102, May 2022.
\newblock ISSN 2057-3960.
\newblock \doi{10.1038/s41524-022-00784-w}.
\newblock URL \url{https://www.nature.com/articles/s41524-022-00784-w}.

\bibitem[Ho \& Salimans(2022)Ho and Salimans]{ho2022classifierfree}
Jonathan Ho and Tim Salimans.
\newblock Classifier-free diffusion guidance, 2022.

\bibitem[Ho et~al.(2020)Ho, Jain, and Abbeel]{ho2020denoising}
Jonathan Ho, Ajay Jain, and Pieter Abbeel.
\newblock Denoising diffusion probabilistic models.
\newblock \emph{Advances in neural information processing systems}, 33:\penalty0 6840--6851, 2020.

\bibitem[Hoffmann et~al.(2019)Hoffmann, Maestrati, Sawada, Tang, Sellier, and Bengio]{hoffmann2019data}
Jordan Hoffmann, Louis Maestrati, Yoshihide Sawada, Jian Tang, Jean~Michel Sellier, and Yoshua Bengio.
\newblock Data-driven approach to encoding and decoding 3-d crystal structures.
\newblock \emph{arXiv preprint arXiv:1909.00949}, 2019.

\bibitem[Hoogeboom et~al.(2021)Hoogeboom, Nielsen, Jaini, Forr{\'e}, and Welling]{hoogeboom2021argmax}
Emiel Hoogeboom, Didrik Nielsen, Priyank Jaini, Patrick Forr{\'e}, and Max Welling.
\newblock Argmax flows and multinomial diffusion: Learning categorical distributions.
\newblock \emph{Advances in Neural Information Processing Systems}, 34:\penalty0 12454--12465, 2021.

\bibitem[Jain et~al.(2013{\natexlab{a}})Jain, Ong, Hautier, Chen, Richards, Dacek, Cholia, Gunter, Skinner, Ceder, and Persson]{10.1063/1.4812323}
Anubhav Jain, Shyue~Ping Ong, Geoffroy Hautier, Wei Chen, William~Davidson Richards, Stephen Dacek, Shreyas Cholia, Dan Gunter, David Skinner, Gerbrand Ceder, and Kristin~A. Persson.
\newblock {Commentary: The Materials Project: A materials genome approach to accelerating materials innovation}.
\newblock \emph{APL Materials}, 1\penalty0 (1):\penalty0 011002, 07 2013{\natexlab{a}}.
\newblock ISSN 2166-532X.
\newblock \doi{10.1063/1.4812323}.
\newblock URL \url{https://doi.org/10.1063/1.4812323}.

\bibitem[Jain et~al.(2013{\natexlab{b}})Jain, Ong, Hautier, Chen, Richards, Dacek, Cholia, Gunter, Skinner, Ceder, et~al.]{jain2013materials}
Anubhav Jain, Shyue~Ping Ong, Geoffroy Hautier, Wei Chen, William~Davidson Richards, Stephen Dacek, Shreyas Cholia, Dan Gunter, David Skinner, Gerbrand Ceder, et~al.
\newblock The materials project: A materials genome approach to accelerating materials innovation, apl mater.
\newblock \emph{APL Mater.}, 2013{\natexlab{b}}.

\bibitem[Jiao et~al.(2023)Jiao, Huang, Lin, Han, Chen, Lu, and Liu]{jiao2023crystal}
Rui Jiao, Wenbing Huang, Peijia Lin, Jiaqi Han, Pin Chen, Yutong Lu, and Yang Liu.
\newblock Crystal structure prediction by joint equivariant diffusion.
\newblock \emph{arXiv preprint arXiv:2309.04475}, 2023.

\bibitem[Jiao et~al.(2024)Jiao, Huang, Liu, Zhao, and Liu]{jiao2024space}
Rui Jiao, Wenbing Huang, Yu~Liu, Deli Zhao, and Yang Liu.
\newblock Space group constrained crystal generation.
\newblock \emph{arXiv preprint arXiv:2402.03992}, 2024.

\bibitem[Kim et~al.(2020)Kim, Noh, Gu, Aspuru-Guzik, and Jung]{kim2020generative}
Sungwon Kim, Juhwan Noh, Geun~Ho Gu, Alan Aspuru-Guzik, and Yousung Jung.
\newblock Generative adversarial networks for crystal structure prediction.
\newblock \emph{ACS central science}, 6\penalty0 (8):\penalty0 1412--1420, 2020.

\bibitem[Kingma \& Welling(2013)Kingma and Welling]{kingma2013auto}
Diederik~P Kingma and Max Welling.
\newblock Auto-encoding variational bayes.
\newblock \emph{arXiv preprint arXiv:1312.6114}, 2013.

\bibitem[Kohn \& Sham(1965)Kohn and Sham]{kohn1965self}
Walter Kohn and Lu~Jeu Sham.
\newblock Self-consistent equations including exchange and correlation effects.
\newblock \emph{Physical review}, 140\penalty0 (4A):\penalty0 A1133, 1965.

\bibitem[Kreuk et~al.(2022)Kreuk, Synnaeve, Polyak, Singer, D{\'e}fossez, Copet, Parikh, Taigman, and Adi]{kreuk2022audiogen}
Felix Kreuk, Gabriel Synnaeve, Adam Polyak, Uriel Singer, Alexandre D{\'e}fossez, Jade Copet, Devi Parikh, Yaniv Taigman, and Yossi Adi.
\newblock Audiogen: Textually guided audio generation.
\newblock \emph{arXiv preprint arXiv:2209.15352}, 2022.

\bibitem[Liu et~al.(2021)Liu, Yan, Oztekin, and Ji]{liu2021graphebm}
Meng Liu, Keqiang Yan, Bora Oztekin, and Shuiwang Ji.
\newblock Graphebm: Molecular graph generation with energy-based models.
\newblock \emph{arXiv preprint arXiv:2102.00546}, 2021.

\bibitem[Long et~al.(2021)Long, Fortunato, Opahle, Zhang, Samathrakis, Shen, Gutfleisch, and Zhang]{long2021constrained}
Teng Long, Nuno~M Fortunato, Ingo Opahle, Yixuan Zhang, Ilias Samathrakis, Chen Shen, Oliver Gutfleisch, and Hongbin Zhang.
\newblock Constrained crystals deep convolutional generative adversarial network for the inverse design of crystal structures.
\newblock \emph{npj Computational Materials}, 7\penalty0 (1):\penalty0 66, 2021.

\bibitem[Louis et~al.(2020)Louis, Zhao, Nasiri, Wang, Song, Liu, and Hu]{louis2020graph}
Steph-Yves Louis, Yong Zhao, Alireza Nasiri, Xiran Wang, Yuqi Song, Fei Liu, and Jianjun Hu.
\newblock Graph convolutional neural networks with global attention for improved materials property prediction.
\newblock \emph{Physical Chemistry Chemical Physics}, 22\penalty0 (32):\penalty0 18141--18148, 2020.

\bibitem[Luo et~al.(2022)Luo, Su, Peng, Wang, Peng, and Ma]{luo2022antigen}
Shitong Luo, Yufeng Su, Xingang Peng, Sheng Wang, Jian Peng, and Jianzhu Ma.
\newblock Antigen-specific antibody design and optimization with diffusion-based generative models for protein structures.
\newblock \emph{Advances in Neural Information Processing Systems}, 35:\penalty0 9754--9767, 2022.

\bibitem[Luo et~al.(2023{\natexlab{a}})Luo, Li, Liu, Wu, Yang, He, Wang, and Tian]{luo2023text}
Yanchen Luo, Sihang Li, Zhiyuan Liu, Jiancan Wu, Zhengyi Yang, Xiangnan He, Xiang Wang, and Qi~Tian.
\newblock Text-guided diffusion model for 3d molecule generation.
\newblock \emph{https://openreview.net/pdf?id=FdUloEgBSE}, 2023{\natexlab{a}}.

\bibitem[Luo et~al.(2023{\natexlab{b}})Luo, Liu, and Ji]{luo2023towards}
Youzhi Luo, Chengkai Liu, and Shuiwang Ji.
\newblock Towards symmetry-aware generation of periodic materials.
\newblock In \emph{Thirty-seventh Conference on Neural Information Processing Systems}, 2023{\natexlab{b}}.
\newblock URL \url{https://openreview.net/forum?id=Jkc74vn1aZ}.

\bibitem[Miller et~al.(2024)Miller, Chen, Sriram, and Wood]{miller2024flowmm}
Benjamin~Kurt Miller, Ricky~TQ Chen, Anuroop Sriram, and Brandon~M Wood.
\newblock Flowmm: Generating materials with riemannian flow matching.
\newblock \emph{arXiv preprint arXiv:2406.04713}, 2024.

\bibitem[Nichol et~al.(2021)Nichol, Dhariwal, Ramesh, Shyam, Mishkin, McGrew, Sutskever, and Chen]{nichol2021glide}
Alex Nichol, Prafulla Dhariwal, Aditya Ramesh, Pranav Shyam, Pamela Mishkin, Bob McGrew, Ilya Sutskever, and Mark Chen.
\newblock Glide: Towards photorealistic image generation and editing with text-guided diffusion models.
\newblock \emph{arXiv preprint arXiv:2112.10741}, 2021.

\bibitem[Noh et~al.(2019)Noh, Kim, Stein, Sanchez-Lengeling, Gregoire, Aspuru-Guzik, and Jung]{noh2019inverse}
Juhwan Noh, Jaehoon Kim, Helge~S Stein, Benjamin Sanchez-Lengeling, John~M Gregoire, Alan Aspuru-Guzik, and Yousung Jung.
\newblock Inverse design of solid-state materials via a continuous representation.
\newblock \emph{Matter}, 1\penalty0 (5):\penalty0 1370--1384, 2019.

\bibitem[Ong et~al.(2013)Ong, Richards, Jain, Hautier, Kocher, Cholia, Gunter, Chevrier, Persson, and Ceder]{ong2013python}
Shyue~Ping Ong, William~Davidson Richards, Anubhav Jain, Geoffroy Hautier, Michael Kocher, Shreyas Cholia, Dan Gunter, Vincent~L Chevrier, Kristin~A Persson, and Gerbrand Ceder.
\newblock Python materials genomics (pymatgen): A robust, open-source python library for materials analysis.
\newblock \emph{Computational Materials Science}, 68:\penalty0 314--319, 2013.

\bibitem[Park \& Wolverton(2020)Park and Wolverton]{Wolverton2020}
Cheol~Woo Park and Chris Wolverton.
\newblock Developing an improved crystal graph convolutional neural network framework for accelerated materials discovery.
\newblock \emph{Physical Review Materials}, 4\penalty0 (6), Jun 2020.
\newblock ISSN 2475-9953.
\newblock \doi{10.1103/physrevmaterials.4.063801}.
\newblock URL \url{http://dx.doi.org/10.1103/PhysRevMaterials.4.063801}.

\bibitem[Pickard.(2020)]{carbon}
Chris~J. Pickard.
\newblock Airss data for carbon at 10gpa and the c+n+h+o system at 1gpa.
\newblock \emph{materialscloud:2020.0026/v1}, 2020.

\bibitem[Radford et~al.(2021)Radford, Kim, Hallacy, Ramesh, Goh, Agarwal, Sastry, Askell, Mishkin, Clark, Krueger, and Sutskever]{radford2021learning}
Alec Radford, Jong~Wook Kim, Chris Hallacy, Aditya Ramesh, Gabriel Goh, Sandhini Agarwal, Girish Sastry, Amanda Askell, Pamela Mishkin, Jack Clark, Gretchen Krueger, and Ilya Sutskever.
\newblock Learning transferable visual models from natural language supervision, 2021.

\bibitem[Ramesh et~al.(2022)Ramesh, Dhariwal, Nichol, Chu, and Chen]{ramesh2022hierarchical}
Aditya Ramesh, Prafulla Dhariwal, Alex Nichol, Casey Chu, and Mark Chen.
\newblock Hierarchical text-conditional image generation with clip latents.
\newblock \emph{arXiv preprint arXiv:2204.06125}, 1\penalty0 (2):\penalty0 3, 2022.

\bibitem[Ren et~al.(2020)Ren, Noh, Tian, Oviedo, Xing, Liang, Aberle, Liu, Li, Jayavelu, et~al.]{ren2020inverse}
Zekun Ren, Juhwan Noh, Siyu Tian, Felipe Oviedo, Guangzong Xing, Qiaohao Liang, Armin Aberle, Yi~Liu, Qianxiao Li, Senthilnath Jayavelu, et~al.
\newblock Inverse design of crystals using generalized invertible crystallographic representation.
\newblock \emph{arXiv preprint arXiv:2005.07609}, 3\penalty0 (6):\penalty0 7, 2020.

\bibitem[Rombach et~al.(2022)Rombach, Blattmann, Lorenz, Esser, and Ommer]{rombach2022high}
Robin Rombach, Andreas Blattmann, Dominik Lorenz, Patrick Esser, and Bj{\"o}rn Ommer.
\newblock High-resolution image synthesis with latent diffusion models.
\newblock In \emph{Proceedings of the IEEE/CVF conference on computer vision and pattern recognition}, pp.\  10684--10695, 2022.

\bibitem[Saharia et~al.(2022)Saharia, Chan, Saxena, Li, Whang, Denton, Ghasemipour, Gontijo~Lopes, Karagol~Ayan, Salimans, et~al.]{saharia2022photorealistic}
Chitwan Saharia, William Chan, Saurabh Saxena, Lala Li, Jay Whang, Emily~L Denton, Kamyar Ghasemipour, Raphael Gontijo~Lopes, Burcu Karagol~Ayan, Tim Salimans, et~al.
\newblock Photorealistic text-to-image diffusion models with deep language understanding.
\newblock \emph{Advances in neural information processing systems}, 35:\penalty0 36479--36494, 2022.

\bibitem[Satorras et~al.(2021)Satorras, Hoogeboom, and Welling]{satorras2021n}
V{\i}ctor~Garcia Satorras, Emiel Hoogeboom, and Max Welling.
\newblock E (n) equivariant graph neural networks.
\newblock In \emph{International conference on machine learning}, pp.\  9323--9332. PMLR, 2021.

\bibitem[Schmidt et~al.(2021)Schmidt, Pettersson, Verdozzi, Botti, and Marques]{schmidt2021crystal}
Jonathan Schmidt, Love Pettersson, Claudio Verdozzi, Silvana Botti, and Miguel~AL Marques.
\newblock Crystal graph attention networks for the prediction of stable materials.
\newblock \emph{Science Advances}, 7\penalty0 (49):\penalty0 eabi7948, 2021.

\bibitem[Shi et~al.(2021)Shi, Luo, Xu, and Tang]{shi2021learning}
Chence Shi, Shitong Luo, Minkai Xu, and Jian Tang.
\newblock Learning gradient fields for molecular conformation generation.
\newblock In \emph{International conference on machine learning}, pp.\  9558--9568. PMLR, 2021.

\bibitem[Sohl-Dickstein et~al.(2015)Sohl-Dickstein, Weiss, Maheswaranathan, and Ganguli]{sohl2015deep}
Jascha Sohl-Dickstein, Eric Weiss, Niru Maheswaranathan, and Surya Ganguli.
\newblock Deep unsupervised learning using nonequilibrium thermodynamics.
\newblock In \emph{International conference on machine learning}, pp.\  2256--2265. PMLR, 2015.

\bibitem[Song \& Ermon(2019)Song and Ermon]{song2019generative}
Yang Song and Stefano Ermon.
\newblock Generative modeling by estimating gradients of the data distribution.
\newblock \emph{Advances in neural information processing systems}, 32, 2019.

\bibitem[Song \& Ermon(2020)Song and Ermon]{song2020improved}
Yang Song and Stefano Ermon.
\newblock Improved techniques for training score-based generative models.
\newblock \emph{Advances in neural information processing systems}, 33:\penalty0 12438--12448, 2020.

\bibitem[Song et~al.(2020)Song, Sohl-Dickstein, Kingma, Kumar, Ermon, and Poole]{song2020score}
Yang Song, Jascha Sohl-Dickstein, Diederik~P Kingma, Abhishek Kumar, Stefano Ermon, and Ben Poole.
\newblock Score-based generative modeling through stochastic differential equations.
\newblock \emph{arXiv preprint arXiv:2011.13456}, 2020.

\bibitem[Vaswani et~al.(2017)Vaswani, Shazeer, Parmar, Uszkoreit, Jones, Gomez, Kaiser, and Polosukhin]{vaswani2017attention}
Ashish Vaswani, Noam Shazeer, Niki Parmar, Jakob Uszkoreit, Llion Jones, Aidan~N Gomez, {\L}ukasz Kaiser, and Illia Polosukhin.
\newblock Attention is all you need.
\newblock \emph{Advances in neural information processing systems}, 30, 2017.

\bibitem[Wu et~al.(2021)Wu, Shen, Lan, Bian, and Huang]{wu2021se}
Jiaxiang Wu, Tao Shen, Haidong Lan, Yatao Bian, and Junzhou Huang.
\newblock Se (3)-equivariant energy-based models for end-to-end protein folding.
\newblock \emph{bioRxiv}, pp.\  2021--06, 2021.

\bibitem[Xie \& Grossman(2018)Xie and Grossman]{xie2018crystal}
Tian Xie and Jeffrey~C Grossman.
\newblock Crystal graph convolutional neural networks for an accurate and interpretable prediction of material properties.
\newblock \emph{Phys. Rev. Lett.}, 120\penalty0 (14):\penalty0 145301, 2018.

\bibitem[Xie et~al.(2021)Xie, Fu, Ganea, Barzilay, and Jaakkola]{xie2021crystal}
Tian Xie, Xiang Fu, Octavian-Eugen Ganea, Regina Barzilay, and Tommi Jaakkola.
\newblock Crystal diffusion variational autoencoder for periodic material generation.
\newblock \emph{arXiv preprint arXiv:2110.06197}, 2021.

\bibitem[Xu et~al.(2022)Xu, Yu, Song, Shi, Ermon, and Tang]{xu2022geodiff}
Minkai Xu, Lantao Yu, Yang Song, Chence Shi, Stefano Ermon, and Jian Tang.
\newblock Geodiff: A geometric diffusion model for molecular conformation generation.
\newblock \emph{arXiv preprint arXiv:2203.02923}, 2022.

\bibitem[Yan et~al.(2022)Yan, Liu, Lin, and Ji]{yan2022periodic}
Keqiang Yan, Yi~Liu, Yuchao Lin, and Shuiwang Ji.
\newblock Periodic graph transformers for crystal material property prediction.
\newblock In \emph{The 36th Annual Conference on Neural Information Processing Systems}, 2022.

\bibitem[Yang et~al.(2022)Yang, Yu, Wang, Wang, Weng, Zou, and Yu]{yang2207discrete}
D~Yang, J~Yu, H~Wang, W~Wang, C~Weng, Y~Zou, and D~Diffsound Yu.
\newblock Discrete diffusion model for text-to-sound generation. arxiv 2022.
\newblock \emph{arXiv preprint arXiv:2207.09983}, 2022.

\bibitem[Yang et~al.(2023)Yang, Cho, Merchant, Abbeel, Schuurmans, Mordatch, and Cubuk]{yang2023scalable}
Mengjiao Yang, KwangHwan Cho, Amil Merchant, Pieter Abbeel, Dale Schuurmans, Igor Mordatch, and Ekin~Dogus Cubuk.
\newblock Scalable diffusion for materials generation.
\newblock \emph{arXiv preprint arXiv:2311.09235}, 2023.

\bibitem[Zee(2016)]{zee2016group}
Anthony Zee.
\newblock \emph{Group theory in a nutshell for physicists}, volume~17.
\newblock Princeton University Press, 2016.

\bibitem[Zeni et~al.(2023)Zeni, Pinsler, Z{\"u}gner, Fowler, Horton, Fu, Shysheya, Crabb{\'e}, Sun, Smith, et~al.]{zeni2023mattergen}
Claudio Zeni, Robert Pinsler, Daniel Z{\"u}gner, Andrew Fowler, Matthew Horton, Xiang Fu, Sasha Shysheya, Jonathan Crabb{\'e}, Lixin Sun, Jake Smith, et~al.
\newblock Mattergen: a generative model for inorganic materials design.
\newblock \emph{arXiv preprint arXiv:2312.03687}, 2023.

\bibitem[Zhao et~al.(2021)Zhao, Al-Fahdi, Hu, Siriwardane, Song, Nasiri, and Hu]{zhao2021high}
Yong Zhao, Mohammed Al-Fahdi, Ming Hu, Edirisuriya~MD Siriwardane, Yuqi Song, Alireza Nasiri, and Jianjun Hu.
\newblock High-throughput discovery of novel cubic crystal materials using deep generative neural networks.
\newblock \emph{Advanced Science}, 8\penalty0 (20):\penalty0 2100566, 2021.

\bibitem[Zhu et~al.(2024)Zhu, Xiao, and Honavar]{zhu20243mdiffusion}
Huaisheng Zhu, Teng Xiao, and Vasant~G Honavar.
\newblock 3m-diffusion: Latent multi-modal diffusion for text-guided generation of molecular graphs, 2024.

\end{thebibliography}
\bibliographystyle{iclr2025_conference}
\newpage
\appendix
\section*{Periodic Materials Generation using Text-Guided Joint Diffusion Model (Technical Appendix)}
\section{Limitations and Future Work}
\label{appendix_limitation}
1) One of the major limitations and scope of the future work of our proposed work is the lack of independent textual datasets for material generation tasks. In our experimental setup, our model relied on textual data extracted from existing datasets. Initially, we extracted text data from CIF files of materials in the test sets of Perov-5, Carbon-24, and MP-20, utilizing this data to evaluate our model and all baseline models. While the experimental results show promise, a more robust evaluation could have been achieved with an independent dataset containing only textual prompts. This would enable us to assess how effectively these models can generate the underlying 3D structure of materials through a text-guided diffusion process. Hence curating an independent textual dataset for material generation containing a diverse set of meta-information will be a future scope for research.\\\\
2) Given text prompts/descriptions, we generate contextual representation using a text encoder in \our{}, where we adopted a pre-trained MatSciBERT~\citep{gupta_matscibert_2022} model, which is a domain-specific language model for materials science. Also, while training \our{}, we freeze the MatSciBERT parameters and do not tune them further. Moreover, during sampling, the user must follow a specific format (Long/Short) to provide the text description of the target material. This setup limits the expressive power of the textual representation. We investigated the robustness of \our{} with much shorter prompts to sample from pre-trained \our{} model in \ref{sec-shorter-prompt}, but observed performance degradation across all the benchmark dataset on \textit{Gen} task. Hence, Exploring state-of-the-art LLMs and further fine-tuning them during training may create more powerful text conditional diffusion models and provide flexibility to process text prompts of different formats. However, that might create computational overhead as it will increase the number of parameters significantly. This
provides scope for further investigation and we keep it as scope for future work.

\section{More Related Work}
\label{appendix_related}
\subsection{Crystal Representation Learning}
In recent times, graph neural network (GNN) based approaches have emerged as a powerful model in learning robust representation of crystal materials, which enhance fast and accurate property prediction. CGCNN ~\citep{xie2018crystal} is the first proposed model, which represents a 3D crystal structure as an undirected weighted multi-edge graph and builds a graph convolution neural network directly on the graph. Following CGCNN, there are a lot of subsequent studies ~\citep{chen2019graph,choudhary2021atomistic,das2023crysmmnet,louis2020graph, Wolverton2020,schmidt2021crystal}, where authors proposed different variants of GNN architectures for effective crystal representation learning.   Recently, graph transformer-based architecture Matformer ~\citep{yan2022periodic} is proposed to learn the periodic graph representation of the material, which marginally improves the performance, however, is much faster than the prior SOTA model. Moreover, scarcity of labeled data makes these models difficult to train for all the properties, and recently, some key studies ~\citep{das2022crysxpp,das2023crysgnn} have shown promising results to mitigate this issue using transfer learning, pre-training, and knowledge distillation respectively.
\subsection{Diffusion Models}
The fundamental idea of the diffusion model, as initially proposed by ~\citep{sohl2015deep}, is to gradually corrupt data with diffusion noise and learn a neural model to recover back data from noise. Idea of diffusion further developed in two broad categories - 1) \textit{Score Matching Network} ~\citep{song2019generative,song2020improved} and 2) \textit{Denoising Diffusion Probabilistic Models (DDPM)} ~\citep{ho2020denoising}. In recent times diffusion models have emerged as a powerful new family of deep generative models, achieving remarkable performance records across numerous applications such as image synthesis ~\citep{dhariwal2021diffusion,ramesh2022hierarchical,rombach2022high}, molecular conformer generation ~\citep{shi2021learning,xu2022geodiff}, molecular graph generation ~\citep{liu2021graphebm}, protein folding ~\citep{luo2022antigen,wu2021se} etc. 
\subsection{Conditional Diffusion Models}
The initial DDPM model ~\citep{ho2020denoising} demonstrated unconditional diffusion models for image generation, where the output cannot be directed towards a desired characteristic or property. In guided diffusion models, the sampling process can be steered by a prompt, which can be a textual description of the desired output, reference image, or any other type of media.

In the field of image generation by diffusion models, Ramesh et al ~\citep{ramesh2022hierarchical} came up with a text-guided diffusion model called Dall-E2 which showed how textual prompt can be used to steer the sampling process. While training the model, both the image and its textual description are encoded and mapped together, and the encoding of the prompt is used to generate the image during sampling. Another way of guiding the diffusion process using a separate classifier model was shown by ~\citep{dhariwal2021diffusion}. They trained a classifier on the noised images and used the gradient of the classifier to guide the sampling process. In the classifier-free setting, ~\citep{ho2022classifierfree} trained two diffusion models, one guided and one unguided, and combined the resulting score estimated during sampling to get the desired outcome. OpenAI's CLIP ~\citep{radford2021learning} further improved the relevance of the generated image to the given prompt by scoring the correctness of the generated image given the textual prompt.

Similar efforts have been made in the field of molecular generative models. The shortcoming of SMILES-based autoregressive models were addressed by TGM-DLM ~\citep{gong2024text} by utilizing diffusion models. This necessitates a two step process, text-guided generation phase, where the SMILES representation is generated from Gaussian noise with the help of a textual description, and correction phase, where necessary rectification are made for the correctness of SMILES string format. This is one of the drawbacks of the SMILES string format, which was addressed by 3M-Diffusion ~\citep{zhu20243mdiffusion}, where they have generated molecular graphs from a given textual description.

\begin{table*}
  \centering
  % \begin{tabular}{ ccc }
  \scalebox{0.8}{\begin{tabularx}{\linewidth}{ c|X|X }
    \toprule
     & DiffCSP & TGDMat \\
    \midrule
    Tasks & Only CSP Task & Both CSP and Gen Tasks \\
    \midrule
    Diffusion on Atom Type & - & Discrete Diffusion (D3PM) \\
    \midrule
    Model Category & Unconditional; unable to specify the criteria required by the user & Conditional; able to specify the criteria required by the user (in Text Format) \\
    \midrule
    Text Guided Diffusion & No & Yes\\
    \bottomrule
  \end{tabularx}}
  \caption{Key Differences between TGDMat from DiffCSP}
  \label{tbl-tgdmat-diffcsp}
\end{table*}

\subsection{Crystal Material Generation}
In the past, there were limited efforts in creating novel periodic materials, with researchers concentrating on generating the atomic composition of periodic materials while largely neglecting the 3D structure. With the advancement of generative models, the majority of the research focuses on using popular generative models like VAEs or GANs to generate 3D periodic structures of materials, however, they either represent materials as three-dimensional voxel images ~\citep{court20203,hoffmann2019data,long2021constrained,noh2019inverse}  and generate images to depict material structures (atom types, coordinates, and lattices), or they directly encode material structures as embedding vectors ~\citep{kim2020generative,ren2020inverse,zhao2021high}. However, these models neither incorporate stability in the generated structure nor are invariant to any Euclidean and periodic transformations. 
In recent times equivariant diffusion models ~\citep{xie2021crystal,luo2023towards,jiao2023crystal,yang2023scalable,jiao2024space,miller2024flowmm} have become the leading method for generating stable crystal materials, thanks to their capability to utilize the physical symmetries of periodic material structures. In specific, state-of-the-art models like CDVAE~\citep{xie2021crystal} and SyMat~\citep{luo2023towards} integrate a variational autoencoder (VAE) and powerful score-based decder network, work directly with the atomic coordinates of the structures and uses an equivariant graph neural network to ensure euclidean and periodic invariance. However, both CDVAE and SyMat first predict the lattice parameters and atomic composition using the VAE model and subsequently update the coordinates using score based diffusion model. Moreover, given atomic composition, DiffCSP~\citep{jiao2023crystal} jointly optimizes the atom coordinates and lattice using a diffusion framework to predict the crystal structure with high precision.\\\\
\xhdr{Relations with Prior Methods} Among the existing models, DiffCSP~\citep{jiao2023crystal} comes close to our methodology, however our work differs from it in multiple ways.  DiffCSP primarily focuses only on the Crystal Structure Prediction (CSP) task and they didn't explore the Crystal Generation task, whereas \our{} focuses on both tasks. 
Moreover, unlike DiffCSP, \our{} can leverage the informative textual descriptions during the reverse diffusion process and can jointly learn lattices, atom types, and fractional coordinates from randomly sampled noise. This makes \our{} more flexible and robust in Crystal Generation and Structure Prediction tasks.

\section{Invariances in Crystal Structure}
\label{appendix_prelim_inv}
The basic idea of using generative models for crystal generation is to learn the underlying data distribution of material structure $p(\textbf{\textit{M}})$. Since crystal materials satisfy physical symmetry properties ~\citep{dresselhaus2007group,zee2016group}, one of the major challenges here is the learned distribution must satisfy periodic E(3) invariance i.e. invariance to permutation, translation, rotation, and periodic transformations.
\begin{itemize}
    \item \textbf{\textit{Permutation Invariance :}} If we permute the indices of constituent atoms it will not change the material. Formally, given any material $ \textbf{\textit{M}}=(\textbf{\textit{A}},\textbf{\textit{X}},\textbf{\textit{L}})$, using any permutation matrix $\mathbf{P}$ if we permute $\textbf{\textit{A}}$ and $\textbf{\textit{X}}$ as $\mathbf{P}(\textbf{\textit{A}})$ and $\mathbf{P}(\textbf{\textit{X}})$, then new material $\textbf{\textit{M}}_\textbf{\textit{P}}=(\mathbf{P}(\textbf{\textit{A}}),\mathbf{P}(\textbf{\textit{X}}),\textbf{\textit{L}})$ will remains unchanged. Hence the underlying distribution is also the same i.e $p( \textbf{\textit{M}}) = p(\textbf{\textit{M}}_\textbf{\textit{P}})$.
    
    \item \textbf{\textit{Translation Invariance :}} If we translate the atom coordinates by a random vector it will not change the structure of the material. Formally, given any material $ \textbf{\textit{M}}=(\textbf{\textit{A}},\textbf{\textit{X}},\textbf{\textit{L}})$, if we translate $\textbf{\textit{X}}$ by an arbitrary translation vector $\mathbf{u} \in \mathbb{R}^{3}$, new generated material $\textbf{\textit{M}}_\textbf{\textit{P}} =(\textbf{\textit{A}},\textbf{\textit{X}}+\mathbf{u}\mathbf{1}^T,\textbf{\textit{L}})$ will be the same as $ \mathbf{\textit{M}}$. Hence $p( \textbf{\textit{M}}) = p(\textbf{\textit{M}}_\textbf{\textit{T}})$ must satisfy.
    
    \item \textbf{\textit{Rotational Invariance :}} If we rotate the atom coordinates and lattice matrix, the material remains unchanged. Formally, using any orthogonal rotational matrix $\mathbf{Q} \in R^{3 \times 3}$ (satisfying $\mathbf{Q}^T\mathbf{Q}= \mathbf{I}$), if we rotate $\textbf{\textit{X}}$ and $\textbf{\textit{L}}$ of any material $ \textbf{\textit{M}}$ and generate new $\textbf{\textit{M}}_\textbf{\textit{R}}=(\textbf{\textit{A}},\textbf{\textit{QX}},\textbf{\textit{QL}})$, then actually different representations of the same material. Hence $p( \textbf{\textit{M}}) = p(\textbf{\textit{M}}_\textbf{\textit{R}})$ must satisfy.
    
    \item \textbf{\textit{Periodic Invariance :}} Finally, since the atoms in the unit cell can periodically repeat itself infinite times along the lattice vector, there can be many choices of unit cells and coordinate matrices representing the same material. Formally, given coordinates $\textbf{\textit{X}}$, after applying periodic transformation using random matrix $\textbf{\textit{K}} \in R^{n \times 3}$, new coordinates $\mathbf{X'}=\textbf{\textit{X}}+\textbf{\textit{K}}\textbf{\textit{L}}$ are periodically equivalent. Hence $\mathbf{\textit{M}}=(\textbf{\textit{A}},\textbf{\textit{X}},\textbf{\textit{L}})$ and $\textbf{M'}=(\textbf{\textit{A}},\mathbf{X'},\textbf{\textit{L}})$ are same material and $p( \textbf{\textit{M}}) = p(\textbf{M'})$ must hold.
\end{itemize}

\section{Textual Dataset}
\label{appendix_text_data}
\begin{figure}[ht]
	\centering
	\includegraphics[width=\columnwidth]{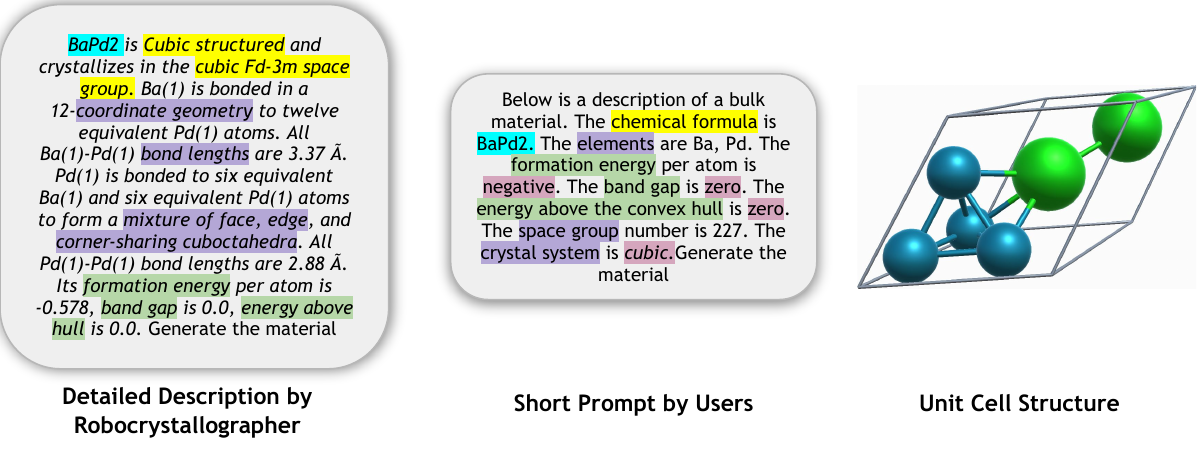}
 \caption{Detailed textual description generated by Robocrystallographer, short/less-detailed prompts by experts, and crystal unit cell structure of $\mathbf{BaPd_2}$ from Material Projects dataset. Text generated by Robocrystallographer contains both local chemical compositional information related to atom/bonds (like site coordination, geometry, polyhedral connectivity, and tilt angles) and global structural knowledge (like mineral type, space group information, symmetry, and dimensionality).The shorter prompt encodes minimal information about the material like its chemical formula, constituent elements, crystal system, and few chemical properties.}
	\label{fig:appendix_text_data}
\end{figure}

Leveraging textual information to guide the reverse diffusion process remains unexplored in the material design community. To the best of our knowledge, there is currently no dataset available that includes textual descriptions of the materials present in standard benchmark databases (Section \ref{results_data_task}) used for material generation. In specific, we propose two methods for generating textual descriptions of materials. Hence, we first curate the textual dataset containing textual descriptions of these materials to train our model. \\\\
\textbf{\textit{Long Detailed Textual Description:}} First, we utilize a freely available utility tool known as \textit{Robocrystallographer}~\citep{ganose2019robocrystallographer} to generate detailed textual descriptions about the periodic structure of crystal materials encoded in Crystallographic Information Files (CIF Files). Robocrystallographer breaks down crystal structures into two main components: local compositional details such as atomic coordination, geometry, polyhedral connectivity, and tilt angles, as well as global structural aspects like crystal formula, mineral type, space group information, symmetry, and dimensionality. This information is presented in three formats: JSON for machine processing, human-readable text for easy comprehension akin to descriptions provided by humans, and machine learning format for specialized analysis. We choose the human-readable text format to compile textual datasets, which closely resemble descriptions given of the crystal structure by humans. \\\\
\textbf{\textit{Short Custom Prompts:}} Secondly, we utilized shorter and less detailed prompts that are more easily interpretable by users. We extend the prompt template proposed by ~\citep{gruver2024fine}, which encodes minimal information about the material like its chemical formula, constituent elements, crystal system it belongs to, and its space group number. Further, we specify a few chemical properties, and instead of mentioning their actual values, we provide generic information like negative/positive formation energy, zero/nonzero band gaps, etc.  We used the Pymatgen tool~\citep{ong2013python} to extract this information from the Crystallographic Information Files (CIF Files) and curate the textual prompts.\\\\
An illustrative example of both these textual descriptions and the unit cell structure is provided in Figure \ref{fig:appendix_text_data}.

\section{Joint Equivariant Diffusion on \textbf{\textit{M}}}\label{appendix_diffusion}
Given an input crystal material $ \textbf{\textit{M}}_0= (\textbf{\textit{A}}_0,\textbf{\textit{X}}_0,\textbf{\textit{L}}_0)$, we define a forward diffusion process through a Markov chain over T steps to defuse  \textbf{\textit{A}}, \textbf{\textit{X}}, \textbf{\textit{L}} independently as follows :
\begin{equation}
    \label{eq:joint_dist}
    % \begin{split}
        q(\textbf{\textit{A}}_{t},\textbf{\textit{X}}_{t},\textbf{\textit{L}}_{t} | \textbf{\textit{A}}_{t-1},\textbf{\textit{X}}_{t-1},\textbf{\textit{L}}_{t-1}) = 
        q(\textbf{\textit{A}}_{t} | \textbf{\textit{A}}_{t-1}) q(\textbf{\textit{X}}_{t} | \textbf{\textit{X}}_{t-1}) 
        q(\textbf{\textit{L}}_{t} | \textbf{\textit{L}}_{t-1}) \: \ t=1,2,...T
    % \end{split}
\end{equation}
\subsection{Diffusion on Lattice (\textbf{\textit{L}})} Lattice Matrix $\textbf{\textit{L}}=[{l}_1,{l}_2,{l}_3]^T \in \mathbb{R}^{3 \times 3}$ is a global feature of the material which determines the shape and symmetry of the unit cell structure. Since \textbf{\textit{L}} is in continuous space, we leverage the idea of the Denoising Diffusion Probabilistic Model (DDPM) for diffusion on \textbf{\textit{L}}. In specific, given input lattice matrix $\textbf{\textit{L}}_0 \sim p(\textbf{\textit{L}})$, the forward diffusion process iteratively diffuses it over T timesteps to a noisy lattice matrix $\textbf{\textit{L}}_T$ 
through a transition probability $q(\textbf{\textit{L}}_{t} | \textbf{\textit{L}}_{0})$ at each $t^{th}$ step, which can be derived as follows :
\begin{equation}
    \label{eq:lattice_forward}
        q(\textbf{\textit{L}}_{t} \ | \ \textbf{\textit{L}}_{0}) = \mathcal{N} \bigg(\textbf{\textit{L}}_{t} \ | \ \sqrt{\bar{\alpha}_{t}}\textbf{\textit{L}}_{0}, \ (1 \ - \ \bar{\alpha}_{t})\mathbf{I} \bigg)
\end{equation}
where, $\bar{\alpha}_{t} = \prod_{k=1}^{t} \alpha_{k}$, $\alpha_{t} = 1-\beta_{t}$ and $\{ \beta_{t} \in (0,1) \}^{T}_{t=1}$ controls the variance of diffusion step following certain variance scheduler. 
By reparameterization, we can rewrite equation \ref{eq:lattice_forward} as:
\begin{equation}
    \label{eq:lattice_forward_reparam}
        \textbf{\textit{L}}_{t} = \sqrt{\bar{\alpha}_{t}}\textbf{\textit{L}}_{0} +  \sqrt{1-\bar{\alpha}_{t}}\bm{\epsilon}^{\textbf{\textit{L}}}
\end{equation}
where, $\bm{\epsilon}^{l}$ is a noise, sampled from $\mathcal{N}(\mathbf{0},\mathbf{I})$, added with original input sample $\textbf{\textit{L}}_{0}$  at $t^{th}$ step to generate $\textbf{\textit{L}}_{t}$. 
After T such diffusion steps, noisy lattice matrix $\textbf{\textit{L}}_T $ is generated from prior noise distribution $\sim \mathcal{N}(\mathbf{0},\mathbf{I})$. In the reverse denoising process, given noisy $\textbf{\textit{L}}_T \sim \mathcal{N}(\mathbf{0},\mathbf{I})$ we reconstruct true lattice structure $\textbf{\textit{L}}_0$ thorough iterative denoising step via learning reverse conditional distribution, which we formulate as follows :
\begin{equation}
    \label{eq:lattice_backward}
        p(\textbf{\textit{L}}_{t-1} | \textbf{\textit{M}}_{t},\textbf{\textit{C}}_\textbf{\textit{p}}) = \mathcal{N} \big \{ \textbf{\textit{L}}_{t-1} \ | \ \mu^{\textbf{\textit{L}}}(\textbf{\textit{M}}_{t},\textbf{\textit{C}}_\textbf{\textit{p}}),\beta_{t}\frac{(1 - \bar{\alpha}_{t-1})}{(1 -\bar{\alpha}_{t})}\mathbf{I} \big \}
\end{equation}
where $\mu^{\textbf{\textit{L}}}(\textbf{\textit{M}}_{t},\textbf{\textit{C}}_\textbf{\textit{p}}) = \frac{1}{\sqrt{\alpha_{t}}}\big(\textbf{\textit{L}}_{t}-\frac{1 - \alpha_{t}}{\sqrt{1 - \bar{\alpha}_{t}}} \ \hat{\bm{\epsilon}}^{\textbf{\textit{L}}} (\textbf{\textit{M}}_{t},\textbf{\textit{C}}_\textbf{\textit{p}},t)\big)$.
Intuitively, $\hat{\bm{\epsilon}}^{l}$ is the denoising term that needs to be subtracted from $\textbf{\textit{L}}_{t}$ to generate $\textbf{\textit{L}}_{t-1}$ and textual representation $\textbf{\textit{C}}_\textbf{\textit{p}}$ will steer this reverse diffusion process. We use a text-guided denoising network $\Phi_\theta(\textbf{\textit{A}}_{t},\textbf{\textit{X}}_{t},\textbf{\textit{L}}_{t},t,\textbf{\textit{C}}_\textbf{\textit{p}})$ to model the noise term $\hat{\bm{\epsilon}}^{\textbf{\textit{L}}} (\textbf{\textit{M}}_{t},\textbf{\textit{C}}_\textbf{\textit{p}},t)$. Following the simplified training objective proposed by ~\citep{ho2020denoising}, we train the aforementioned denoising network using $l_2$ loss between $\hat{\bm{\epsilon}}^{\textbf{\textit{L}}}$ and $\bm{\epsilon}^{\textbf{\textit{L}}}$
\begin{equation}
    \label{eq:appendix_lattice_loss}
        \mathcal{L}_{lattice} = \mathbb{E}_{\bm{\epsilon}^{\textbf{\textit{L}}},t \sim \mathcal{U}(1,T)}
        \lVert  \bm{\epsilon}^{\textbf{\textit{L}}} \ - \ \hat{\bm{\epsilon}}^{\textbf{\textit{L}}} {\rVert}^2_2 
\end{equation}
\subsection{Diffusion on Atom Types (\textbf{\textit{A}})} 
Prior studies ~\citep{jiao2023crystal,xie2021crystal} consider Atom Type Matrix $\textbf{\textit{A}}$ as the logits/probability distribution for k classes $\in$ $\mathbb{R}^{N \times k}$ (continuous variable in real space) and apply DDPM to learn the distribution. However for discrete data these models are inappropriate and produce suboptimal results ~\citep{austin2021structured,campbell2022continuous,hoogeboom2021argmax}. Hence we consider $\textbf{\textit{A}}$ as N discrete variables belonging to k classes and leverage discrete denoising diffusion probabilistic model (D3PM) ~\citep{austin2021structured} for diffusion on \textbf{\textit{A}}. In specific, denoting row vector \textbf{\textit{a}} as a one-hot representation of an atom \textit{a}, we can write transition probability for forward process as:
\begin{equation}
        q(\textbf{\textit{a}}_t | \textbf{\textit{a}}_{t-1}) = Cat(\textbf{\textit{a}}_t;\textbf{\textit{p}} = \textbf{\textit{a}}_{t-1}\textbf{\textit{Q}}_{t})
\end{equation}
where $Cat(\textbf{\textit{a}};\textbf{\textit{p}})$ is a categorical distribution over the one-hot row vector \textbf{\textit{a}} with probabilities given by the row vector \textbf{\textit{p}} and $\textbf{\textit{Q}}_{t}$ is the Markov transition matrix at time step t defined as $[\textbf{\textit{Q}}_{t}]_{i,j} = q(a_t = i \ | \ a_{t-1} = j)$. Different choices of $\textbf{\textit{Q}}_{t}$ and corresponding stationary distributions are proposed by ~\citep{austin2021structured} which provides flexibility to control the data corruption and denoising process. We adopted the absorbing state diffusion process, introducing a new absorbing state [MASK] in $\textbf{\textit{Q}}_{t}$. At each time step t, we can formally define the transition matrix as:
\begin{equation}
        [\textbf{\textit{Q}}_{t}]_{i,j} = 
        \begin{cases}
        1,           & \text{if $i=j=[MASK]$}.\\
        1-\beta_{t}, & \text{if $i=j \neq [MASK]$}\\
        \beta_{t}, & \text{if $i=j = [MASK]$}.
        \end{cases}
\end{equation}
Intuitively, at each time step t, an atom either stays in its type state with probability $1-\beta_{t}$ or moves to [MASK] state with probability $\beta_{t}$ and once it moves to [MASK] state, it stays in that state. Hence, the stationary distribution
of this diffusion process has all the mass on the [MASK] state. During reverse denoising process, given textual representation $\textbf{\textit{C}}_\textbf{\textit{p}}$, we first sample noisy $\textbf{\textit{a}}_T$ and obtain $\textbf{\textit{a}}_0$ thorough iterative denoising step via learning reverse conditional transition:
\begin{equation}
        p_{\theta}(\textbf{\textit{a}}_{t-1} | \textbf{\textit{a}}_t, \textbf{\textit{C}}_\textbf{\textit{p}}) \propto \sum_{\textbf{\textit{a}}_0} q(\textbf{\textit{a}}_{t-1}, \textbf{\textit{a}}_t | \textbf{\textit{a}}_0)p_{\theta}(\textbf{\textit{a}}_0 | \textbf{\textit{a}}_t, \textbf{\textit{C}}_\textbf{\textit{p}})
\end{equation}
We use the text-guided denoising network $\Phi_\theta(\textbf{\textit{A}}_{t},\textbf{\textit{X}}_{t},\textbf{\textit{L}}_{t},t,\textbf{\textit{C}}_\textbf{\textit{p}})$ to model this backward denoising process, which is trained using the following loss function as proposed by ~\citep{austin2021structured} :
\begin{equation}
    \label{eq:appendix_type_loss}
        \mathcal{L}_{type} = \mathcal{L}_{VB}+\lambda\mathcal{L}_{CE}
\end{equation}
where $\mathcal{L}_{VB}$ is the variational lower bound loss defined as follows:
\begin{equation}
    \label{eq:variational_loss}
    \begin{aligned}
        \mathcal{L}_{VB} = 
        \mathbb{E}_{\substack{q(\textbf{\textit{a}}_0)}} 
        \bigg[ 
        \underbrace{D_{KL} \{q(\textbf{\textit{a}}_{T}| \textbf{\textit{a}}_0) || p(\textbf{\textit{a}}_{T})\}}_{L_T} \ + \ 
        \sum^T_{t=2} \mathbb{E}_{\substack{q(\textbf{\textit{a}}_t | \textbf{\textit{a}}_0)}}
        \underbrace{[ D_{KL} \{ q(\textbf{\textit{a}}_{t-1}| \textbf{\textit{a}}_t,\textbf{\textit{a}}_0) || p_{\theta}(\textbf{\textit{a}}_{t-1}| \textbf{\textit{a}}_t)\}]}_{L_{t-1}} \\
        - \underbrace{\mathbb{E}_{\substack{q(\textbf{\textit{a}}_1 | \textbf{\textit{a}}_0)}}[\text{log} \ p_{\theta}(\textbf{\textit{a}}_0| \textbf{\textit{a}}_1)\}]}_{L_0}
        \bigg]
    \end{aligned}
\end{equation}
and $\mathcal{L}_{CE}$ is the cross-entropy loss defined as follows:
\begin{equation}
    \label{eq:ce_loss}
    \begin{aligned}
        \mathcal{L}_{CE} = 
        \mathbb{E}_{\substack{q(\textbf{\textit{a}}_0)}} 
        \bigg[
        \sum^T_{t=2} \mathbb{E}_{\substack{q(\textbf{\textit{a}}_t | \textbf{\textit{a}}_0)}}[\text{log} \ p_{\theta}(\textbf{\textit{a}}_0| \textbf{\textit{a}}_t)\}]
        \bigg]
    \end{aligned}
\end{equation}
and $\lambda$ is a hyperparameter.
\subsection{Diffusion on Atom Coordinates (\textbf{\textit{X}})}
Coordinate Matrix $\textbf{\textit{X}}=[\textbf{\textit{x}}_1,\textbf{\textit{x}}_2,...,\textbf{\textit{x}}_N]^T \in \mathbb{R}^{N \times 3}$ denotes atomic coordinate positions, where ${x}_i \in \mathbb{R}^{3}$ corresponds to coordinates of $i^{th}$ atom in the unit cell. We can diffuse the atom coordinates in two ways: either by diffusing cartesian coordinates or fractional coordinates. Prior works like CDVAE~\citep{xie2021crystal} and SyMat~\citep{luo2023towards} diffuse cartesian coordinates whereas DiffCSP~\citep{jiao2023crystal} diffuse fractional coordinates. In our setup, as we are jointly learning atom coordinates and lattice matrix simultaneously, we follow the line of work by DiffCSP and diffuse fractional coordinates. Atomic fractional coordinates in crystal material lives in quotient space $\mathbb{R}^{N \times 3} / \mathbb{Z}^{N \times 3}$ induced by the crystal periodicity. Since the Gaussian distribution used in DDPM is unable to model the cyclical and bounded domain of \textbf{\textit{X}}, it is not suitable to apply DDPM to model \textbf{\textit{X}}. Hence at each step of forward diffusion, we add noise sample from Wrapped Normal (WN) distribution ~\citep{de2022riemannian} to \textbf{\textit{X}} and during backward diffusion leverage Score Matching Diffusion Networks ~\citep{song2019generative,song2020improved} to model underlying transition probability $q(\textbf{\textit{X}}_{t} \ | \ \textbf{\textit{X}}_{0}) = \mathcal{N}_W (\textbf{\textit{X}}_{t} \ | \ \textbf{\textit{X}}_{0}, \sigma_t^2 \mathbf{I})$. In specific, at each $t^{th}$ step of diffusion, we derive $\textbf{\textit{X}}_{t}$ as : 
$\textbf{\textit{X}}_{t} = f_w(\textbf{\textit{X}}_{0} + \bm{\sigma_t}\bm{\epsilon}^{\textbf{\textit{X}}})$
where, $\bm{\epsilon}^{\textbf{\textit{X}}}$ is a noise, sampled from $\mathcal{N}(\mathbf{0},\mathbf{I})$, $\bm{\sigma_t}$ is the noise scale following exponential scheduler and $f_w(.)$ is a truncation function. Given a fractional coordinate matrix X, truncation function $f_w(\textbf{\textit{X}}) = (\textbf{\textit{X}} - \lfloor \textbf{\textit{X}} \rfloor)$ returns the fractional part of each element of \textbf{\textit{X}}. \\
As argued in ~\citep{jiao2023crystal}, $q(X_t|X_0)$ is periodic translation equivariant, and approaches uniform distribution $\mathcal{U}(0,1)$ for sufficiently large values of $\sigma_T$. Hence during the backward denoising process, we first sample $\textbf{\textit{X}}_{T} \sim \mathcal{U}(0,1)$ and iteratively denoise via score network for T steps to recover back the true fractional coordinates $\textbf{\textit{X}}_{0}$. We use the text-guided denoising network $\Phi_\theta(\textbf{\textit{A}}_{t},\textbf{\textit{X}}_{t},\textbf{\textit{L}}_{t},t,\textbf{\textit{C}}_\textbf{\textit{p}})$ to model the backward diffusion process, which is trained using the following score-matching objective function :
\begin{equation}
    \label{eq:appendix_coord_loss}
        \mathcal{L}_{coord} = \mathbb{E}_{\substack{\textbf{\textit{X}}_{t} \sim q(\textbf{\textit{X}}_{t} | \textbf{\textit{X}}_{0}) \\
        t \sim \mathcal{U}(1,T)}}
        \lVert  \nabla_{\textbf{\textit{X}}_{t}} \text{log} q(\textbf{\textit{X}}_{t} | \textbf{\textit{X}}_{0}) - \hat{\bm{\epsilon}}^{\textbf{\textit{X}}}(\textbf{\textit{M}}_{t},\textbf{\textit{C}}_\textbf{\textit{p}},t) {\rVert}^2_2
\end{equation}
where $\nabla_{\textbf{\textit{X}}_{t}} \text{log} q(\textbf{\textit{X}}_{t} | \textbf{\textit{X}}_{0}) \propto \sum_{\textbf{\textit{K}} \in \mathbb{Z}^{N \times 3}} \text{exp}(- \ \frac{\lVert \textbf{\textit{X}}_{t} - \textbf{\textit{X}}_{0} + \textbf{\textit{K}} {\rVert}^2_F }{2\bm{\sigma_t}^2})$ is the score function of transitional distribution and $\hat{\bm{\epsilon}}^{\textbf{\textit{X}}}(\textbf{\textit{M}}_{t},\textbf{\textit{C}}_\textbf{\textit{p}},t)$ denoising term.
\subsection{Text Guided Denoising Network} 
\label{appendix_text_guided_diffusion}
In this subsection, we will illustrate the detailed architecture of our proposed Text Guided Denoising Network $\Phi_\theta(\textbf{\textit{A}}_{t},\textbf{\textit{X}}_{t},\textbf{\textit{L}}_{t},t,\textbf{\textit{C}}_\textbf{\textit{p}})$, which we used to denoise \textbf{\textit{A}}, \textbf{\textit{X}} and \textbf{\textit{L}}. As mentioned in \ref{prelim_inv}, the learned distribution of material structure $p(\textbf{\textit{M}})$  must satisfy periodic E(3) invariance. Hence we leverage an periodic-E(3)-equivariant Graph Neural Network (GNN) integrated with a pre-trained textual encoder to model the denoising process. In particular, as a text encoder, we adopt a pre-trained MatSciBERT ~\citep{gupta_matscibert_2022} model, which is a domain-specific language model for materials science, followed by a projection layer. MatSciBERT is effectively a pre-trained SciBERT model on a scientific text corpus of 3.17B words, which is further trained on a huge text corpus of materials science containing around 285 M words.  We feed textual description of material $\mathcal{T}$ and extract embedding of [CLS] token $\textbf{\textit{h}}_{CLS}$ as a representation of the whole text. Further. we pass $\textbf{\textit{h}}_{CLS}$ through a projection layer to generate the contextual textual embedding for the material $\textbf{\textit{C}}_\textbf{\textit{p}} \in \mathbb{R}^{d}$, which we pass to the equivariant GNN model to guide the denoising process. Practically, as the backbone network for the backward diffusion process, we extend CSPNet architecture~\citep{jiao2023crystal}, originally developed for crystal structure prediction (CSP) task. CSPNet is built upon EGNN~\citep{satorras2021n}, satisfying periodic E(3) invariance condition on periodic crystal structure. At the $k^{th}$ layer message passing, the Equivariant Graph Convolutional Layer (EGCL) takes as input the set of atom embeddings $\textbf{\textit{h}}^{k}=[\textbf{\textit{h}}^{k}_1,\textbf{\textit{h}}^{k}_2,...,\textbf{\textit{h}}^{k}_N]$, atom coordinates $\textbf{\textit{x}}^{k}=[\textbf{\textit{x}}^{k}_1,\textbf{\textit{x}}^{k}_2,...,\textbf{\textit{x}}^{k}_N]$ and Lattice Matrix \textbf{\textit{L}} and outputs a transformation on $\textbf{\textit{h}}^{k+1}$. Formally, we can define the $k^{th}$ layer message passing operation as follows :
\begin{equation}
\label{eq:appendix_msg_pass_1}
    \textbf{\textit{m}}_{i,j} = \rho_m \{\textbf{\textit{h}}^{k}_i, \ \textbf{\textit{h}}^{k}_j, \ \textbf{\textit{L}}^T\textbf{\textit{L}}, \ 
    \psi_{FT}(\textbf{\textit{x}}^{k}_i - \textbf{\textit{x}}^{k}_j ) \} ; \;
\end{equation}
\begin{equation}
\label{eq:appendix_msg_pass_2}
    \textbf{\textit{h}}^{k+1}_{i} = \textbf{\textit{h}}^{k}_{i} + \rho_h \{ \textbf{\textit{h}}^{k}_{i}, \textbf{\textit{m}}_{i} \}
\end{equation}
where $\textbf{\textit{m}}_{i} = \sum_{j=1}^{N} \textbf{\textit{m}}_{i,j}$,  $\rho_m, \rho_h$ are multi-layer perceptrons and $\psi_{FT}$ is a Fourier Transformation function applied on relative difference between fractional coordinates $\textbf{\textit{x}}^{k}_i$, $ \textbf{\textit{x}}^{k}_j$. Fourier Transformation is used since it is invariant to periodic translation and extracts various frequencies of all relative fractional distances that are helpful for crystal structure modeling.\\
We fuse textual representation $\textbf{\textit{C}}_\textbf{\textit{p}}$ into input atom feature $\textbf{\textit{h}}^{0}_{i}$ as 
\begin{equation}
\label{eq:appendix_text_fusion}
   \textbf{\textit{h}}^{0}_{i} = \rho \ \{ \ f_{atom}(\textbf{\textit{a}}_i) \ \lvert \rvert \  f_{pos}(t) \ \lvert \rvert  \ \textbf{\textit{C}}_\textbf{\textit{p}} 
\end{equation}
where t is the timestamp of the diffusion model, $f_{pos}(.)$ is sinusoidal
positional encoding ~\citep{ho2020denoising,vaswani2017attention}, $f_{atom}(.)$ learned atomic embedding function and $\lvert \rvert$ is concatenation operation. Input atom features $\textbf{\textit{h}}^{0}$ and coordinates $\textbf{\textit{x}}^{0}$ are fed through $\mathcal{K}$ layers of EGCL to produce $\hat{\bm{\epsilon}}^{\textbf{\textit{L}}}$, $p(\textbf{\textit{A}}_{t-1} \ | \ \textbf{\textit{M}}_{t})$ and $\hat{\bm{\epsilon}}^{\textbf{\textit{X}}}$ as follows :
\begin{equation}
    \label{eq:appendix_msg_pass_3}
    \begin{split}
        \hat{\bm{\epsilon}}^{\textbf{\textit{L}}} = \textbf{\textit{L}} \rho_L (\frac{1}{N} \sum^{i=1}_{N} \textbf{\textit{h}}^{\mathcal{K}}) ; \; \\
        p(\textbf{\textit{A}}_{t-1} \ | \ \textbf{\textit{M}}_{t}) = \rho_A (\textbf{\textit{h}}^{\mathcal{K}}) ; \; \\
        \hat{\bm{\epsilon}}^{\textbf{\textit{X}}} = \rho_X (\textbf{\textit{h}}^{\mathcal{K}})
    \end{split}
\end{equation}
where $\rho_L, \rho_A, \rho_X$ are multi-layer perceptrons on the final layer embeddings. Intuitively, we feed global structural knowledge about the crystal structure into the network by injecting contextual representation $\textbf{\textit{C}}_\textbf{\textit{p}}$ into input atom features. This added signal will participate through message-passing operations in Eq. \ref{eq:appendix_msg_pass_1} and guides in denoising atom types, coordinates, and lattice parameters such that it can capture the global crystal geometry and aligned with the input stable structure specified by textual description.
\begin{algorithm}
\caption{Training Algorithm}\label{alg:training}
\begin{algorithmic}[1]
\State \textbf{Input:} Atom type Matrix $\textbf{\textit{A}}_0$ (One hot Vector Representation), Coordinate Matrix $\textbf{\textit{X}}_0$, Lattice matrix $\textbf{\textit{L}}_0$, Markov Transition Matrix $[\textbf{\textit{Q}}_{t}]^T_{t=1}$, Textual Representation $\textbf{\textit{C}}_\textbf{\textit{p}}$, Number of diffusion step T and hyperparameters $\lambda_{\textbf{\textit{A}}}$, $\lambda_{\textbf{\textit{X}}}$, $\lambda_{\textbf{\textit{L}}}$.
\Repeat 
\State Sample $t \sim  \mathcal{U}(\mathbf{0},\mathbf{T})$
\State Sample Noise $\epsilon^\textbf{X}, \epsilon^\textbf{L} \sim N(0,I)$
\State $\textbf{\textit{L}}_{t} = \sqrt{\bar{\alpha}_{t}}\textbf{\textit{L}}_{0} +  \sqrt{1-\bar{\alpha}_{t}}\bm{\epsilon}^{\textbf{\textit{L}}} $
\State $\textbf{\textit{X}}_{t} = f_w(\textbf{\textit{X}}_{0} + \bm{\sigma_t}\bm{\epsilon}^{x})$
\State $\textbf{\textit{A}}_{t} = Cat(\textbf{\textit{A}}_t;\textbf{\textit{p}} = \textbf{\textit{A}}_{t-1}\textbf{\textit{Q}}_{t})$
\State $\hat{\epsilon}^{\textbf{L}},\hat{\epsilon}^{\textbf{X}},\textbf{A}'_{t}  \leftarrow \Phi_\theta(\textbf{A}_{t},\textbf{X}_{t},\textbf{L}_{t},t,\textbf{C}_\textbf{p})$
\State $\mathcal{L}_{lattice} = \lVert  \bm{\epsilon}^{\textbf{\textit{L}}} \ - \ \hat{\bm{\epsilon}}^{\textbf{\textit{L}}} {\rVert}^2_2 $
\State $\mathcal{L}_{coord} = 
        \lVert  \nabla_{\textbf{\textit{X}}_{t}} \text{log} q(\textbf{\textit{X}}_{t} | \textbf{\textit{X}}_{0}) - \hat{\bm{\epsilon}}^{X} {\rVert}^2_2$
\State $\mathcal{L}_{type} = \mathcal{L}_{VB}+\lambda\mathcal{L}_{CE}$
\State Minimize $\mathcal{L} = \lambda_{L} \mathcal{L}_{lattice} + \lambda_{A} \mathcal{L}_{type} + \lambda_{X} \mathcal{L}_{coord}$ and update parameters of $\Phi_\theta$
\Until{converged}
\end{algorithmic}
\end{algorithm}
\begin{algorithm}
\caption{Sampling Algorithm}\label{alg:sampling}
\begin{algorithmic}[1]
\State Sample $\textbf{\textit{L}}_T \sim  \mathcal{N}(\mathbf{0},\mathbf{I}),\textbf{\textit{X}}_T \sim  \mathcal{U}(0,1)$
\State Randomly sample each atom type between 0 to 99 (Max possible atom type) and form $\textbf{\textit{A}}_T$
\State $\textbf{\textit{C}}_\textbf{\textit{p}} \gets \text{Textual Representation}$
\For{$t \gets T$ to $1$}
    \State $\epsilon^\textbf{A},\epsilon^\textbf{X},\epsilon^\textbf{L} \sim N(0,I) /*Sample*/$
    \State $\hat{\textbf{A}},\hat{\epsilon}^{\textbf{X}},\hat{\epsilon}^{\textbf{L}} \leftarrow \Phi_\theta(\textbf{A}_{t},\textbf{X}_{t},\textbf{L}_{t},t,\textbf{C}_\textbf{p})$
    
    \State $\textbf{L}_{t-1} \leftarrow \frac{1}{\sqrt{\alpha_{t}}} (\textbf{L}_{t} - \frac{\beta_t}{\sqrt{1-\bar{\alpha}_{t}}}\hat{\epsilon}^{\textbf{L}})+\sqrt{\beta_t\frac{1-\bar{\alpha}_{t-1}}{1-\bar{\alpha}_{t}}}\epsilon^{\textbf{L}}$
    \State $\textbf{A}_{t-1} \leftarrow \text{Softmax}(\hat{\textbf{A}}+\bm{\sigma_t}\epsilon^\textbf{A})$
    \State $\textbf{X}_{t-\frac{1}{2}} \leftarrow w(\textbf{X}_{t}+(\sigma^2_t - \sigma^2_{t-1})\hat{\epsilon}^{\textbf{X}}+\frac{\sigma_{t-1}\sqrt{\sigma^2_t - \sigma^2_{t-1}}}{\sigma_{t}}\epsilon^\textbf{X})$
    \State $_,\hat{\epsilon}^{\textbf{X}} \leftarrow \Phi_\theta(\textbf{A}_{t},\textbf{X}_{t-\frac{1}{2}},\textbf{L}_{t-1},t,\textbf{C}_\textbf{p})$
    \State $\eta_t \leftarrow step\_size * \frac{\sigma_{t-1}}{\sigma_t}$
    \State $\textbf{X}_{t-1} \leftarrow w(\textbf{X}_{t-\frac{1}{2}}+\eta_t\hat{\epsilon}^{\textbf{X}}+\sqrt{2\eta_t}\epsilon^\textbf{X})$   
\EndFor
\end{algorithmic}
\end{algorithm}
\subsection{Training and Sampling}
\label{appendix_training}
\our{} is trained using the following combined loss:\\
\begin{equation}
    \label{eq:appendix_loss}
        \mathcal{L} = \lambda_{L} \mathcal{L}_{lattice} + \lambda_{A} \mathcal{L}_{type} + \lambda_{X} \mathcal{L}_{coord}
\end{equation}
\\
where $\mathcal{L}_{lattice}$, $\mathcal{L}_{type}$ and $\mathcal{L}_{coord}$ are lattice $l_2$ loss (Eq. \ref{eq:appendix_lattice_loss}), type cross-entropy loss (Eq. \ref{eq:appendix_type_loss}) and coordinate score matching loss (Eq. \ref{eq:appendix_coord_loss}) respectively and $\lambda_{L}$, $\lambda_{A}$, $\lambda_{X}$ are hyperparameters control the relative weightage between these different loss components. During training, we freeze the MatSciBERT parameters and do not tune it further. During sampling, we use the Predictor-Corrector sampling mechanism to sample $\textbf{\textit{A}}_0$, $\textbf{\textit{X}}_0$ and $\textbf{\textit{L}}_0$. Next we explain algorithms for training and sampling.

\section{Experiments}\label{appendix_results}
\subsection{Experimental Setup}\label{appendix_results_expsetup}
\begin{figure}
    \centering
	\subfloat[Random Material Generation (Gen) Task]{\includegraphics[width=0.45\columnwidth]{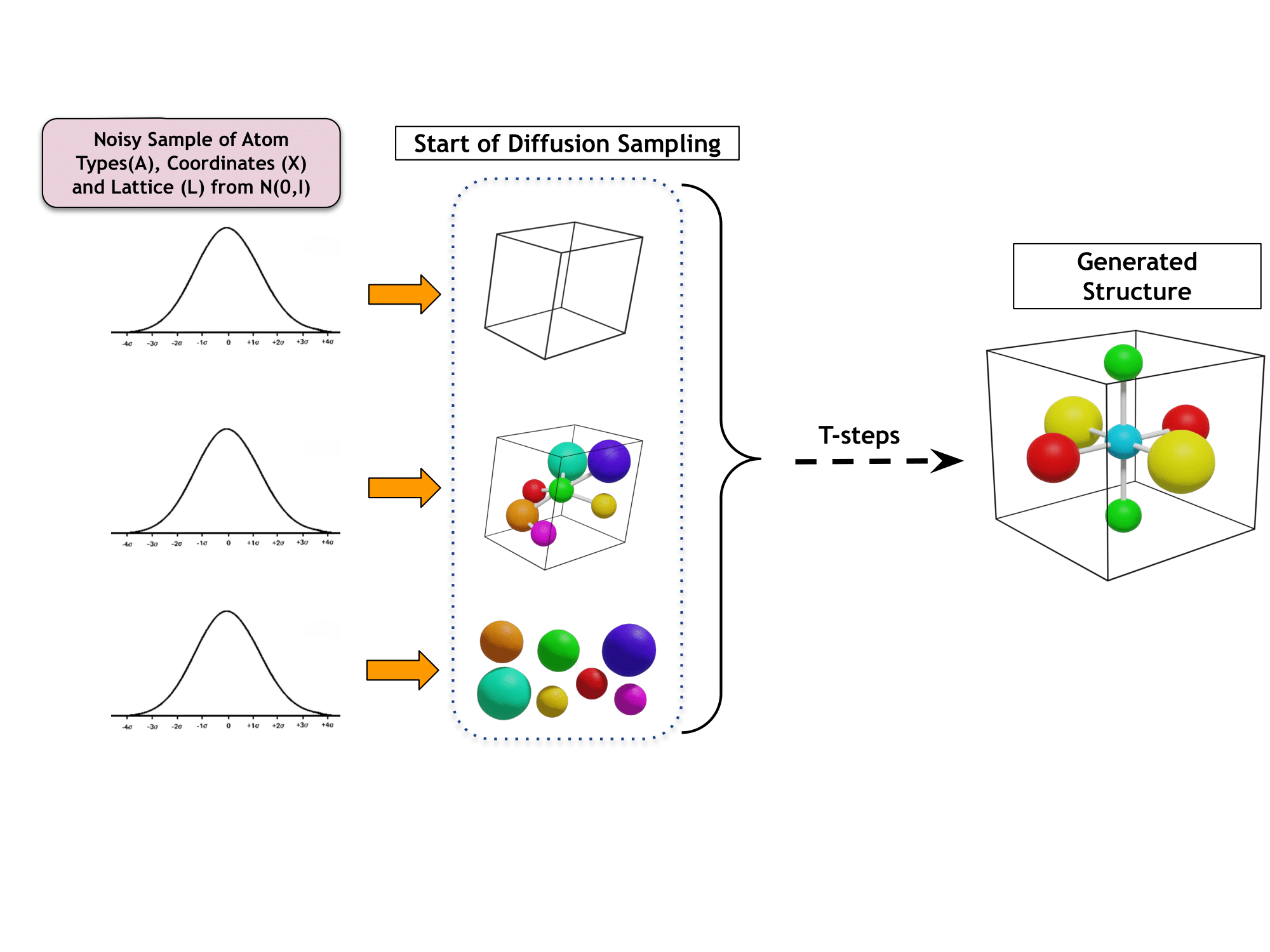}}
    \hspace{10mm}
    \subfloat[Crystal Structure Prediction (CSP) Task]
    {\includegraphics[width=0.45\columnwidth]{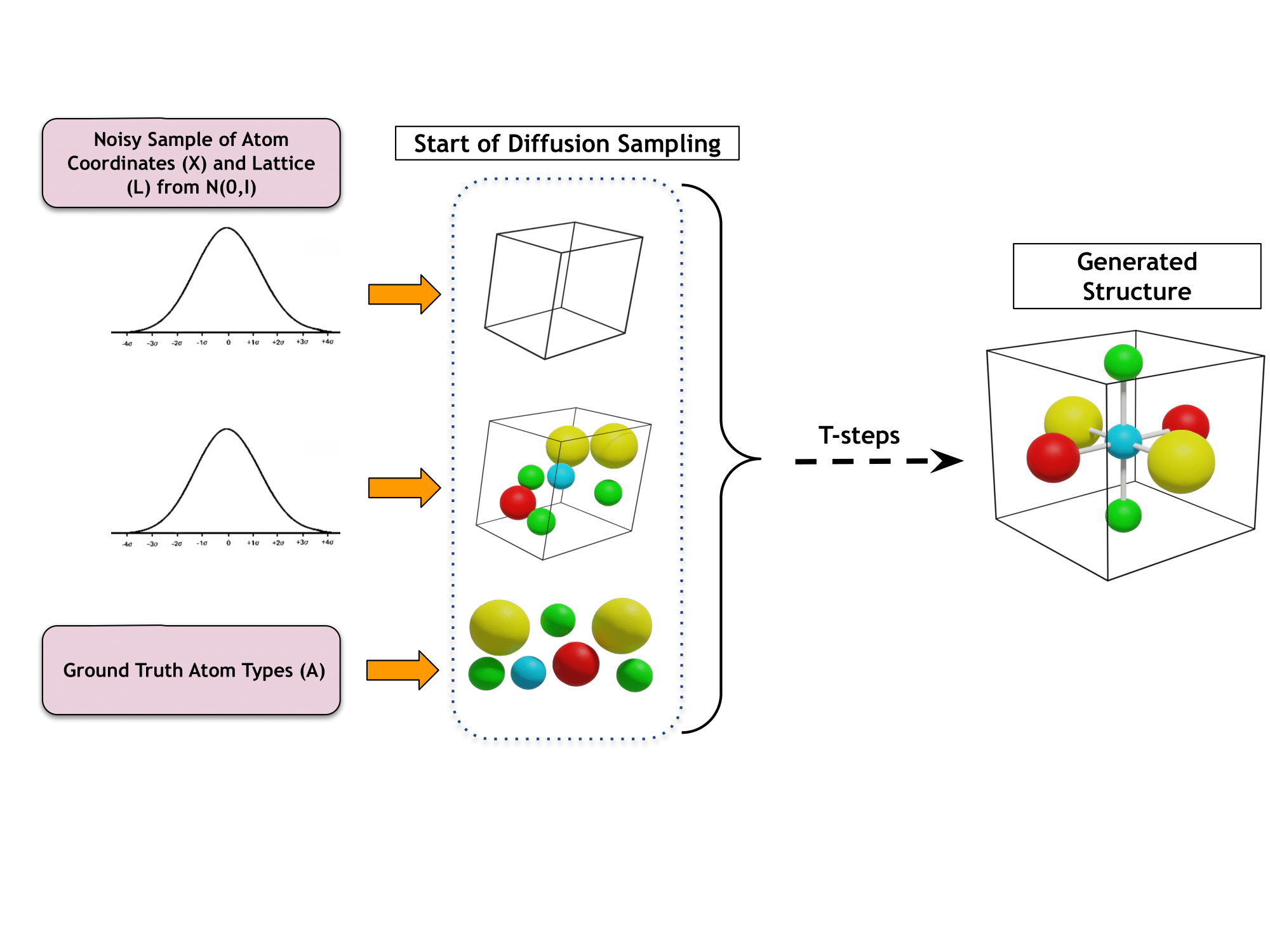}}
	\caption{}
	\label{fig:task}
\end{figure}

\xhdr{Benchmark Tasks}
We evaluate our proposed model \our{} on two different categories of tasks for material generation, \textit{Random Material Generation (Gen)} and \textit{Crystal Structure Prediction (CSP)}. In \textit{Gen} task, the goal of the generative model is to generate novel stable materials (atom types, fractional coordinates, and lattice structure). In \textit{CSP} task, atom types of the materials are given and the goal is to predict/match the crystal structure (atom coordinates and lattice). In \our{} model, by design choice, we use the textual description of crystal materials during each step of the reverse diffusion process to enhance the generation capability in both tasks. A pictorial illustration of both tasks is provided at \ref{fig:task}
\\\\
\xhdr{Dataset} Following Xie et al ~\citep{xie2021crystal} we evaluate our model on three baseline datasets: \textbf{Perov-5}, \textbf{Carbon-24} and \textbf{MP-20}. \textbf{Perov-5} ~\citep{castelli2012new,castelli2012computational} dataset consists of 18,928 perovskite materials, each with 5 atoms in a cell. They generally can be denoted by $\mathbf{ABX_3}$ indicating the three different types of atoms usually observed in such materials. \textbf{Carbon-24} ~\citep{carbon} dataset has 10,153 materials with 6 to 24 atoms of carbon in the crystal lattice. Finally, \textbf{MP-20}~\citep{jain2013materials} dataset has 45,231 materials curated from the Materials Project library ~\citep{10.1063/1.4812323}, where each material has at most 20 atoms in the lattice. Crystals from \textbf{Perov-5} dataset share the same structure but differ in composition, whereas Crystals from \textbf{Carbon-24} share the same composition but differ in structure.  Crystals from \textbf{MP-20}  differs in both structure and composition. We curated textual data for these datasets with a textual description of each material. Specifically, we generate both long detailed textual descriptions and shorter prompts using approaches mentioned in Appendix \ref{appendix_text_data}. \\
The structures in all three datasets are derived from quantum mechanical simulations and are all at local energy minima. Most materials in \textbf{Perov-5} and \textbf{Carbon-24} are hypothetical, whereas \textbf{MP-20} represents a realistic dataset that includes many experimentally known inorganic materials, each with a maximum of 20 atoms in the unit cell, most of which are globally stable. A model that performs well on MP-20 could potentially generate novel materials that can be synthesized experimentally. While training \our{}, we split the datasets into the train, test, and validation sets following the convention of 60:20:20 as done by Xie et al ~\citep{xie2021crystal}.\\\\
\xhdr{Hyper-Parameters Details}
In our \our{} model, we adopted 4 layers CSPNet as message passing layer with hidden dimension set as 512. Further, we use pre-trained MatSciBERT~\citep{gupta_matscibert_2022} followed by a two-layer projection layer (projection dimension 64) as the text encoder module. We keep the dimension of time embedding at each diffusion timestep as 64. We train it for 500 epochs using the same optimizer, and learning rate scheduler as DiffCSP and keep the batch size as 512. We perform all the experiments in the Tesla P100-PCIE-16GB GPU server.
\subsection{Evaluation Metrics}\label{appendix_results_metrics}
\xhdr{Random Material Generation (Gen) Task} Following CDVAE~\citep{xie2021crystal}, we evaluate the performance of \our{} and baseline models on generating novel material structure using seven metrics under three broad categories: \textbf{Validity}, \textbf{Coverage}, and \textbf{Property Statistics}. Under \textbf{Validity}, following the prior line of work ~\citep{court20203,xie2021crystal}, we measure structural and compositional validity, representing the percentages of generated crystals with valid periodic structures and atom types, respectively. A structure is valid as long as the shortest distance between any pair of atoms is larger than 0.5\textup{~\AA}  whereas the composition is valid if the overall charge is neutral as computed by SMACT~\citep{davies2019smact}. In \textbf{Coverage}, we consider two coverage metrics, COV-R (Recall) and COV-P (Precision). COV-R measures the percentage of the test set materials being correctly predicted, whereas COV-P measures the percentage of generated materials that cover at least one of the test set materials. (More detailed discussions can be found in  ~\citep{xie2021crystal} and ~\citep{ganea2021geomol}). Finally, we evaluate the similarity between the generated materials and those in the test set using various \textbf{Property Statistics}, where we compute the earth mover’s distance (EMD) between the distributions in element number (\# Elem), density ($\rho$, unit g/cm3), and formation energy ($\mathcal{E}$, unit eV/atom) predicted by a GNN model.\\\\
\xhdr{Crystal Structure Prediction (CSP) Task} We evaluate the performance of \our{} and baseline models on stable structure prediction using standard metrics proposed by the prior works ~\citep{jiao2023crystal,xie2021crystal}, by matching the generated structure and the input ground truth structure in the test set. In Specific, for each material structure in the test set, we generate k samples given the textual description and then identify the matching if at least one of the samples matches the ground truth structure. We calculate the \textbf{Match Rate} and \textbf{RMSE} metrics using the StructureMatcher class in Pymatgen, which identifies the best match between two structures while accounting for all material invariances. Match rate indicates the percentage of the matched structures over the test set satisfying thresholds stol=0.5, angle\_tol=10, ltol=0.3. RMSE is computed between the ground truth and the best-matching candidate, normalized by $\sqrt[3]{V/N}$ where V is the volume of the lattice, and averaged over the matched structures. For baselines and \our{}, we evaluate using $k = 1$ and $k = 20$.

\subsection{Complete and Detailed Results}
\label{sec-full-results}
In this subsection, we provide full comprehensive results on both Gen and CSP tasks across three benchmark datasets and evaluate the performance of all the baseline models, their text-guided variants (both short and long), and our proposed \our{}(Long) \& \our{}(Short). We report the CSP and Gen task results in Table \ref{tbl-csp-full} and \ref{tbl-gen-full} respectively. \\\\
Following are the Insights or Observations:
\begin{itemize}
    \item For both tasks, across all the datasets, text guidance outperforms the vanilla diffusion models in almost all metrics.
    \item Our experiments suggest that using shorter prompts text-guided models outperforms the vanilla baseline models. However, performance is even superior when using text-guided diffusion using longer prompts.
    \item For the CSP task, using text guidance during the reverse denoising process, with just one generated sample per test material, text-guided variants outperform respective vanilla models, thereby reducing computational overhead.
    \item Our proposed TGDMat (Long) stands out as the leading model when compared to all baseline models and their text-guided variants across three benchmark datasets. In specific, for Gen Task, TGDMat (Long) outperforms the closest baseline DiffCSP+ (Long) because we leveraged discrete diffusion on atom types, which is more powerful in learning discrete variables like atom types.
    \item Finally, results indicate that utilizing shorter prompts TGDMat (Short) results in a slight decrease in overall performance compared to the longer variant TGDMat (Long). Nonetheless, the performance remains superior or comparable to baseline models (vannila and text-guided variants).
\end{itemize}
\begin{table*}
  \centering
  % \begin{tabular}{ c|c|c|c|c|c|c|c }
  \begin{tabular}{ c | c |cccccc }
    % \hline
    \toprule
    \multirow{2}*{Method} & \multirow{2}*{\# samples} & \multicolumn{2}{c}{Perov-5} & \multicolumn{2}{c}{Carbon-24} & \multicolumn{2}{c}{MP-20} \\
    % \cline{3-8}
    \cmidrule{3-8}
     & & Match & RMSE & Match & RMSE & Match & RMSE \\
    % \hline
    \midrule
    \multirow{2}*{CDVAE} & 1 & 45.31 & 0.1138 & 17.09 & 0.2969 & 33.9 & 0.1045 \\
    % \cline{2-8}
    \cmidrule{2-8}
     & 20 & 88.51 & 0.0464 & 88.37 & 0.2286 & 66.95 & 0.1026 \\
    % \hline
    \midrule
    \multirow{2}*{CDVAE+(short)} & 1 & 48.97 & 0.1063 & 22.65 & 0.264 & 40.33 & 0.1037 \\
    % \cline{2-8}
    \cmidrule{2-8}
     & 20 & 89.54 & 0.0423 & 89.61 & 0.2188 & 70.22 & 0.0876 \\
    % \hline
    \midrule
    \multirow{2}*{CDVAE+(long)} & 1	 & 49.25 & 0.1055 & 23.73 & 0.259 & 41.8 & 0.1021 \\
    % \cline{2-8}
    \cmidrule{2-8}
     & 20 & 89.73 & 0.0417 & 89.77 & 0.2053 & 72.56 & 0.084 \\
    % \hline
    \midrule
    \multirow{2}*{SyMat} & 1 & 47.32 & 0.1074 & 20.81 & 0.2655 & 33.92 & 0.1039 \\
    % \cline{2-8}
    \cmidrule{2-8}
     & 20 & 90.25 & 0.0316 & 89.29 & 0.2184 & 71.03 & 0.0945 \\
    % \hline
    \midrule
    \multirow{2}*{SyMat+(short)} & 1 & 49.39 & 0.0985 & 23.71 & 0.2567 & 40.84 & 0.1027 \\
    % \cline{2-8}
    \cmidrule{2-8}
     & 20 & 92.1 & 0.0255 & 90.86 & 0.2069 & 71.31 & 0.0875 \\
    % \hline
    \midrule
    \multirow{2}*{SyMat+(long)} & 1 & 50.88 & 0.0963 & 28.18 & 0.251 & 43.17 & 0.1016 \\
    % \cline{2-8}
    \cmidrule{2-8}
     & 20 & 92.3 & 0.0201 & 91.65 & 0.187 & 72.96 & 0.082 \\
    % \hline
    \midrule
    \multirow{2}*{DiffCSP} & 1 & 52.02 & 0.076 & 17.54 & 0.2759 & 51.49 & 0.0631 \\
    % \cline{2-8}
    \cmidrule{2-8}
     & 20 & 98.6 & 0.0128 & 88.47 & 0.2192 & 77.93 & 0.0492 \\
    % \hline
    \midrule
    \multirow{2}*{DiffCSP+(short)} & 1 & 56.54 & 0.0583 & 24.13 & 0.2424 & 52.22 & 0.0597 \\
    % \cline{2-8}
    \cmidrule{2-8}
     & 20 & 98.25 & 0.0137 & 88.28 & 0.2252 & 80.97 & 0.0443 \\
    % \hline
    \midrule
    \multirow{2}*{DiffCSP+(long)} & 1 & 90.46 & 0.0203 & 44.63 & 0.2266 & 55.15 & 0.0572 \\
    % \cline{2-8}
    \cmidrule{2-8}
     & 20 & 98.59 & 0.0072 & 95.27 & 0.1534 & 82.02 & 0.0391 \\
    % \hline
    \midrule
    \multirow{2}*{TGDMat (short} & 1 & 56.54 & 0.0583 & 24.13 & 0.2424 & 52.22 & 0.0597 \\
    % \cline{2-8}
    \cmidrule{2-8}
     & 20 & 98.25 & 0.0137 & 88.28 & 0.2252 & 80.97 & 0.0443 \\
    % \hline
    \midrule
    \multirow{2}*{TGDMat (long)} & 1 & 90.46 & 0.0203 & 44.63 & 0.2266 & 55.15 & 0.0572 \\
    % \cline{2-8}
    \cmidrule{2-8}
     & 20 & 98.59 & 0.0072 & 95.27 & 0.1534 & 82.02 & 0.0391 \\
    % \hline
    \bottomrule
  \end{tabular}
  \caption{Summary of the Complete and Detailed Results on the CSP Task.}
  \label{tbl-csp-full}
\end{table*}
\begin{table*}
  \centering
  \begin{tabular}{ ccccccccc }
    \toprule
    \multirow{2}*{Dataset} & \multirow{2}*{Method} & \multicolumn{2}{c}{Validity} & \multicolumn{2}{c}{Coverage} & \multicolumn{3}{c}{Property} \\ \cmidrule{3-9}
     & & Comp & Struct & Cov-R & Cov-P & \# element & Density & form\_energy \\ \midrule
    \multirow{11}*{Perov 5} & CDVAE & 98.29 & 100 & 99.25 & 98.39 & 0.0731 & 0.1462 & 0.0291 \\ \cmidrule{2-9}
     & CDVAE+ (Short) & 98.17 & 100 & 99.4 & 99.01 & 0.0706 & 0.1395 & 0.0246 \\ \cmidrule{2-9}
     & CDVAE+ (Long) & 98.45 & 100 & 99.53 & 99.09 & 0.0609 & 0.1276 & 0.0223 \\ \cmidrule{2-9}
     & SyMat & 96.83 & 100 & 99.16 & 98.29 & 0.0193 & 0.1991 & 0.2827 \\ \cmidrule{2-9}
     & SyMat+ (Short) & 96.94 & 100 & 99.22 & 98.4 & 0.0192 & 0.1827 & 0.2633 \\ \cmidrule{2-9}
     & SyMat+ (Long) & 97.88 & 100 & 99.71 & 98.79 & 0.0172 & 0.1755 & 0.2566 \\ \cmidrule{2-9}
     & DiffCSP & 98.15 & 100 & 99.28 & 98.08 & 0.0132 & 0.1281 & 0.0267 \\ \cmidrule{2-9}
     & DiffCSP+ (Short) & 98.21 & 100 & 99.61 & 98.39 & 0.0123 & 0.1193 & 0.0266 \\ \cmidrule{2-9}
     & DiffCSP+ (Long) & 98.44 & 100 & 99.85 & 98.53 & 0.0119 & 0.1071 & 0.0241 \\ \cmidrule{2-9}
     & TGDMat(Short) & 98.28 & 100 & 99.71 & 99.24 & 0.0108 & 0.0947 & 0.0237 \\ \cmidrule{2-9}
     & TGDMat(Long) & 98.63 & 100 & 99.87 & 99.52 & 0.009 & 0.0497 & 0.0187 \\ \midrule
    \multirow{11}*{Carbon 24} & CDVAE & - & 100 & 99.35 & 82.66 & - & 0.1539 & 0.2889 \\ \cmidrule{2-9}
     & CDVAE+ (Short) & - & 100 & 99.34 & 82.96 & - & 0.1398 & 0.2804 \\ \cmidrule{2-9}
     & CDVAE+ (Long) & - & 100 & 99.82 & 84.76 & - & 0.1377 & 0.266 \\ \cmidrule{2-9}
     & SyMat & - & 100 & 99.42 & 97.17 & - & 0.1234 & 3.9628 \\ \cmidrule{2-9}
     & SyMat+ (Short) & - & 100 & 99.52 & 97.2 & - & 0.1206 & 3.7422 \\ \cmidrule{2-9}
     & SyMat+ (Long) & - & 100 & 99.9 & 97.63 & - & 0.1171 & 3.862 \\ \cmidrule{2-9}
     & DiffCSP & - & 99.9 & 99.49 & 97.27 & - & 0.0861 & 0.0876 \\ \cmidrule{2-9}
     & DiffCSP+ (Short) & - & 100 & 99.61 & 97.29 & - & 0.0811 & 0.087 \\ \cmidrule{2-9}
     & DiffCSP+ (Long) & - & 100 & 99.93 & 97.33 & - & 0.0763 & 0.0853 \\ \cmidrule{2-9}
     & TGDMat(Short) & - & 100 & 99.81 & 91.77 & - & 0.0681 & 0.0865 \\ \cmidrule{2-9}
     & TGDMat(Long) & - & 100 & 99.91 & 92.43 & - & 0.0436 & 0.0632 \\ \midrule
    \multirow{11}*{MP 20} & CDVAE & 86.3 & 100 & 99.15 & 99.49 & 1.4921 & 0.7085 & 0.3039 \\ \cmidrule{2-9}
     & CDVAE+ (Short) & 87.05 & 100 & 99.36 & 99.6 & 0.993 & 0.642 & 0.297 \\ \cmidrule{2-9}
     & CDVAE+ (Long) & 87.42 & 100 & 99.57 & 99.81 & 0.972 & 0.6388 & 0.2977 \\ \cmidrule{2-9}
     & SyMat	& 87.96 & 99.9 & 98.3 & 99.37 & 0.5236 & 0.4012 & 0.3877 \\ \cmidrule{2-9}
     & SyMat+ (Short) & 88.08 & 99.9 & 98.59 & 99.47 & 0.5031 & 0.3917 & 0.3622 \\ \cmidrule{2-9}
     & SyMat+ (Long) & 88.47 & 99.9 & 99.01 & 99.95 & 0.4865 & 0.3879 & 0.3489 \\ \cmidrule{2-9}
     & DiffCSP & 83.25 & 100 & 99.41 & 99.76 & 0.3411 & 0.3802 & 0.1497 \\ \cmidrule{2-9}
     & DiffCSP+ (Short) & 84.57 & 100 & 99.52 & 99.85 & 0.331 & 0.38 & 0.1379 \\ \cmidrule{2-9}
     & DiffCSP+ (Long) & 85.07 & 100 & 99.81 & 99.89 & 0.3122 & 0.3799 & 0.1355 \\ \cmidrule{2-9}
     & TGDMat(Short) & 86.6 & 100 & 99.79 & 99.88 & 0.3337 & 0.3296 & 0.1189 \\ \cmidrule{2-9}
     & TGDMat(Long) & 92.97 & 100 & 99.89 & 99.95 & 0.289 & 0.3082 & 0.1154 \\ \bottomrule
  \end{tabular}
  \caption{Summary of the Complete and Detailed Results on the Gen Task.}
  \label{tbl-gen-full}
\end{table*}

\subsection{Correctness of Generated Materials}
\label{result_corectness_appendix}
\xhdr{Setup} In this section, we investigate whether the generated material matches different features specified by the textual prompts. \our{} has the capability to process textual prompts given by the user, enabling it to manage global attributes about crystal materials such as Formula, Space group, Crystal System, and different property values like formation energy, band-gap, etc. To ensure the fidelity of our model's outputs concerning these specified global attributes from the text prompt, We randomly generated 1000 materials (sampled from all three Datasets) based on their respective textual descriptions(both Long and Short) and assessed the percentage of generated materials that matched the global features outlined in the text prompt. In specific, we matched the Formula, Space group, and Crystal System, and Dimensions of generated materials with the textual descriptions. Moreover, we examined whether properties such as formation energy and bandgap matched the specified criteria as per the text prompt (positive/negative, zero/nonzero). \\\\\\
\xhdr{Results and Discussions} We report the results in Table \ref{tbl-match-appendix}. 
In general, using longer text, considering Perov-5 and Carbon-24 datasets, the generated material meets the specified criteria effectively. However, when dealing with the MP-20 dataset, which is more intricate due to its complex structure and composition, performance tends to decline. Additionally, when using shorter prompts, overall performance suffers across all datasets compared to longer text inputs. This is because the longer text, provided by the robocrystallographer, offers a comprehensive range of information, both global and local, thereby enhancing the generation capabilities of \our{}.

\begin{table*}[ht]
  \centering
  \small
    \setlength{\tabcolsep}{10 pt}
    \scalebox{1}{
\begin{tabular}{c | c | c c c }
\toprule
Method & Global Features  & \multicolumn{3}{c}{\% of Matched Materials}  \\
& in Text Prompt   & Perov-5 & Carbon-24 & MP-20 \\
\midrule
% RS
\multirow{4}*{\our{}(Long)} 
& Formula           & 97.50 & 98.20 & 70.54\\
& Space Group       & 87.00 & 80.79 & 67.88\\
& Crystal System    & 92.60 & 91.55 & 73.54\\
& Formation Energy  & 95.49 & - & 92.88\\
& Band Gap          & - & 98.61 & 96.73\\
\midrule
\midrule
\multirow{4}*{\our{}(Long)} 
& Formula                   & 90.70 & 92.56 & 65.22\\
& Space Group               & 86.51 & 80.50 & 58.77\\
& Crystal System            & 83.19 & 81.64 & 72.77 \\
& Formation Energy          & 90.33 & - & 91.00\\
& Band Gap                  & - & 95.90 & 93.33\\

\bottomrule
\end{tabular} 
}
\caption{Summary of results on \% of generated materials matching different global features specified by the textual prompts.
}
  \label{tbl-match-appendix}
\end{table*}

\subsection{Performance on More Shorter Prompts}
\label{sec-shorter-prompt}
In this section, we explore the generalizability and robustness of our model by examining potential variability in text description lengths. The goal of this paper is, given the text prompt, to generate specific material, not any generic or class of materials. Hence some minimum essential information about the crystal, like formula, space group, crystal system, property value, etc must be given as input to the pre-trained model. However, to investigate the robustness of our proposed \our{} model with more custom and shorter prompts, we did an experiment where we evaluated \our{} (trained with full text) with even shorter custom prompts with very little information as follows:
\begin{itemize}
    \item \textbf{Specifying only Formula: }\textit{"The chemical formula is GaSiSO2. The elements are Ga, Si, S, O. Generate the material."}
    \item \textbf{Specifying only Space Group Info: }\textit{"The spacegroup number is 1. Generate the material."}
    \item \textbf{Specifying only Property Info: }\textit{"The formation energy per atom is positive. Generate the material."}
\end{itemize}
We report the results in table \ref{tbl-short-prompt}. We observe that though \our{} can handle more custom prompts, but it affects the quality of generated materials. Hence we conclude some minimum essential information about the crystal must be given as input to \our{} to generate high quality crystal materials.

\begin{table*}[ht]
  \centering
  \small
    \setlength{\tabcolsep}{10 pt}
   \scalebox{0.8}{
\begin{tabular}{ c|c|c|c|c|c|c }
% \hline
\toprule
% Header
\multirow{2}*{Text Encoder} & \multicolumn{2}{c|}{Perov-5} & \multicolumn{2}{c|}{Carbon-24} & \multicolumn{2}{c}{MP-20} \\
% \cline{2-7}
& Comp(\%) $\uparrow$ & Struct (\%)  $\uparrow$ & Comp(\%) $\uparrow$ &  Struct (\%) $\uparrow$ & Comp(\%) $\uparrow$ &  Struct (\%) $\uparrow$ \\
\midrule
 Only Formula     & 97.06 & 99.19 & - & 98.76 & 86.16 & 96.01\\
 Only Space Group & 85.91 & 98.97 & - & 95.39 & 84.22 & 96.88\\
 Only Property    & 96.62 & 98.53 & - & 94.21 & 86.53 & 91.73\\
 Full Text & \textbf{98.28} &\textbf{100}   & - & \textbf{100}   & \textbf{86.60} & \textbf{100} \\
 \bottomrule
\end{tabular}
}
\caption{Summary of results on generated materials using more custom/shorter Prompt.}
  \label{tbl-short-prompt}
\end{table*}

\newpage

\begin{table*}[ht]
\centering
  \small
\caption{Ablation study results on different choices of Text Encoders.}
\label{tbl-text-enc}
\renewcommand{\arraystretch}{0.9}
\setlength{\tabcolsep}{6 pt}
\scalebox{0.8}{
\begin{tabular}{ c|c|c|c|c|c|c }
% \hline
\toprule
% Header
\multirow{2}*{Text Encoder} & \multicolumn{2}{c|}{Perov-5} & \multicolumn{2}{c|}{Carbon-24} & \multicolumn{2}{c}{MP-20} \\
% \cline{2-7}
& MR $\uparrow$ & RMSE $\downarrow$ & MR $\uparrow $& RMSE $\downarrow$ & MR $\uparrow $& RMSE $\downarrow$ \\
% \hline
\midrule
% BERT
BERT & 96.64 & 0.0109 & 72.21 & 0.2679 & 79.53 & 0.057 \\
% \hline

% MatSciBERT
MatSciBERT & \textbf{98.63} & \textbf{0.0072} & \textbf{95.27} & \textbf{0.1534} & \textbf{82.02} & \textbf{0.039} \\
% \hline
\midrule
% \midrule
& Comp $\uparrow$ & Struct $\uparrow$ & Comp $\uparrow$ & Struct $\uparrow$ & Comp $\uparrow$ & Struct $\uparrow$ \\
% \hline
\midrule
% T5
 % T5 & 94.02 & 99.86 & - & \textbf{100} \\
 % \hline
 BERT & 97.44 & 99.97 & - & \textbf{100} & 84.73 & 98.37\\
 % \hline
 MatSciBERT & \textbf{98.63} & \textbf{100} & - & \textbf{100} & \textbf{92.97} & \textbf{100} \\
 \bottomrule
\end{tabular}
}
\end{table*}
\subsection{Ablation Study : Choice of Text Encoder}
\label{result_text_enc_appendix}
Further, we investigate the expressiveness of textual representation during the reverse diffusion process. In particular, we are interested in understanding whether there are any benefits we are gaining from using a domain-specific pre-trained text encoder MatSciBERT. We conduct an ablation study where we substitute MatSciBERT with pre-trained BERT~\citep{devlin2018bert} model (which is domain agnostic) as text encoder in \our{} and evaluate the performance on both tasks. The results presented in Table \ref{tbl-text-enc} demonstrate that MatSciBERT surpasses BERT~\citep{devlin2018bert}
in performance for both tasks. This highlights the richer expressiveness of contextual representation achieved through the use of a domain-specific pre-trained language model.

\subsection{More Visualization on \textbf{Perov-5}, \textbf{Carbon-24} and \textbf{MP-20}}
\label{appendix-visualization}
%%%%%%%%% Perov-Sample %%%%%%%%%%%%%%%
\begin{table*}[ht]
  \centering
    \setlength{\tabcolsep}{1.5 pt}
    \renewcommand{\arraystretch}{1}
    \resizebox{1.0\textwidth}{!}{
    % \scalebox{0.7}{
    \begin{tabular}{m{3in}|m{2.2in}|c|cccc| } % angstrom symbol Å
\toprule
Detailed Description & Short Prompt &  Ground truth & \multicolumn{4}{c|}{Generated Samples} \\
\midrule
 
YCoSO2 crystallizes in the orthorhombic Pmm2 space group. Y is bonded in a distorted square co-planar geometry to two equivalent S, two equivalent O, and two equivalent O atoms. Both Y-S bond lengths are 2.74 Å. Both Y-O bond lengths are 2.24 Å. There is one shorter (2.09 Å) and one longer (2.39 Å) Y-O bond length. Co is bonded in a distorted see-saw-like geometry to two equivalent S and two equivalent O atoms. Both Co-S bond lengths are 2.31 Å. Both Co-O bond lengths are 2.40 Å. S is bonded in a 6-coordinate geometry to two equivalent Y, two equivalent Co, and two equivalent O atoms. Both S-O bond lengths are 2.30 Å. There are two inequivalent O sites. In the first O site, O is bonded in a rectangular see-saw-like geometry to two equivalent Y and two equivalent Co atoms. In the second O site, O is bonded in a distorted square co-planar geometry to two equivalent Y and two equivalent S atoms. & 
Below is a description of a bulk material. The chemical formula is YCoSO2. The elements are Y, Co, S, O. The formation energy per atom is positive. The spacegroup number is 24. The crystal system is orthorhombic. Generate the material:& 
\begin{minipage}{.1\linewidth}
  \includegraphics[width=\linewidth]{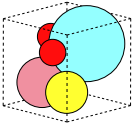}
\end{minipage} & 
\begin{minipage}{.1\linewidth}
  \includegraphics[width=\linewidth]{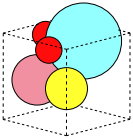}
\end{minipage} & 
\begin{minipage}{.1\linewidth}
  \includegraphics[width=\linewidth]{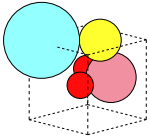}
\end{minipage} & 
\begin{minipage}{.1\linewidth}
  \includegraphics[width=\linewidth]{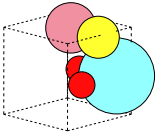}
\end{minipage} & 
\begin{minipage}{.1\linewidth}
  \includegraphics[width=\linewidth]{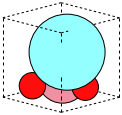}
\end{minipage} 
\\
  
\midrule
ScMoN2O is (Cubic) Perovskite-derived structured and crystallizes in the tetragonal P4mm space group. Sc is bonded to four equivalent N and two equivalent O atoms to form ScN4O2 octahedra that share corners with six equivalent ScN4O2 octahedra and faces with eight equivalent MoN8O4 cuboctahedra. The corner-sharing octahedral tilt angles range from 0-1°. All Sc-N bond lengths are 2.00 Å. There is one shorter (2.00 Å) and one longer (2.01 Å) Sc-O bond length. Mo is bonded to eight equivalent N and four equivalent O atoms to form MoN8O4 cuboctahedra that share corners with twelve equivalent MoN8O4 cuboctahedra, faces with six equivalent MoN8O4 cuboctahedra, and faces with eight equivalent ScN4O2 octahedra. There are four shorter (2.83 Å) and four longer (2.84 Å) Mo-N bond lengths. All Mo-O bond lengths are 2.83 Å. N is bonded in a linear geometry to two equivalent Sc and four equivalent Mo atoms. O is bonded in a linear geometry to two equivalent Sc and four equivalent Mo atoms. The formation energy per atom is 1.8931.&
Below is a description of a bulk material. The chemical formula is ScMoN2O. The elements are Sc, Mo, N, O. The formation energy per atom is positive. The spacegroup number is 98. The crystal system is tetragonal. Generate the material:& 
\begin{minipage}{.1\linewidth}
  \includegraphics[width=\linewidth]{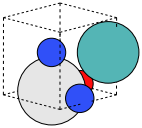}
\end{minipage} & \begin{minipage}{.1\linewidth}
  \includegraphics[width=\linewidth]{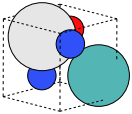}
\end{minipage} & \begin{minipage}{.1\linewidth}
  \includegraphics[width=\linewidth]{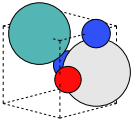}
\end{minipage} & \begin{minipage}{.1\linewidth}
  \includegraphics[width=\linewidth]{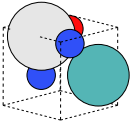}
\end{minipage} & \begin{minipage}{.1\linewidth}
  \includegraphics[width=\linewidth]{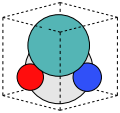}
\end{minipage} 
\\

\midrule
ScNO2Ga is alpha Rhenium trioxide-derived structured and crystallizes in the orthorhombic Pmm2 space group. The structure consists of one Ga cluster inside a ScNO2 framework. In the Ga cluster, Ga is bonded in a 1-coordinate geometry to  atoms. In the ScNO2 framework, Sc is bonded to two equivalent N, two equivalent O, and two equivalent O atoms to form corner-sharing ScN2O4 octahedra. The corner-sharing octahedral tilt angles range from 0-1°. Both Sc-N bond lengths are 2.09 Å. Both Sc-O bond lengths are 2.09 Å. Both Sc-O bond lengths are 2.09 Å. N is bonded in a linear geometry to two equivalent Sc atoms. There are two inequivalent O sites. In the first O site, O is bonded in a linear geometry to two equivalent Sc atoms. In the second O site, O is bonded in a linear geometry to two equivalent Sc atoms.The formation energy per atom is 1.4796.&
Below is a description of a bulk material. The chemical formula is ScGaNO2. The elements are Sc, Ga, N, O. The formation energy per atom is positive. The spacegroup number is 24. The crystal system is orthorhombic. Generate the material:& 
\begin{minipage}{.1\linewidth}
  \includegraphics[width=\linewidth]{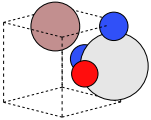}
\end{minipage} & \begin{minipage}{.1\linewidth}
  \includegraphics[width=\linewidth]{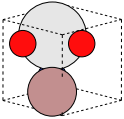}
\end{minipage} & \begin{minipage}{.1\linewidth}
  \includegraphics[width=\linewidth]{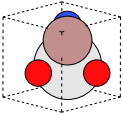}
\end{minipage} & \begin{minipage}{.1\linewidth}
  \includegraphics[width=\linewidth]{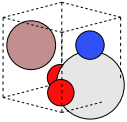}
\end{minipage} & \begin{minipage}{.1\linewidth}
  \includegraphics[width=\linewidth]{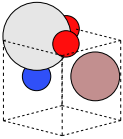}
\end{minipage} 
\\

\midrule
OsAuO3 is (Cubic) Perovskite structured and crystallizes in the cubic Pm-3m space group. Os is bonded to six equivalent O atoms to form OsO6 octahedra that share corners with six equivalent OsO6 octahedra and faces with eight equivalent AuO12 cuboctahedra. The corner-sharing octahedra are not tilted. All Os-O bond lengths are 1.97 Å. Au is bonded to twelve equivalent O atoms to form distorted AuO12 cuboctahedra that share corners with twelve equivalent AuO12 cuboctahedra, faces with six equivalent AuO12 cuboctahedra, and faces with eight equivalent OsO6 octahedra. All Au-O bond lengths are 2.79 Å. O is bonded in a linear geometry to two equivalent Os and four equivalent Au atoms.The formation energy per atom is 1.4248.&
Below is a description of a bulk material. The chemical formula is OsAuO3. The elements are Os, Au, O. The formation energy per atom is 1.4248. The spacegroup number is 220. The crystal system is cubic. Generate the material.& 
\begin{minipage}{.1\linewidth}
  \includegraphics[width=\linewidth]{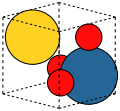}
\end{minipage} & \begin{minipage}{.1\linewidth}
  \includegraphics[width=\linewidth]{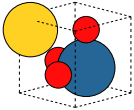}
\end{minipage} & \begin{minipage}{.1\linewidth}
  \includegraphics[width=\linewidth]{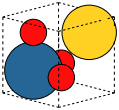}
\end{minipage} & \begin{minipage}{.1\linewidth}
  \includegraphics[width=\linewidth]{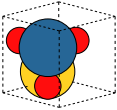}
\end{minipage} & \begin{minipage}{.1\linewidth}
  \includegraphics[width=\linewidth]{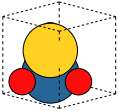}
\end{minipage} 
\\

\bottomrule
\end{tabular}
 }
   \caption{Visualization of the generated structures given textual description for \textbf{Perov-5} dataset}
  \label{tbl-samples_perov_appendix}
\end{table*}
%%%%%%%%% Carbon-Sample %%%%%%%%%%%%%%%
\begin{table*}[ht]
  \centering
    \setlength{\tabcolsep}{1.5 pt}
    \renewcommand{\arraystretch}{1}
    \resizebox{1.0\textwidth}{!}{
    % \scalebox{0.7}{
    \begin{tabular}{m{3in}|m{2.2in}|c|cccc| } % angstrom symbol Å
\toprule
Detailed Description & Short Prompt &  Ground truth & \multicolumn{4}{c|}{Generated Samples} \\
\midrule
 
C crystallizes in the triclinic P1 space group. There are twenty-two inequivalent C sites. In the first C site, C(1) is bonded to one C(18), one C(5), and two equivalent C(9) atoms to form corner-sharing CC4 tetrahedra. \dots two equivalent C(12) atoms to form a mixture of distorted corner and edge-sharing CC4 trigonal pyramids. The energy per atom is -154.1336. & 
Below is a description of a bulk material. The chemical formula is C. The elements are C. The energy per atom is negative. The spacegroup number is 0. The crystal system is triclinic. Generate the material& 
\begin{minipage}{.1\linewidth}
  \includegraphics[width=\linewidth]{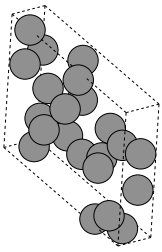}
\end{minipage} & 
\begin{minipage}{.1\linewidth}
  \includegraphics[width=\linewidth]{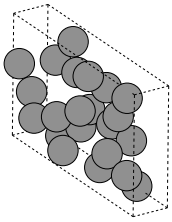}
\end{minipage} & 
\begin{minipage}{.1\linewidth}
  \includegraphics[width=\linewidth]{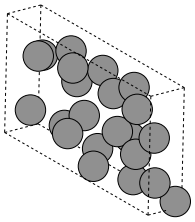}
\end{minipage} & 
\begin{minipage}{.1\linewidth}
  \includegraphics[width=\linewidth]{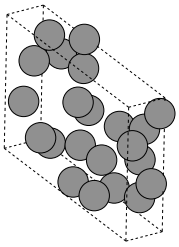}
\end{minipage} & 
\begin{minipage}{.1\linewidth}
  \includegraphics[width=\linewidth]{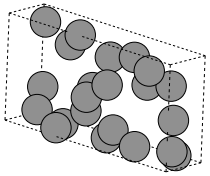}
\end{minipage} 
\\
  
\midrule
C crystallizes in the orthorhombic Cmcm space group. There are two inequivalent C sites. In the first C site, C(1) is bonded to one C(2) and three equivalent C(1) atoms to form a mixture of corner and edge-sharing CC4 trigonal pyramids. The C(1)-C(2) bond length is 1.49 Å. There are two shorter (1.51 Å) and one longer (1.56 Å) C(1)-C(1) bond length. In the second C site, C(2) is bonded to one C(1) and three equivalent C(2) atoms to form corner-sharing CC4 tetrahedra. There are two shorter (1.54 Å) and one longer (1.56 Å) C(2)-C(2) bond length. The energy per atom is -154.2425.&
Below is a description of a bulk material. The chemical formula is C. The elements are C. The energy per atom is negative. The spacegroup number is 62. The crystal system is orthorhombic. Generate the material.& 
\begin{minipage}{.1\linewidth}
  \includegraphics[width=\linewidth]{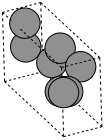}
\end{minipage} & 
\begin{minipage}{.1\linewidth}
  \includegraphics[width=\linewidth]{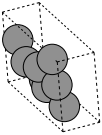}
\end{minipage} & 
\begin{minipage}{.1\linewidth}
  \includegraphics[width=\linewidth]{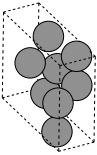}
\end{minipage} & 
\begin{minipage}{.1\linewidth}
  \includegraphics[width=\linewidth]{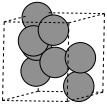}
\end{minipage} & 
\begin{minipage}{.1\linewidth}
  \includegraphics[width=\linewidth]{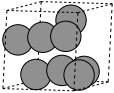}
\end{minipage} 
\\

\midrule
C crystallizes in the triclinic P-1 space group. There are six inequivalent C sites. In the first C site, C(1) is bonded to one C(3), one C(5), and two equivalent C(4) atoms to form corner-sharing CC4 tetrahedra. \dots In the sixth C site, C(6) is bonded to one C(2), one C(4), and two equivalent C(3) atoms to form distorted corner-sharing CC4 tetrahedra. The energy per atom is -154.1338.&
Below is a description of a bulk material. The chemical formula is C. The elements are C. The energy per atom is negative. The spacegroup number is 1. The crystal system is triclinic. Generate the material& 
\begin{minipage}{.1\linewidth}
  \includegraphics[width=\linewidth]{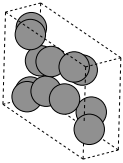}
\end{minipage} & 
\begin{minipage}{.1\linewidth}
  \includegraphics[width=\linewidth]{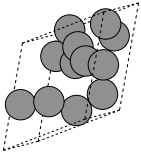}
\end{minipage} & 
\begin{minipage}{.1\linewidth}
  \includegraphics[width=\linewidth]{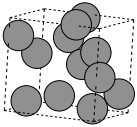}
\end{minipage} & 
\begin{minipage}{.1\linewidth}
  \includegraphics[width=\linewidth]{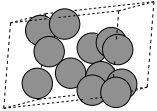}
\end{minipage} & 
\begin{minipage}{.1\linewidth}
  \includegraphics[width=\linewidth]{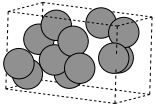}
\end{minipage} 
\\

\midrule
C is a Theoretical Carbon Structure-like structure and crystallizes in the triclinic P-1 space group. There are nine inequivalent C sites. In the first C site, C(1) is bonded to one C(5), one C(6), one C(7), and one C(8) atom to form a mixture of corner and edge-sharing CC4 tetrahedra. The C(1)-C(5) bond length is 1.51 Å. The C(1)-C(6) bond length is 1.56 Å. The C(1)-C(7) bond length is 1.54 Å. \dots In the ninth C site, C(9) is bonded to one C(4), one C(6), one C(7), and one C(8) atom to form a mixture of corner and edge-sharing CC4 tetrahedra. The energy per atom is -154.2197.&
Below is a description of a bulk material. The chemical formula is C. The elements are C. The energy per atom is -154.2197. The spacegroup number is 1. The crystal system is triclinic. Generate the material.& 
\begin{minipage}{.1\linewidth}
  \includegraphics[width=\linewidth]{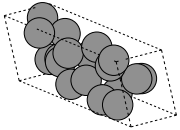}
\end{minipage} & 
\begin{minipage}{.1\linewidth}
  \includegraphics[width=\linewidth]{Images/carbon_samples/556/1.png}
\end{minipage} & 
\begin{minipage}{.1\linewidth}
  \includegraphics[width=\linewidth]{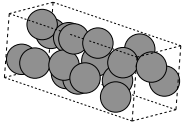}
\end{minipage} & 
\begin{minipage}{.1\linewidth}
  \includegraphics[width=\linewidth]{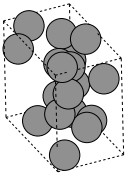}
\end{minipage} & 
\begin{minipage}{.1\linewidth}
  \includegraphics[width=\linewidth]{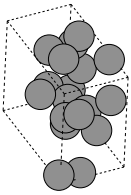}
\end{minipage} 
\\

\bottomrule
\end{tabular}
 }
   \caption{Visualization of the generated structures given textual description for \textbf{Carbon-24} dataset}
  \label{tbl-samples_carbon_appendix}
\end{table*}
%%%%%%%%% MP-Sample %%%%%%%%%%%%%%%
\begin{table*}[ht]
\centering
\setlength{\tabcolsep}{1.5 pt}
\renewcommand{\arraystretch}{1}
\resizebox{1.0\textwidth}{!}{
% \scalebox{0.7}{
\begin{tabular}{m{3in}|m{2.2in}|c|cccc| } % angstrom symbol Å
\toprule
Detailed Description & Short Prompt &  Ground truth & \multicolumn{4}{c|}{Generated Samples} \\
\midrule
 
Eu2PCl is Caswellsilverite-like structured and crystallizes in the tetragonal P4/mmm space group. There are two inequivalent Eu sites. In the first Eu site, Eu(1) is bonded to two equivalent P(1) and four equivalent Cl(1) atoms to form EuP2Cl4 octahedra that share corners with six equivalent Eu(1)P2Cl4 octahedra, edges with four equivalent Eu(1)P2Cl4 octahedra, and edges with eight equivalent Eu(2)P4Cl2 octahedra. \dots The corner-sharing octahedra are not tilted. The formation energy per atom is -1.7615. The band gap is zero. The energy above the convex hull is zero.& 
Below is a description of a bulk material. The chemical formula is Eu2PCl. The elements are Eu, P, and Cl. The formation energy per atom is -1.7615. The band gap is 0.0. The energy above the convex hull is 0.0. The spacegroup number is 122. The crystal system is tetragonal. Generate the material.& 
\begin{minipage}{.1\linewidth}
  \includegraphics[width=\linewidth]{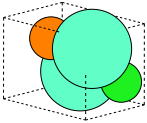}
\end{minipage} & 
\begin{minipage}{.1\linewidth}
  \includegraphics[width=\linewidth]{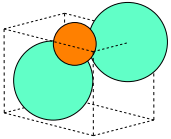}
\end{minipage} & 
\begin{minipage}{.1\linewidth}
  \includegraphics[width=\linewidth]{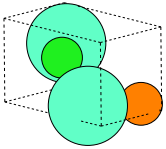}
\end{minipage} & 
\begin{minipage}{.1\linewidth}
  \includegraphics[width=\linewidth]{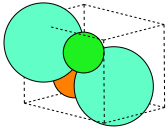}
\end{minipage} & 
\begin{minipage}{.1\linewidth}
  \includegraphics[width=\linewidth]{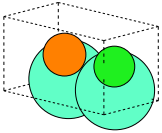}
\end{minipage} 
\\

\midrule
MgNdHg2 is Heusler structured and crystallizes in the cubic Fm-3m space group. Mg(1) is bonded in a body-centered cubic geometry to eight equivalent Hg(1) atoms. All Mg(1)-Hg(1) bond lengths are 3.18 Å. Nd(1) is bonded in a body-centered cubic geometry to eight equivalent Hg(1) atoms. All Nd(1)-Hg(1) bond lengths are 3.18 Å. Hg(1) is bonded in a body-centered cubic geometry to four equivalent Mg(1) and four equivalent Nd(1) atoms. The formation energy per atom is -0.4708. The band gap is 0.0. The energy above the convex hull is 0.0. The spacegroup number is 224.& 
Below is a description of a bulk material. The chemical formula is NdMgHg2. The elements are Nd, Mg, and Hg. The formation energy per atom is -0.4708. The band gap is 0.0. The energy above the convex hull is 0.0. The spacegroup number is 224. The crystal system is cubic. Generate the material.& 
\begin{minipage}{.1\linewidth}
  \includegraphics[width=\linewidth]{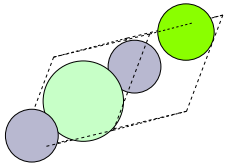}
\end{minipage} & 
\begin{minipage}{.1\linewidth}
  \includegraphics[width=\linewidth]{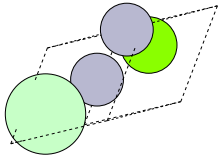}
\end{minipage} & 
\begin{minipage}{.1\linewidth}
  \includegraphics[width=\linewidth]{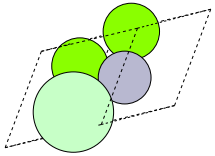}
\end{minipage} & 
\begin{minipage}{.1\linewidth}
  \includegraphics[width=\linewidth]{Images/mp_samples/713/1.png}
\end{minipage} & 
\begin{minipage}{.1\linewidth}
  \includegraphics[width=\linewidth]{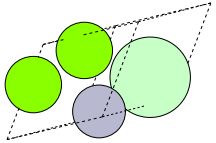}
\end{minipage} 
\\

\midrule
MgNdTl crystallizes in the hexagonal P-62m space group. Mg(1) is bonded in a 4-coordinate geometry to two equivalent Tl(1) and two equivalent Tl(2) atoms. Both Mg(1)-Tl(1) bond lengths are 3.01 Å. Both Mg(1)-Tl(2) bond lengths are 3.03 Å. Nd(1) is bonded in a 5-coordinate geometry to one Tl(2) and four equivalent Tl(1) atoms. The Nd(1)-Tl(2) bond length is 3.31 Å. All Nd(1)-Tl(1) bond lengths are 3.32 Å. There are two inequivalent Tl sites. In the first Tl site, Tl(2) is bonded in a distorted q6 geometry to six equivalent Mg(1) and three equivalent Nd(1) atoms. In the second Tl site, Tl(1) is bonded in a 9-coordinate geometry to three equivalent Mg(1) and six equivalent Nd(1) atoms. The formation energy per atom is -0.355. The band gap is 0.0. The energy above the convex hull is 0.0. The spacegroup number is 188.& 
Below is a description of a bulk material. The chemical formula is NdMgTl. The elements are Nd, Mg, and Tl. The formation energy per atom is -0.355. The band gap is 0.0. The energy above the convex hull is 0.0. The spacegroup number is 188. The crystal system is hexagonal. Generate the material.& 
\begin{minipage}{.1\linewidth}
  \includegraphics[width=\linewidth]{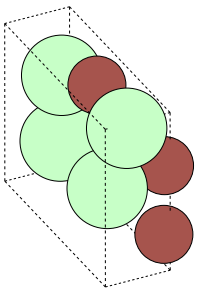}
\end{minipage} & 
\begin{minipage}{.1\linewidth}
  \includegraphics[width=\linewidth]{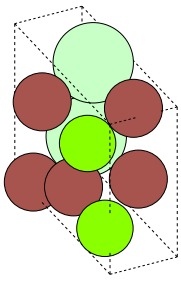}
\end{minipage} & 
\begin{minipage}{.1\linewidth}
  \includegraphics[width=\linewidth]{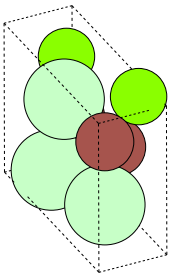}
\end{minipage} & 
\begin{minipage}{.1\linewidth}
  \includegraphics[width=\linewidth]{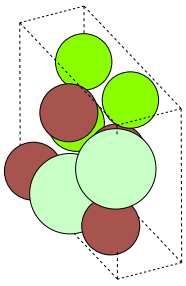}
\end{minipage} & 
\begin{minipage}{.1\linewidth}
  \includegraphics[width=\linewidth]{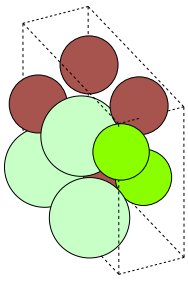}
\end{minipage} 
\\

\midrule
LaNi2Ge2 crystallizes in the tetragonal I4/mmm space group. La(1) is bonded in a 16-coordinate geometry to eight equivalent Ni(1) and eight equivalent Ge(1) atoms. All La(1)-Ni(1) bond lengths are 3.25 Å. All La(1)-Ge(1) bond lengths are 3.26 Å. Ni(1) is bonded in a 4-coordinate geometry to four equivalent La(1) and four equivalent Ge(1) atoms. All Ni(1)-Ge(1) bond lengths are 2.39 Å. Ge(1) is bonded in a 9-coordinate geometry to four equivalent La(1), four equivalent Ni(1), and one Ge(1) atom. The Ge(1)-Ge(1) bond length is 2.66 Å. The formation energy per atom is -0.691. The band gap is 0.0. The energy above the convex hull is 0.0.& 
Below is a description of a bulk material. The chemical formula is La(NiGe)2. The elements are La, Ni, and Ge. The formation energy per atom is -0.691. The band gap is 0.0. The energy above the convex hull is 0.0. The spacegroup number is 138. The crystal system is tetragonal. Generate the material.& 
\begin{minipage}{.1\linewidth}
  \includegraphics[width=\linewidth]{Images/mp_samples/1681/1.png}
\end{minipage} & 
\begin{minipage}{.1\linewidth}
  \includegraphics[width=\linewidth]{Images/mp_samples/1681/2.png}
\end{minipage} & 
\begin{minipage}{.1\linewidth}
  \includegraphics[width=\linewidth]{Images/mp_samples/1681/3.png}
\end{minipage} & 
\begin{minipage}{.1\linewidth}
  \includegraphics[width=\linewidth]{Images/mp_samples/1681/5.png}
\end{minipage} & 
\begin{minipage}{.1\linewidth}
  \includegraphics[width=\linewidth]{Images/mp_samples/1681/5.png}
\end{minipage} 
\\
\bottomrule
\end{tabular}
 }
   \caption{Visualization of the generated structures given textual description for \textbf{MP-20} dataset}
  \label{tbl-samples_mp_appendix}
\end{table*}
\newpage

\end{document}